\documentclass[10pt,twocolumn,letterpaper]{article}

\PassOptionsToPackage{table}{xcolor}
\usepackage[pagenumbers]{cvpr} %
\usepackage{svg}

\usepackage{cuted} %

\usepackage{currfile} %

\usepackage{caption} %

\usepackage{pgf}
\usepackage{tikz}
\usepackage{pifont}
\usepackage{multirow}
\usepackage{pgfplots}
\usepackage{adjustbox}

\newcommand{\first}[1]{\cellcolor{red!25}\textbf{#1}}
\newcommand{\second}[1]{\cellcolor{orange!25}\textbf{#1}}
\newcommand{\third}[1]{\cellcolor{yellow!35}\textbf{#1}}
\newcommand{\cmark}{\ding{51}}
\newcommand{\xmark}{\ding{55}}

\newcommand{\reviewonly}[1]{}
\usepackage{afterpage}
\makeatletter
\newcommand{\beforereferences}{%
  \if@firstcolumn\afterpage{\afterpage{\clearpage}}\else\afterpage{\clearpage}\fi}
\makeatother

\definecolor{cvprblue}{rgb}{0.21,0.49,0.74}
\usepackage[pagebackref,breaklinks,colorlinks,allcolors=cvprblue]{hyperref}

\def\confName{CVPR}
\def\confYear{2027}

\title{GTR-Mesh: Geometry-Texture-Ready Mesh Reconstruction \\ from Multiview Images}

\author{
David Berger\thanks{Both authors contributed equally to the paper.} \qquad
Simon Niedermayr\footnotemark[1] \qquad
R\"udiger Westermann \\[0.6em]
Technical University of Munich
}

\begin{document}

\maketitle

\begin{abstract}
We present GTR-Mesh, a fully differentiable pipeline that directly reconstructs watertight, adaptive, textured triangle meshes of unbounded scenes from multiview images. 
We tetrahedralize a point-based SDF representation and employ multiple implicit surfaces to induce non-local gradient flow for shape and appearance optimization. Surfaces are rendered with a novel differentiable triangle rasterizer with analytic anti-aliasing and progressively collapsed into a single opaque, watertight surface free of self-intersections. 
Gradient-driven refinement of the point cloud adds geometry only where the scene demands it. Appearance is modeled by a volumetric multi-resolution hash grid, which decouples view-independent color details from geometric resolution. After optimization, the hash grid is baked into a 2D texture atlas, yielding textured meshes in a shared-vertex representation. GTR-Mesh represents scenes with up to 3.7$\times$ fewer vertices than other watertight mesh reconstruction methods while achieving higher quality and rendering speed in novel view synthesis. 

\end{abstract}

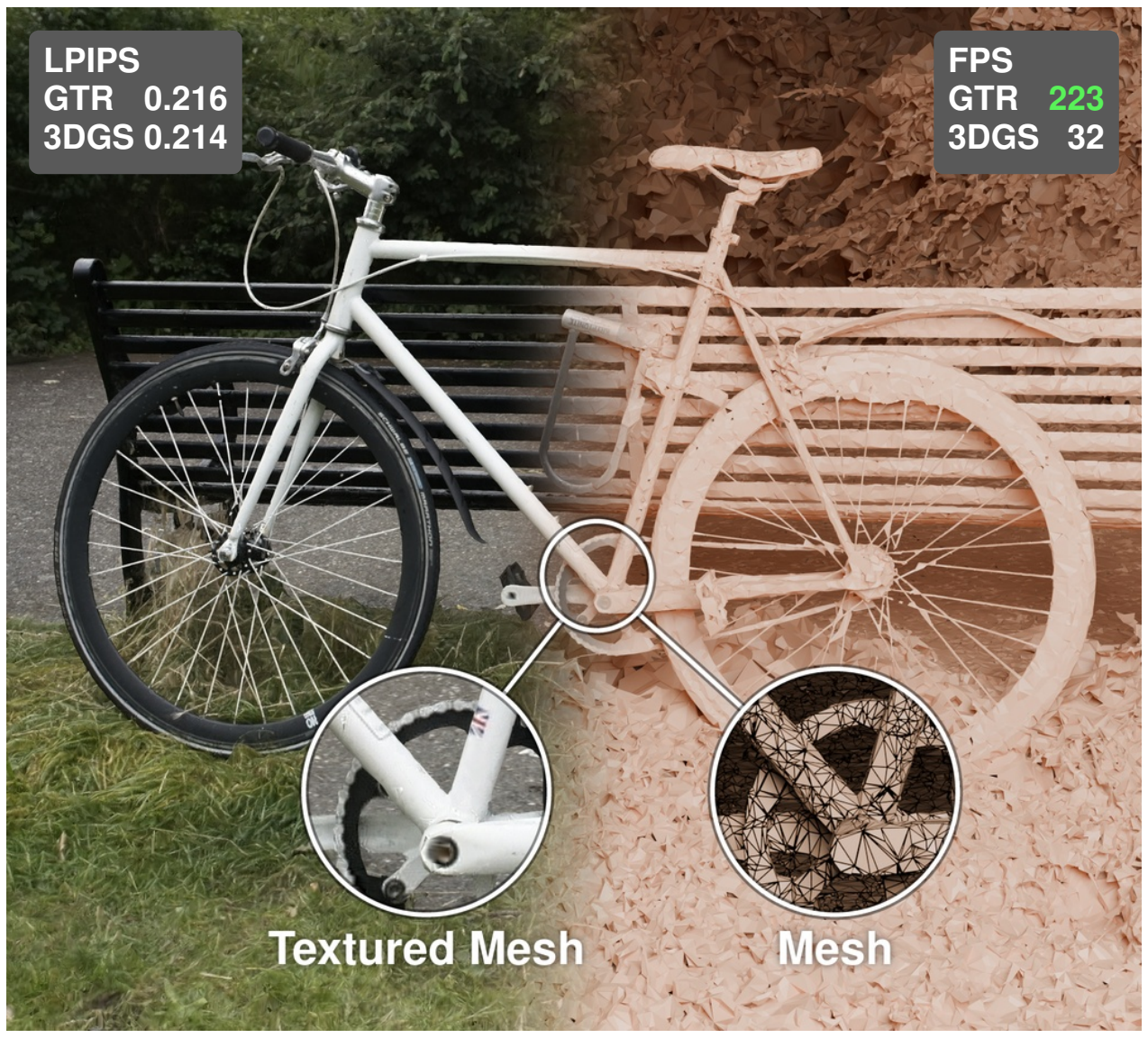
\begin{figure}[t]
  \centering
  \setlength{\unitlength}{\linewidth}%
  \begin{picture}(1,0.9004)
    \put(0,0){\includegraphics[width=\linewidth]{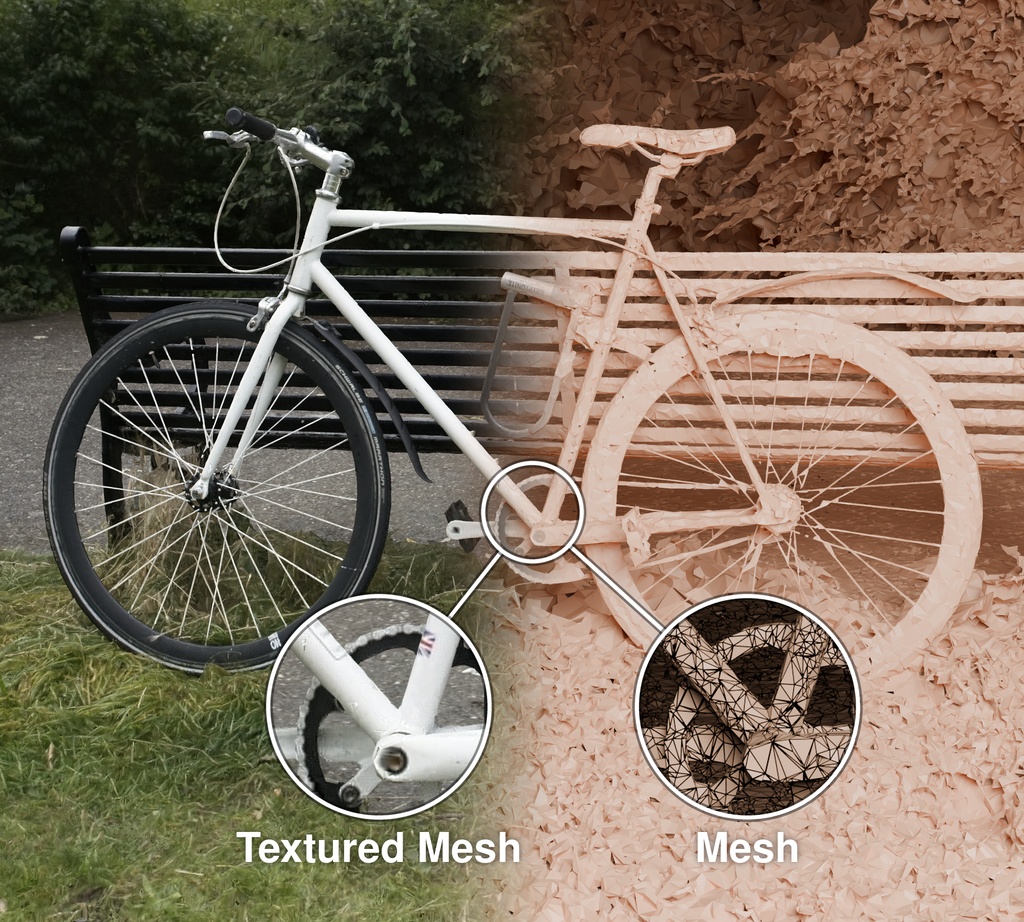}}
    \put(0.02,0.8804){\makebox(0,0)[tl]{\tikz\node[fill=black!65,rounded corners=2pt,inner sep=3pt,text=white,font=\sffamily\bfseries\scriptsize]{\begin{tabular}{@{}l@{\ }l@{}}\multicolumn{2}{@{}l@{}}{LPIPS}\\ GTR & 0.216\\ 3DGS & 0.214\end{tabular}};}}
    \put(0.98,0.8804){\makebox(0,0)[tr]{\tikz\node[fill=black!65,rounded corners=2pt,inner sep=3pt,text=white,font=\sffamily\bfseries\scriptsize]{\begin{tabular}{@{}l@{\ }r@{}}\multicolumn{2}{@{}l@{}}{FPS}\\ GTR & \textcolor[rgb]{0.35,0.95,0.35}{223}\\ 3DGS & 32\end{tabular}};}}
  \end{picture}
  \captionsetup{hypcap=false}
  \captionof{figure}{\textbf{GTR-Mesh} 
  collapses multiple semi-transparent SDF shells into a watertight and textured triangle mesh. 
  It achieves higher quality in novel view synthesis with fewer vertices than other watertight mesh methods,
  and directly produces a textured shared-vertex mesh. Frame rates measured on a MacBook Air M3.}
  \captionsetup{hypcap=true}
  \label{fig:teaser}
\end{figure}

\section{Introduction}

Novel view synthesis (NVS) has progressed rapidly with the move from implicit volumetric representations such as NeRF~\cite{mildenhall_nerf_2021} and Instant-NGP~\cite{muller_instant_2022} to explicit, primitive-based ones such as 3D Gaussian Splatting (3DGS)~\cite{kerbl_3d_2023}.
However, these representations model the radiance field of a scene but do not describe its geometry.
Textured meshes, in contrast, are directly usable in applications that require well-defined surfaces, such as physics simulation and collision detection, and they
can be efficiently edited, animated, and rendered with industry-standard tools such as game engines and 3D modeling software.
Because neighboring triangles share vertices and fine appearance detail is stored in textures rather than additional geometry, meshes are also compact, and hardware rasterization renders them at high frame rates on a wide range of devices.

Several recent methods therefore reconstruct meshes with different optimization strategies.
Triangle Splatting~\cite{held2025triangle} optimizes freely floating triangles and produces a self-intersecting triangle soup.
Both MeshSplatting by Held~\etal~\cite{held2026mesh} and the similarly named Mesh Splatting by Zhang~\etal~\cite{zhang2026meshsplatting} optimize a connected mesh, but the exported meshes can still contain non-manifold structures and self-intersections.  
MILo~\cite{guedon2025milo} and Blobs to Spokes~\cite{gomez2026gaussianwrapping} extract watertight meshes from 3D Gaussians, but the Gaussians remain the primitive that is rendered and supervised during training, and the mesh only serves as a geometric constraint or as the extraction target.
All of these methods store appearance in the geometry itself, either as per-vertex attributes or in the primitives from which the mesh is derived.
To reproduce fine texture detail, they need many small triangles, leading to high vertex counts.

To our knowledge, no existing method optimizes a textured, geometry-adaptive manifold mesh end-to-end with a photometric loss on large unbounded scenes.
We close this gap with GTR-Mesh.
GTR-Mesh iteratively extracts several semi-transparent isosurfaces---called \emph{shells}---from a trainable tetrahedral SDF mesh. 
Instead of deferred rendering at $3\times$ supersampled resolution with a non-differentiable rasterizer~\cite{charatan2026meshtryoshka}, we introduce a differentiable triangle rasterizer with analytic anti-aliasing that is $2.8\times$ faster and needs $3.3\times$ less memory than depth peeling with \emph{nvdiffrast}~\cite{Laine2020diffrast}.
Because several shells surround the current surface, each pixel receives gradients from many tetrahedra along its ray.
The surface can therefore move towards target geometry within the region covered by the shells, and even change its topology, instead of being dragged there vertex by vertex.
During training, the shells' opacity is gradually sharpened until they collapse into a single opaque surface.

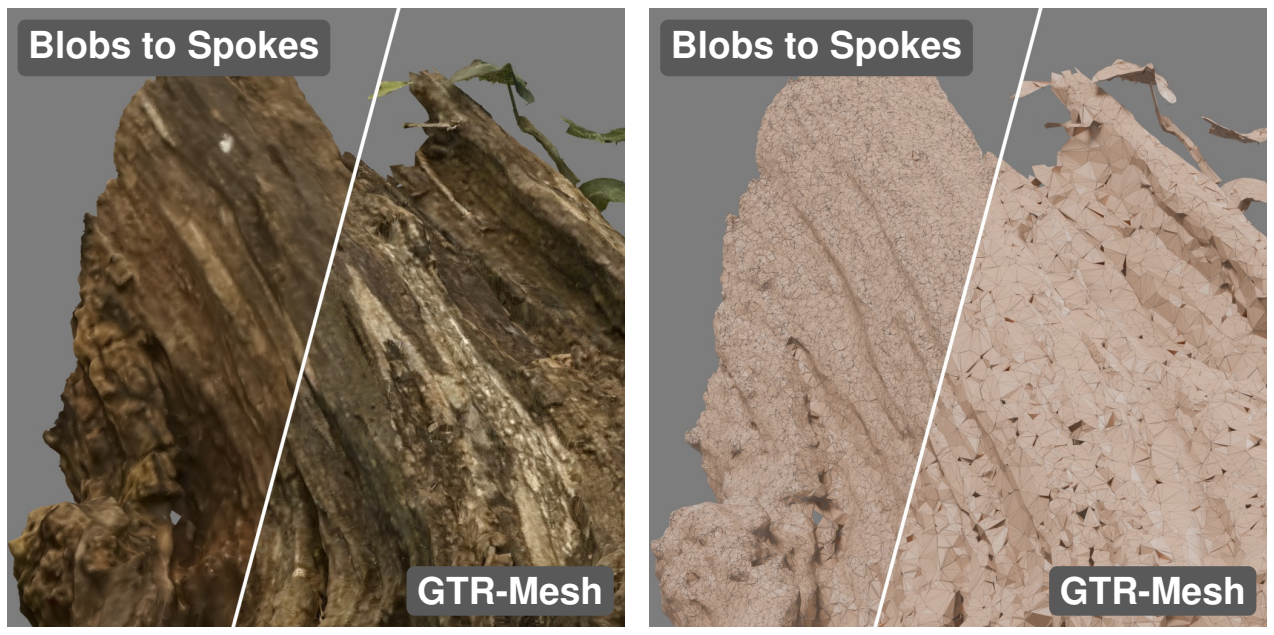
\begin{figure}[tbp]
  \centering
  \newcommand{\meshsplit}[2]{%
    \begin{tikzpicture}
      \node[inner sep=0, anchor=south west] (img) at (0,0) {\includegraphics[width=0.49\linewidth]{#1}};
      \tikzset{lbl/.style={fill=black!65, rounded corners=1.5pt, inner sep=2pt, text=white, font=\sffamily\bfseries\scriptsize}}
      \node[lbl, anchor=north west] at ([shift={(2pt,-2pt)}]img.north west) {Blobs to Spokes};
      \node[lbl, anchor=south east] at ([shift={(-2pt,2pt)}]img.south east) {GTR-Mesh};
      \node[anchor=north, inner sep=1pt] at (img.south) {\small\strut #2};
    \end{tikzpicture}}%
  \meshsplit{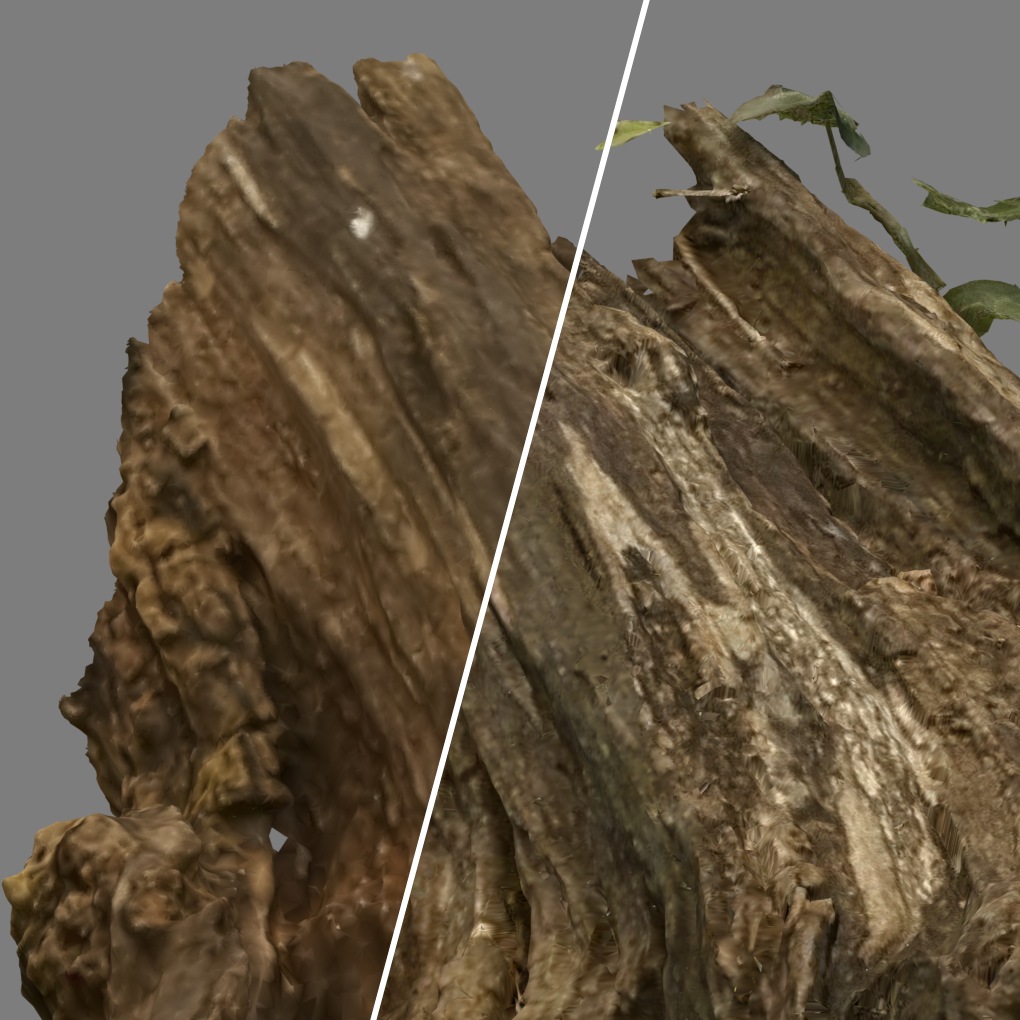}{Appearance}\hfill
  \meshsplit{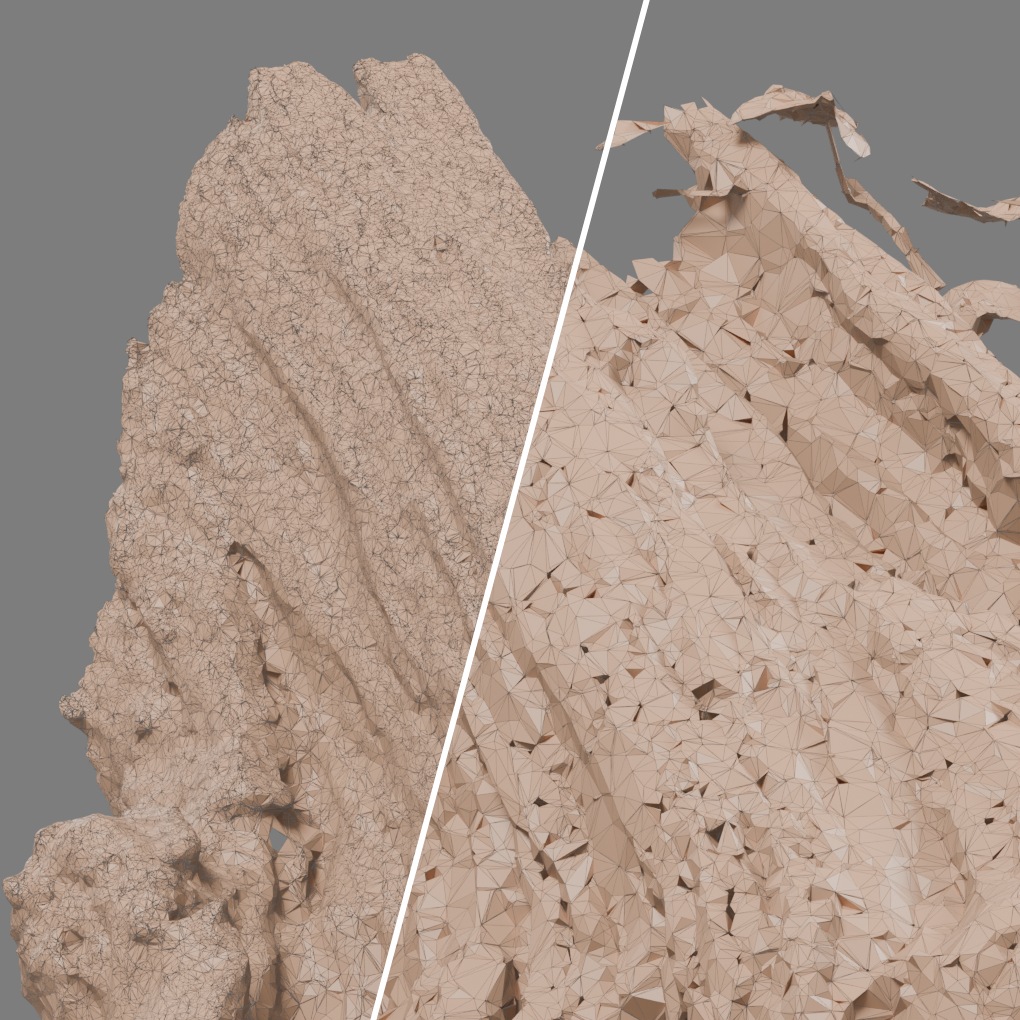}{Geometry}
  \caption{Meshes reconstructed from \textit{stump} with Blobs to Spokes~\cite{gomez2026gaussianwrapping} and GTR-Mesh.}
  \label{fig:mesh}
\end{figure}

Appearance is stored in a multi-resolution hash grid~\cite{muller_instant_2022} that is queried at the rasterized fragment positions and finally baked into a 2D texture.
Gradient-driven refinement adds new points adaptively where the current geometry is too coarse. 
The result is a watertight, geometry-adaptive, textured mesh with shared vertices that can be rendered by standard graphics pipelines.
GTR-Mesh 
matches 3DGS in LPIPS and outperforms all previous mesh-based methods, while using fewer vertices than other watertight methods (Tab.~\ref{tab:main}, Fig.~\ref{fig:mesh}).
Our specific contributions are:
\begin{itemize}
    \item \textbf{Shell-sampled SDF optimization.} A differentiable, end-to-end formulation that bridges volume rendering and triangle rasterization, yielding render-ready 
    meshes for unbounded scenes under a single photometric loss.
    \item \textbf{Anti-aliasing gradient rasterizer.} An efficient differentiable triangle rasterizer with fragment sorting, alpha compositing, and analytic subpixel anti-aliasing. 
\end{itemize}

\begin{figure*}[tbp]
    \centering
    \includegraphics[width=\linewidth]{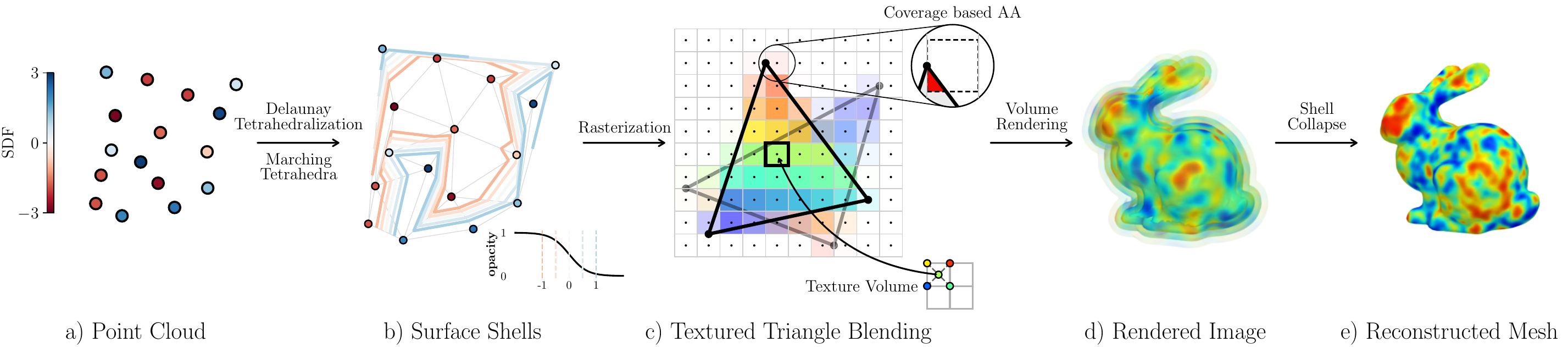}
    \caption{Method overview. GTR-Mesh starts from a point cloud with SDF values (a), and computes a Delaunay tet-mesh from which multiple semi-transparent shells are extracted using Marching Tetrahedra (b). Shells are rasterized with coverage-based anti-aliasing and fragment colors queried from a 3D neural hash grid (c). Shells are alpha-composited into the final image (d), collapsing into a single opaque, textured mesh with increasing temperature (e).}
    \label{fig:pipeline}
\end{figure*}

\section{Related work}

\subsection{Radiance fields}

NeRF~\cite{mildenhall_nerf_2021} represents scenes as continuous fields of density and color, but rendering requires hundreds of samples per ray.
3DGS~\cite{kerbl_3d_2023} replaces this with millions of explicit anisotropic Gaussians that are rasterized and alpha-blended, and follow-up work has further improved its training and rendering speed~\cite{niedermayr2025upscaling3dgs,kelkar2026hybrid,hahlbohm2026fastergs,ren2026fastgs,HansonSpeedy}.
Still, these representations rely on blending many semi-transparent primitives per pixel and lack the surface connectivity that graphics pipelines and downstream applications expect.

\subsection{Surface reconstruction with Gaussian splatting}
Many methods extract meshes from Gaussian representations.
They align the Gaussians with the surface~\cite{guedon2024sugar,huang20242d}, improve the rendered geometry and its supervision~\cite{zhang2024radegs,chen2024pgsr,radl2026come}, or extract level sets of an opacity field~\cite{yu2024gof}.
MILo~\cite{guedon2025milo} integrates a Delaunay-based mesh extraction into the training loop.
Blobs to Spokes~\cite{gomez2026gaussianwrapping} derives a closed-form occupancy field from oriented Gaussians to extract watertight meshes of complete scenes.
In all of these methods, however, the Gaussians remain the primitive that is rendered and supervised.
The mesh is derived from them, and its appearance is limited to per-vertex colors.

\subsection{Mesh-based scene representations}

NeuS~\cite{wang2021neus} and VolSDF~\cite{yariv2021volume} optimize an implicit SDF that is converted into a mesh after training, which introduces discretization artifacts and lowers rendering quality.
MobileNeRF~\cite{chen2023mobilenerf}, BakedSDF~\cite{yariv2023bakedsdf}, NeRF2Mesh~\cite{tang2023nerf2mesh}, LTM~\cite{Choi_2024_CVPR}, and Binary Opacity Grids~\cite{reiser2024binary} bake radiance fields into meshes for real-time rendering, but the mesh is still derived from a volumetric representation.
NVDiffRec~\cite{munkberg2022extracting} optimizes a mesh directly with differentiable rasterization, but requires foreground masks and does not scale to large scenes.

Triangle Splatting~\cite{held2025triangle} and its extensions~\cite{jimenez2026deformable,tojo2026diffsoup} optimize triangle primitives directly, but produce unconnected triangle soups.
MeshSplatting~\cite{held2026mesh} optimizes connected, opaque meshes, though the exported geometry still contains non-manifold structures and fragments, resulting from local mesh operations.
Adaptive Shells~\cite{wang2023adaptive} restrict volume rendering to a narrow band between two meshes, and Volumetric Surfaces~\cite{esposito2025volumetric} represent fuzzy geometry with a fixed number of layered, semi-transparent meshes; both still rely on a volumetric representation during rendering.
Closest to our work, Zhang~\etal~\cite{zhang2026meshsplatting} and Meshtryoshka~\cite{charatan2026meshtryoshka} render several semi-transparent shells around the surface.
Zhang~\etal use continuous remeshing~\cite{werner2022continuous} and rely on a surface parameterization whose grid resolution and extent are fixed, which limits their method to small, bounded scenes. 
Meshtryoshka scales to unbounded scenes. However, since it uses a non-differentiable rasterizer, the geometry is only updated through per-vertex opacities.
Its appearance is stored per vertex on a sparse voxel grid, which the authors report leads to large triangle counts and about four hours of training.

\subsection{Differentiable mesh rendering}

DMTet~\cite{shen2021deep} and FlexiCubes~\cite{shen2023flexicubes} reformulate the mesh extraction so that the output geometry is a continuous, differentiable function of learned parameters, but they assume a fixed background grid, limiting scalability to unbounded scenes. TetWeave~\cite{binninger2025tetweave} extends Marching Tetrahedra (MT) to unstructured point clouds, while DMesh~\cite{son2024dmesh} differentiably models face existence probabilities on a weighted Delaunay triangulation. 
Radiance Meshes~\cite{mai2026radiancemeshes} perform volume rendering on Delaunay tetrahedral cells but represent volumetric density rather than an explicit surface. Differentiable rasterizers such as Soft Rasterizer~\cite{liu2019softras} and \emph{nvdiffrast}~\cite{Laine2020diffrast} target a single opaque surface per pixel. In contrast, our rasterizer sorts and alpha-composites multiple semi-transparent fragments per pixel with anti-aliased subpixel coverage, operating on shells extracted from an unstructured Delaunay tetrahedralization of the point cloud.

\section{Method}

GTR-Mesh represents the scene geometry as an SDF defined on an unstructured point cloud (Fig.~\ref{fig:pipeline}).
Unlike grid-based SDFs, a point cloud adapts naturally to unbounded scenes and spatially varying geometric detail.
The Delaunay tetrahedralization defines a continuous field via barycentric interpolation within each tetrahedron. Both point positions and their SDF values are optimized, and as points move, the tetrahedralization is updated periodically to ensure connectivity follows the points. In each iteration, GTR-Mesh extracts several semi-transparent SDF isosurfaces---called shells---using MT (Sec.~\ref{sec:shells}). 
To keep all shells closed, the SDF of points on the convex hull is constrained to stay outside the outermost shell, so MT yields watertight shells without self-intersections. GTR-Mesh renders the shells with a novel differentiable rasterizer (Sec.~\ref{sec:rendering}), using fragment colors captured by a hash-grid appearance model (Sec.~\ref{sec:color_model}). 

\subsection{Initialization}
The point cloud is initialized with the Structure-from-Motion (SfM) points computed with COLMAP~\cite{schoenberger2016sfm}.
For each SfM point, we add one support point displaced by a random offset proportional to the distance to its nearest SfM neighbor.
SfM points receive a zero SDF value and support points a positive value, excluding high-opacity inner shells for greater initial transparency.

\subsection{Shell extraction from SDF}\label{sec:shells}

Following occupancy-based formulations such as UNISURF~\cite{oechsle2021unisurf}, SDF values $s$ are mapped to opacity with a sigmoid function,
\begin{equation}
    \alpha(s) = \sigma(-T s),
\end{equation}
where the temperature $T$ controls how sharp the transition from inside to outside is, similar to the learnable sharpness in NeuS~\cite{wang2021neus}. As no eikonal constraint is enforced, $s$ is a level-set function rather than a true distance.
This allows the shell spacing to tighten locally, while temperature controls opacity.
Instead of performing volumetric integration along each view ray via ray marching, GTR-Mesh approximates this integral using a discrete set of shells, yielding a discrete, rasterizable representation that mimics the effect of volume rendering. Specifically, we extract $n$ shells at fixed isolevels $\tau_1 > \dots > \tau_n$ and alpha-composite them in depth order.

\paragraph{Marching tetrahedra.}
For every tetrahedron edge whose endpoint values $s_a$ and $s_b$ enclose a level $\tau$, we compute the linear interpolation weight $\lambda = (s_a - \tau)/(s_a - s_b)$ and place a shell vertex at the corresponding position.
The opacity of a shell fragment is computed from the interpolated SDF value
\begin{equation}
    \tilde{s} = (1 - \text{sg}(\lambda))\, s_a + \text{sg}(\lambda)\, s_b,
\end{equation}
where $\text{sg}(\cdot)$ denotes the stop-gradient operator.
In the forward pass, $\tilde{s}$ equals $\tau$, so all fragments of a shell share the opacity $\alpha(\tau)$.
Without the stop-gradient, $\tilde{s}$ remains the constant $\tau$ and its gradients with respect to $s_a$ and $s_b$ vanish.
With it, the photometric loss changes the SDF values at the tetrahedron vertices.
Similar stop-gradient formulations are used by Meshtryoshka~\cite{charatan2026meshtryoshka}, which equivalently interpolates the SDF at Marching Cubes vertices, and by Zhang~\etal~\cite{zhang2026meshsplatting}, who compute the signed distance of shell vertices to a base mesh.

\begin{figure}[ht]
    \centering
    \input{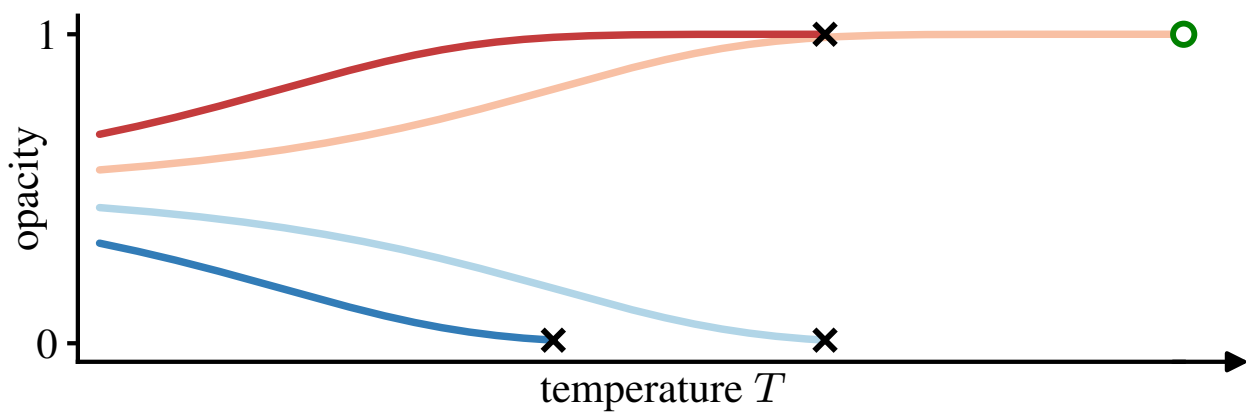}
    \caption{Opacity of the four shells for different temperatures $T$. Shells with an opacity below $0.01$ or hidden behind a shell with an opacity above $0.99$ are not rendered (black cross). For high temperatures, only one practically opaque shell remains. The green circle marks the shell that forms the final mesh.}
    \label{fig:suppl_shell_temperature}
\end{figure}

\paragraph{Shells and temperature schedule.}
We use an even number of shells placed symmetrically around the zero crossing so that 
no shell lies exactly on it.
Starting from a low temperature, where all shells are semi-transparent and together form a soft, volume-like representation, we gradually raise the temperature during training.
The outer shells become transparent and the inner ones opaque, until only a single opaque shell forms the final mesh (Fig.~\ref{fig:suppl_shell_temperature}).
For efficiency, we skip shells that are nearly transparent or hidden behind a nearly opaque shell.
The exact shell levels and the temperature schedule are given in the supplementary material.

\subsection{Color model}\label{sec:color_model}
Similar to 3DGS variants using hash grids to capture a primitive's appearance~\cite{NeST,Nexels,kelkar2026hybrid}, we split appearance into a view-independent part stored in a volumetric texture and a view-dependent part stored on the geometry.
The volumetric texture is a multi-resolution hash grid with a small MLP decoder~\cite{muller_instant_2022}, queried at the 3D position $\mathbf x$ of each rasterized fragment.
Because it is defined everywhere in space, the surface can move and change its topology during training without any texture parameterization; after training, it is baked into a standard UV texture (Sec.~\ref{sec:texture}).
For view-dependent effects, each point carries spherical harmonics coefficients, which are interpolated to the shell vertices and evaluated once per vertex, giving a color $\mathbf v$.

In contrast to previous work, our hash grid predicts a diffuse color $\mathbf b(\mathbf x)$ and a scalar strength $w(\mathbf x) \in [0,1]$, and the fragment color is
\begin{equation}
\mathbf c(\mathbf x) = \mathbf b(\mathbf x) + w(\mathbf x)\,\bar{\mathbf v},
\end{equation}
where $\bar{\mathbf v}$ is $\mathbf v$ interpolated within the triangle.
The strength $w$ is needed because $\mathbf v$ is only stored at the vertices and therefore varies smoothly across each triangle.
View-dependent effects such as specular highlights, however, are often confined to small regions, for example a glossy object next to a diffuse one within the same triangle.
Without $w$, the view-dependent color spreads over the whole triangle, and the mesh has to be subdivided to localize it.
The strength regularizes the separation of geometry and appearance, acting as a high-resolution mask that decides per fragment how much of the view-dependent color is added, while the view-dependent signal itself remains cheap to evaluate. 
After baking, $w$ is stored in the texture's alpha channel (Fig.~\ref{fig:atlas}), so that per fragment a single RGBA texture lookup and the interpolated per-vertex SH colors are required.

\subsection{Differentiable rendering of connected shells}\label{sec:rendering}
For optimizing multiple transparent shells, existing differentiable rasterizers suffer from four key problems. We address each with a dedicated design choice in a tile-based CUDA rasterizer (see additional  details in the supplementary material).

\paragraph{Shell handling.}
The widely used differentiable rasterizer \emph{nvdiffrast} resolves only the single nearest fragment per pixel (smallest depth) and cannot natively capture multiple overlapping transparent layers. Our shell-based approach, by contrast, requires many fragments per pixel to be alpha-composited. Depth peeling could in principle handle this, but it needs one rendering pass per layer and does not scale to large meshes. We therefore bin triangles into screen-space tiles and depth-sort all fragments of each pixel into a single per-pixel list, resolving all layers in one pass.

\paragraph{Inaccurate coverage.}
Geometry gradients arise from how triangle edges move across pixels. During training, many triangles shrink below a pixel, so these gradients are informative only if subpixel coverage is computed accurately. \emph{nvdiffrast} approximates anti-aliasing as a post-process on silhouette edges, and soft rasterizers~\cite{liu2019softras} blur triangles rather than compute their coverage; neither yields accurate gradients for subpixel triangles. Our rasterizer instead computes, for each triangle, the exact fraction $\kappa_i \in [0,1]$ 
of pixel area it covers---yielding exact coverage gradients even for subpixel triangles. Overlap grouping is approximated using a subpixel  occupancy mask, as described below.

\paragraph{Seams at shared edges.}
Simply alpha-blending each fragment with opacity $\kappa_i\alpha_i$ treats adjacent triangles of the same shell as overlapping. For example, two triangles that each cover half a pixel with $\alpha = 1$ yield a combined opacity of only $0.75$, so shared edges turn semi-transparent. Soft rasterizers~\cite{liu2019softras} have this problem because they blend triangles independently, and \emph{nvdiffrast}~\cite{Laine2020diffrast} avoids it only by not weighting fragments by coverage.

\begin{figure}[tbp]
  \centering
  \includegraphics[width=\linewidth]{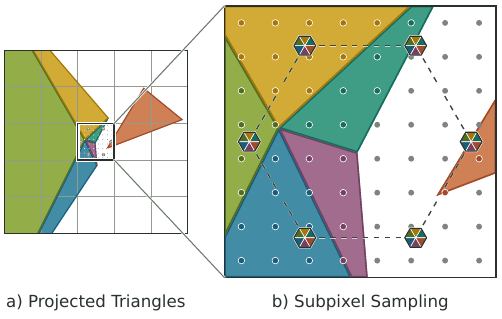}
  \caption{Anti-aliased rendering. (a) Projected triangles. (b) Within a pixel, each triangle's coverage is computed analytically, and an $m\times m$ subpixel mask (dots) groups non-overlapping triangles into layers that are accumulated additively; a triangle hitting an occupied subpixel starts a new layer. The hexagons mark the hash-grid sample positions for prefiltered appearance.}
  \label{fig:aa_render}
\end{figure}

We avoid it by processing a pixel's fragments front to back and marking the subpixels each fragment covers in an $m\times m$ occupancy mask (Fig.~\ref{fig:aa_render}), with $m=8$, similar to the A-buffer~\cite{carpenter1984abuffer}.
As long as a fragment covers only free subpixels, we treat it as non-overlapping and lying beside---not behind---the previous fragments, so we simply add its color and opacity to the current layer. Once a fragment covers an occupied subpixel, it lies behind an earlier fragment; we then alpha-composite the current layer onto the pixel, clear the mask, and start a new layer with this fragment.

\paragraph{Prefiltered hash grid lookup.}
Because a fragment also integrates color over the surface area it covers, anti-aliased coverage alone is not sufficient. A pixel can cover a large surface area on distant or grazing geometry, and a single hash grid lookup at the fragment center then samples one point of a high-frequency signal, causing aliasing and erratic changes when the camera moves. We therefore adapt the prefiltering of Zip-NeRF~\cite{barron2023zipnerf} from rays to surfaces. Zip-NeRF integrates the hash grid over a conical frustum along each ray. We instead place six samples on a hexagon around the pixel center and project them onto the triangle plane, so that all samples lie on the surface. To also account for grazing angles, the footprint size is derived from the screen-to-surface Jacobian, rather than from the cone radius as in Zip-NeRF.

\subsection{Gradient-based densification and pruning}\label{sec:densification}
Following the adaptive density control of 3DGS~\cite{kerbl_3d_2023}, we refine the point cloud where resolution is insufficient. Our split criterion, however, uses the gradient of the loss with respect to the SDF values instead of view-space positional gradients. A persistently large SDF gradient indicates that the geometry cannot satisfy the training views at the current resolution; our fix is more geometric degrees of freedom, not more color capacity. This decouples geometric refinement from texture refinement: a highly textured but geometrically simple region will not be densified, and a smooth but geometrically under-resolved region will be. 

For each point, we accumulate the magnitude of the SDF gradient over the training views and average it over the views in which the point is visible. Because magnitudes do not cancel, the score stays large even where views pull the SDF in conflicting directions and the net gradient is small. We sum the score over the four corners of each tetrahedron and insert a new point inside every tetrahedron whose score exceeds a threshold, interpolating its SDF value so the surface is approximately preserved. Conversely, we periodically remove points that do not contribute to any rendered shell.

\subsection{Optimization}\label{sec:optimization}

\paragraph{Losses.}
Following 3DGS~\cite{kerbl_3d_2023}, we minimize the photometric loss
\begin{equation}
    \mathcal{L} = (1 - \lambda_{\text{SSIM}})\,\mathcal{L}_1 + \lambda_{\text{SSIM}}\,(1 - \text{SSIM}),
\end{equation}
with $\lambda_{\text{SSIM}} = 0.2$.
In addition, we regularize the hash-grid features with a small L2 penalty (weight $0.001$) following Zip-NeRF~\cite{barron2023zipnerf}.

\paragraph{Depth of field.}
Some training images contain noticeable defocus blur.
Our final representation consists of opaque triangles with sharp boundaries and cannot reproduce such blur.
Without an explicit model, the optimization would try to explain it with blurred textures or additional semi-transparent geometry.
We therefore model defocus during training following DoF-Gaussian~\cite{shen2025dofgaussiancontrollabledepthoffield3d}: the rendered image is blurred with a depth-dependent circle of confusion, using a learned aperture and automatic focus estimation.
The blur is only used during training.

\paragraph{Training.}
During training, the temperature is raised (Sec.~\ref{sec:shells}) while points are densified and pruned (Sec.~\ref{sec:densification}), and the tetrahedralization is recomputed every 10 steps.
After training, the remaining opaque shell is extracted as the final mesh and its appearance is baked into a texture (Sec.~\ref{sec:texture}).
All hyperparameters are listed in the supplementary material.

\subsection{Texture export}\label{sec:texture}
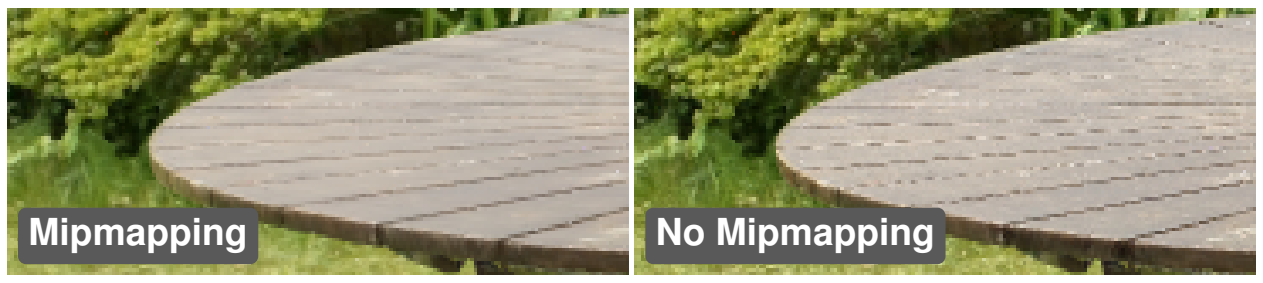
\begin{figure}[htbp]
  \centering
  \newsavebox{\aaimgbox}
  \newlength{\aaW}
  \newlength{\aaClipH}
  \newlength{\aaH}
  \setlength{\aaW}{\dimexpr(\linewidth-3pt)/2\relax}
  \setlength{\aaH}{0.25\linewidth}

  \newcommand{\aaimg}[3][0]{%
  \setlength{\aaClipH}{\aaH}%
    \ifnum#1=1\relax
      \setlength{\aaClipH}{0.85\aaH}%
    \fi
    \ifnum#1=1\relax
      \sbox{\aaimgbox}{%
        \includegraphics[width=\dimexpr2\aaW\relax]{#2}}%
      \ifdim\ht\aaimgbox<\dimexpr2\aaH\relax
        \sbox{\aaimgbox}{%
          \includegraphics[height=\dimexpr2\aaH\relax]{#2}}%
      \fi
    \else
      \sbox{\aaimgbox}{\includegraphics[width=\aaW]{#2}}%
      \ifdim\ht\aaimgbox<\aaH
        \sbox{\aaimgbox}{\includegraphics[height=\aaH]{#2}}%
      \fi
    \fi
    \begin{tikzpicture}[baseline=(current bounding box.south)]
      \path[use as bounding box] (0,0) rectangle (\aaW,\aaClipH);
      \begin{scope}
        \clip (0,0) rectangle (\aaW,\aaClipH);
        \ifnum#1=1\relax
          \node[inner sep=0, anchor=south west]
            at (0,0) {\usebox{\aaimgbox}};
        \else
          \node[inner sep=0]
            at (0.5\aaW,0.5\aaH) {\usebox{\aaimgbox}};
        \fi
      \end{scope}
      \node[
        fill=black!65,
        rounded corners=1.5pt,
        inner sep=2pt,
        text=white,
        font=\sffamily\bfseries\scriptsize,
        anchor=south west
      ] at (2pt,2pt) {#3};
    \end{tikzpicture}%
  }

  \makebox[\linewidth][c]{%
    \vbox{%
        \aaimg[1]{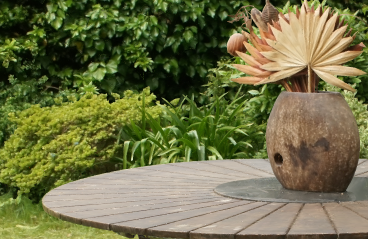}{Mipmapping}%
        \hskip1pt
        \aaimg[1]{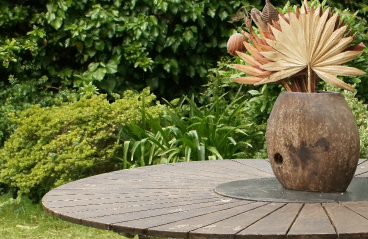}{No Mipmapping}%
      }%
    }%

  \caption{Results w/ and w/o mipmapping in rendering.}
  \label{fig:aa}
\end{figure}

\begin{figure}[tbp]
    \centering
    \includegraphics[width=0.40\linewidth]{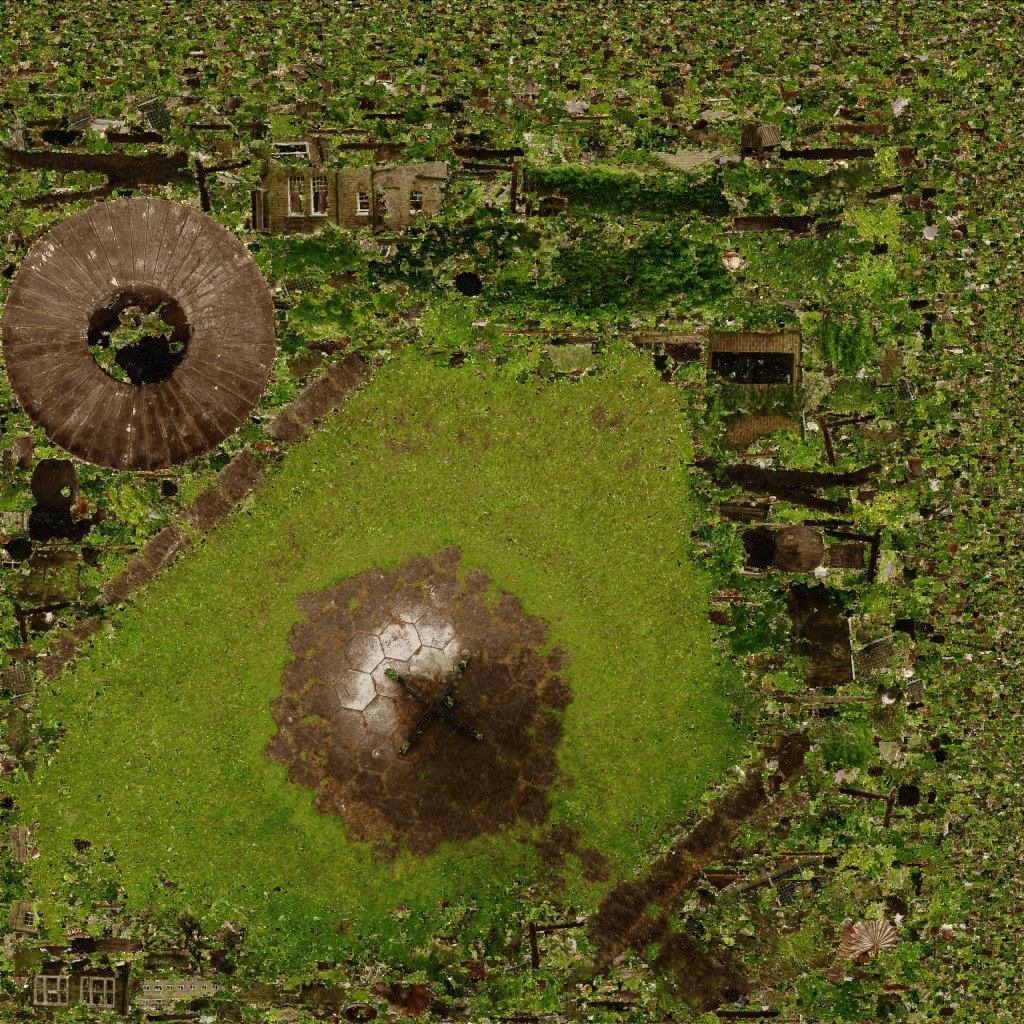}\hspace{0.03\linewidth}%
    \includegraphics[width=0.40\linewidth]{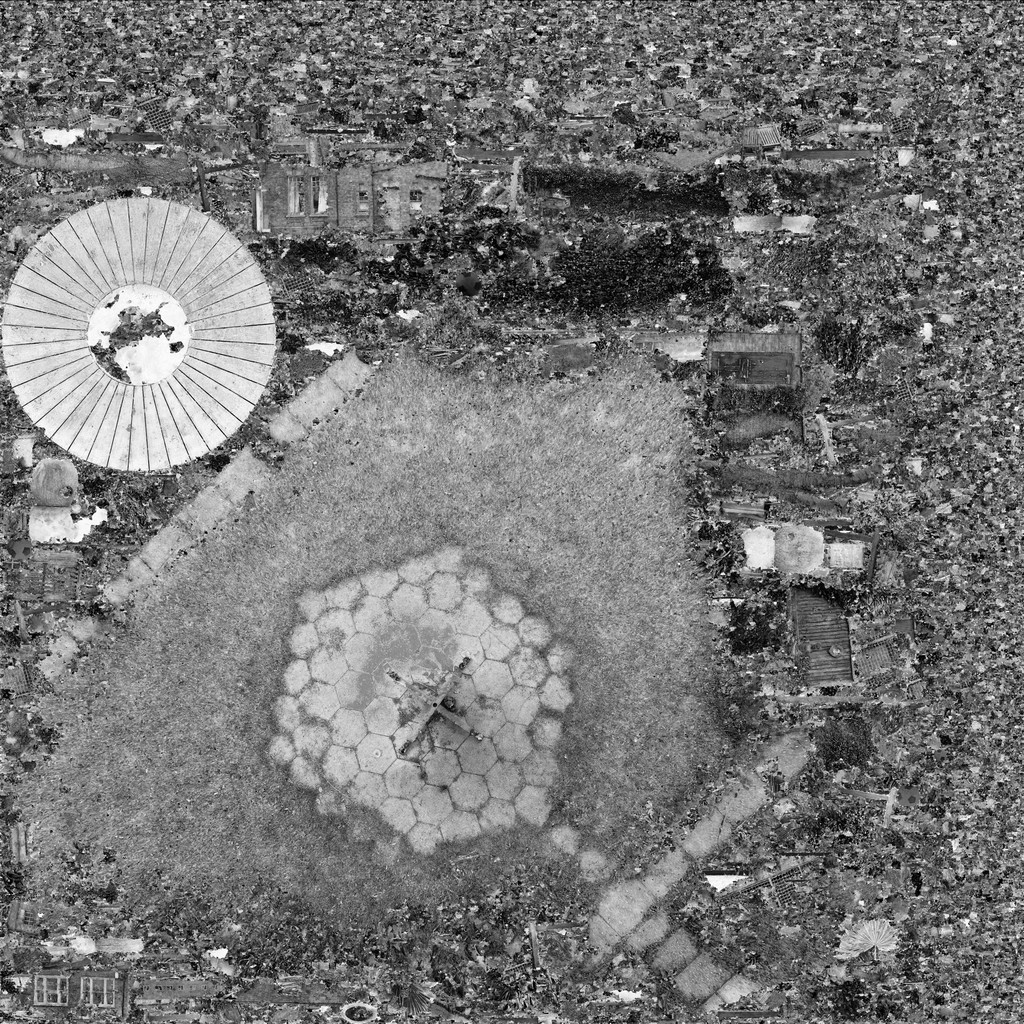}
    \caption{Baked texture atlas of \textit{garden}. Left: diffuse color $\mathbf b$. Right: strength $w$, stored in the alpha channel, which scales the view-dependent vertex color.}
    \label{fig:atlas}
\end{figure}

After training, we extract the remaining shell with MT and export it as a standard glTF~\cite{iso2022gltf} mesh. The SH coefficients are stored as vertex attributes, and the hash grid is baked into a single $8K$ RGBA texture atlas that holds the diffuse color $\mathbf b$ and the strength $w$ (Fig.~\ref{fig:atlas}).
We unwrap the mesh with Blender's Smart UV Project~\cite{blender}, using a standalone implementation~\cite{smartUV}, and pack the UV islands into the atlas with xatlas~\cite{young2018xatlas}.
To spend texels where they are visible, we compute for each triangle its largest screen-space size over all training views and scale its UV island accordingly. 
Large, contiguous UV islands facilitate GPU mipmapping, which reduces filtering artifacts and aliasing problems (Fig.~\ref{fig:aa}).

We initialize the texture by querying the hash grid at the world position of each texel and then fine-tune the texture and the SH coefficients for 1,500 steps on the training views.
All reported metrics are computed on this textured mesh; further details are in the supplementary material.

\begin{table*}[tbp]
\centering

\caption{Quantitative comparison on Mip-NeRF 360~\cite{barron2022mipnerf360} and Tanks \& Temples~\cite{knapitsch_tanks_2017}. We report average PSNR, LPIPS, and SSIM, and the number of vertices $|V|$ and faces $|F|$, over all scenes. Baseline numbers are for the rendered meshes, taken from the respective publications or, for MeshSplatting and Blobs to Spokes on Tanks \& Temples, computed with the published code (see suppl.). $^\dagger$Not watertight due to visibility culling.}
\begingroup
\setlength{\tabcolsep}{3pt}
\resizebox{\textwidth}{!}{%
\begin{tabular}{@{}l|cc|ccccc|ccccc@{}}
\hline
\multirow{2}{*}{Method} & \multirow{2}{*}{2D Textures} & Watertight, no
& \multicolumn{5}{c|}{Mip-NeRF 360} & \multicolumn{5}{c}{Tanks \& Temples} \\
\cline{4-13}
 & & self-intersections
& PSNR$\uparrow$ & LPIPS$\downarrow$ & SSIM$\uparrow$ & $|V|\downarrow$ & $|F|\downarrow$
& PSNR$\uparrow$ & LPIPS$\downarrow$ & SSIM$\uparrow$ & $|V|\downarrow$ & $|F|\downarrow$ \\
\hline

\rowcolor{gray!15}3DGS~\cite{kerbl_3d_2023}  & - & -
& 27.21 & 0.214 & 0.815 & 3.6M & - & 23.14 & 0.183 & 0.841 & 1.8M & - \\
\hline
MeshSplatting~\cite{held2026mesh}  & \xmark & \xmark
& 24.87 & 0.301 & 0.741 & \first{2.8M} & \first{5.3M} & \third{20.57} & \second{0.271} & \second{0.760} & \first{2M} & 3.9M \\
Binary Opacity Grids~\cite{reiser2024binary}  & \xmark & \xmark$^\dagger$
& \second{25.62} & \second{0.247} & \second{0.766} & - & \third{13M} & - & - & - & - & - \\
\hline
MILo~\cite{guedon2025milo}  & \xmark & \cmark
& 24.09 & 0.323 & 0.688 & \third{7M} & 15M & \second{21.46} & 0.348 & \third{0.706} & \third{4M} & 9M \\
Meshtryoshka~\cite{charatan2026meshtryoshka}  & \xmark & \cmark
& 24.30 & 0.369 & 0.684 & - & - & - & - & - & - & - \\
Blobs to Spokes~\cite{gomez2026gaussianwrapping}  & \xmark & \cmark
& \third{25.10} & \third{0.289} & \third{0.752} & 11.2M & 22.1M & 18.34 & \third{0.331} & 0.677 & 6.1M & 12.1M \\
\hline
\textbf{GTR-Mesh}  & \cmark & \cmark
& \first{25.88} & \first{0.216} & \first{0.789} & \second{3.0M} & \second{6.3M} & \first{21.91} & \first{0.167} & \first{0.813} & \second{2.9M} & 6.0M \\
\hline
\end{tabular}}
\endgroup
\label{tab:main}
\end{table*}

\section{Implementation}
Our method is implemented in PyTorch~\cite{paszke2017automatic} with custom CUDA kernels for the rasterizer and Marching Tetrahedra; the Delaunay tetrahedralization uses geogram~\cite{Levy2015geogram}.
We train for 6,000 steps on an NVIDIA RTX 5090.
All metrics are computed on the final representation, i.e., the exported UV-textured mesh with per-vertex view-dependent colors.
Following common practice~\cite{kerbl_3d_2023}, we downscale the Mip-NeRF 360 images by a factor of 4 for outdoor and 2 for indoor scenes.

\section{Experiments}
We evaluate on all nine scenes of Mip-NeRF 360~\cite{barron2022mipnerf360} and on the \textit{train} and \textit{truck} scenes of Tanks \& Temples~\cite{knapitsch_tanks_2017}.
Following 3DGS~\cite{kerbl_3d_2023}, we use every 8th image for testing. We report PSNR, LPIPS (VGG)~\cite{zhang2018unreasonable}, and SSIM.
To evaluate geometric accuracy, we also report the Chamfer distance for the 15 DTU scans~\cite{jensen2014large} commonly used in surface reconstruction, following the standard evaluation protocol.
Camera parameters are taken from COLMAP~\cite{schoenberger2016sfm}, and all scenes use the same hyperparameters.
On Mip-NeRF 360 and Tanks \& Temples, we compare against methods that output opaque meshes: Binary Opacity Grids~\cite{reiser2024binary}, MILo~\cite{guedon2025milo}, MeshSplatting~\cite{held2026mesh}, Meshtryoshka~\cite{charatan2026meshtryoshka}, and Blobs to Spokes~\cite{gomez2026gaussianwrapping}. Binary Opacity Grids and Meshtryoshka report results on Mip-NeRF 360 only; for Meshtryoshka we quote the published zero-level-set results. On DTU, we additionally compare against the surface reconstruction methods listed in Tab.~\ref{tab:dtu}. We also report 3DGS~\cite{kerbl_3d_2023} as a non-mesh reference.

\subsection{Results}
\begin{figure*}[p]
\centering
\includegraphics[width=\textwidth]{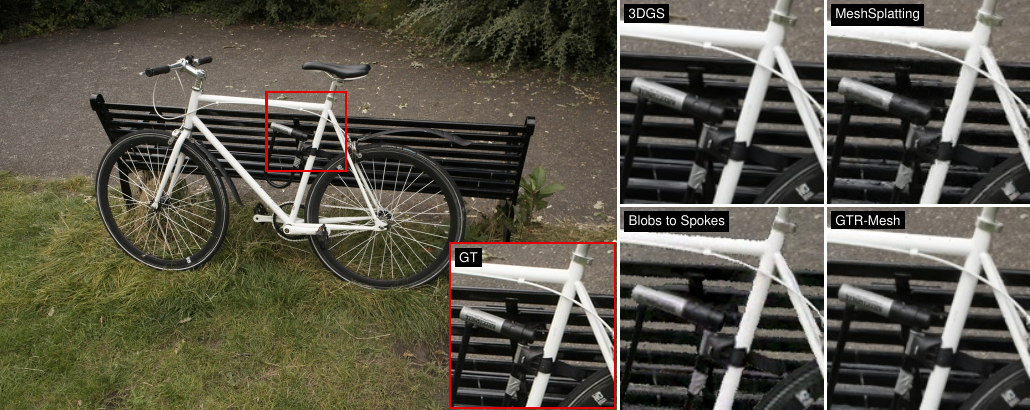}\\[0.6ex]
\includegraphics[width=\textwidth]{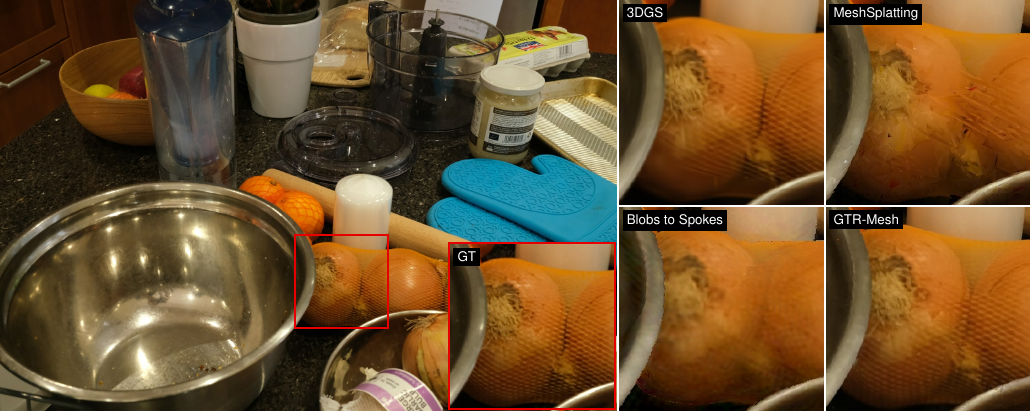}\\[0.6ex]
\includegraphics[width=\textwidth]{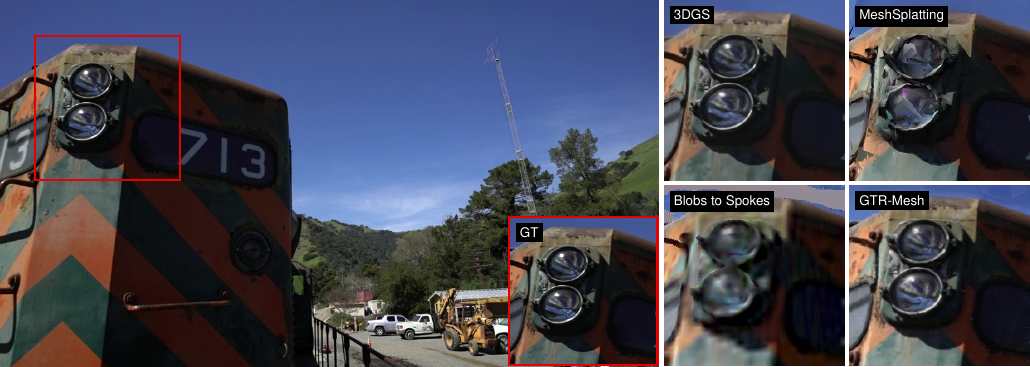}
\caption{Qualitative comparison on test views of \textit{bicycle}, \textit{counter} (Mip-NeRF 360) and \textit{train} (Tanks \& Temples). 
}
\label{fig:qualitative}
\end{figure*}

Tab.~\ref{tab:main} and Fig.~\ref{fig:qualitative} show  quantitative and qualitative results of GTR-Mesh, respectively.
On Mip-NeRF 360, GTR-Mesh outperforms all mesh-based methods in PSNR, SSIM, and LPIPS, and its LPIPS ($0.216$) is on par with 3DGS ($0.214$). On Tanks \& Temples, it achieves a lower LPIPS than 3DGS ($0.167$ vs.\ $0.183$).
Compared to MILo and Blobs to Spokes, the other watertight methods that report mesh sizes, GTR-Mesh uses $2.3\times$ and $3.7\times$ fewer vertices on Mip-NeRF 360, since fine appearance detail is stored in the texture rather than in additional geometry.
MeshSplatting~\cite{held2026mesh} uses fewer vertices, but its meshes are not watertight and contain overlapping triangles that cause artifacts in novel views.
Rendering timings of the final meshes are given in the supplementary material.

\paragraph{Geometric accuracy.}
Tab.~\ref{tab:dtu} reports the Chamfer distance on DTU in millimeters.
Although GTR-Mesh is trained with a photometric loss only and uses no geometric regularization, it is more accurate than NeuS~\cite{wang2021neus}, 2DGS~\cite{huang20242d}, GOF~\cite{yu2024gof}, MILo, and MeshSplatting, matches Zhang~\etal~\cite{zhang2026meshsplatting}, and is on par with Neuralangelo~\cite{li2023neuralangelo}. PGSR~\cite{chen2024pgsr} and Blobs to Spokes~\cite{gomez2026gaussianwrapping} remain more accurate on average; per-scan results are given in the supplementary material.
The largest differences occur on scans 65 and 110, which contain strong view-dependent effects.
Since we do not model such effects correctly as surface reflectance, the optimization partly explains them by displacing the surface.
Geometric regularizers or a more expressive view-dependent appearance model could address this, which we leave for future work.

\begin{table}[tbp]
\centering
\caption{Average Chamfer distance (CD) over the 15 scans from DTU~\cite{jensen2014large}. Numbers are taken from the respective publications.}
\small
\begin{tabular}{@{}lclc@{}}
\hline
Method & CD $\downarrow$ & Method & CD$\downarrow$ \\
\hline
Neuralangelo~\cite{li2023neuralangelo} & \third{0.61} &
NeuS~\cite{wang2021neus} & 0.84 \\
\hline
2DGS~\cite{huang20242d} & 0.80 &
GOF~\cite{yu2024gof} & 0.74 \\
Blobs to Spokes~\cite{gomez2026gaussianwrapping} & \first{0.49} &
MILo~\cite{guedon2025milo} & 0.68 \\
PGSR~\cite{chen2024pgsr} & \second{0.52} & & \\
\hline
Zhang~\etal~\cite{zhang2026meshsplatting} & 0.62 &
MeshSplatting~\cite{held2026mesh} & 0.79 \\
\hline
\textbf{GTR-Mesh} & 0.62 & & \\
\hline
\end{tabular}
\label{tab:dtu}
\end{table}

\begin{figure}[htbp]
  \centering
  \setlength{\aaW}{\dimexpr(\linewidth-3pt)/2\relax}
  \setlength{\aaH}{0.25\linewidth}

  \newcommand{\aaimg}[3][0]{%
  \setlength{\aaClipH}{\aaH}%
    \ifnum#1=1\relax
      \setlength{\aaClipH}{0.85\aaH}%
    \fi
    \ifnum#1=1\relax
      \sbox{\aaimgbox}{%
        \includegraphics[width=\dimexpr2\aaW\relax]{#2}}%
      \ifdim\ht\aaimgbox<\dimexpr2\aaH\relax
        \sbox{\aaimgbox}{%
          \includegraphics[height=\dimexpr2\aaH\relax]{#2}}%
      \fi
    \else
      \sbox{\aaimgbox}{\includegraphics[width=\aaW]{#2}}%
      \ifdim\ht\aaimgbox<\aaH
        \sbox{\aaimgbox}{\includegraphics[height=\aaH]{#2}}%
      \fi
    \fi
    \begin{tikzpicture}[baseline=(current bounding box.south)]
      \path[use as bounding box] (0,0) rectangle (\aaW,\aaClipH);
      \begin{scope}
        \clip (0,0) rectangle (\aaW,\aaClipH);
        \ifnum#1=1\relax
          \node[inner sep=0, anchor=south west]
            at (0,0) {\usebox{\aaimgbox}};
        \else
          \node[inner sep=0]
            at (0.5\aaW,0.5\aaH) {\usebox{\aaimgbox}};
        \fi
      \end{scope}
      \node[
        fill=black!65,
        rounded corners=1.5pt,
        inner sep=2pt,
        text=white,
        font=\sffamily\bfseries\scriptsize,
        anchor=south west
      ] at (2pt,2pt) {#3};
    \end{tikzpicture}%
  }

  \makebox[\linewidth][c]{%
    \vbox{%
      \hbox{%
        \aaimg{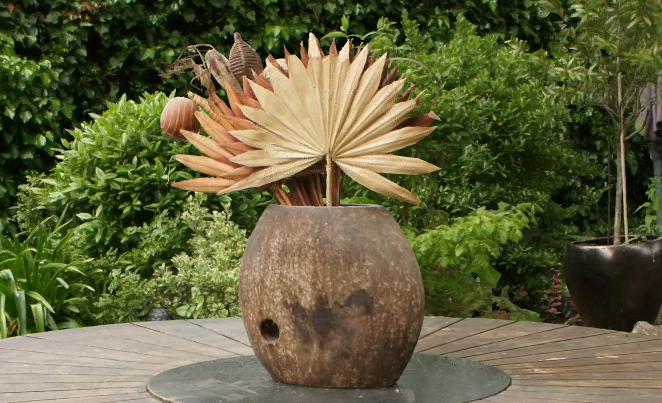}{Anti-aliasing}%
        \hskip1pt
        \aaimg{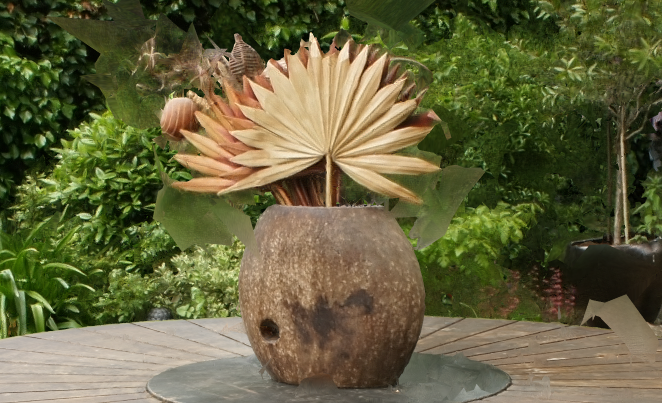}{No Anti-aliasing}%
      }%
    }%
  }
  \caption{Reconstructions w/ and w/o anti-aliasing in training.}
  \label{fig:bb}
\end{figure}
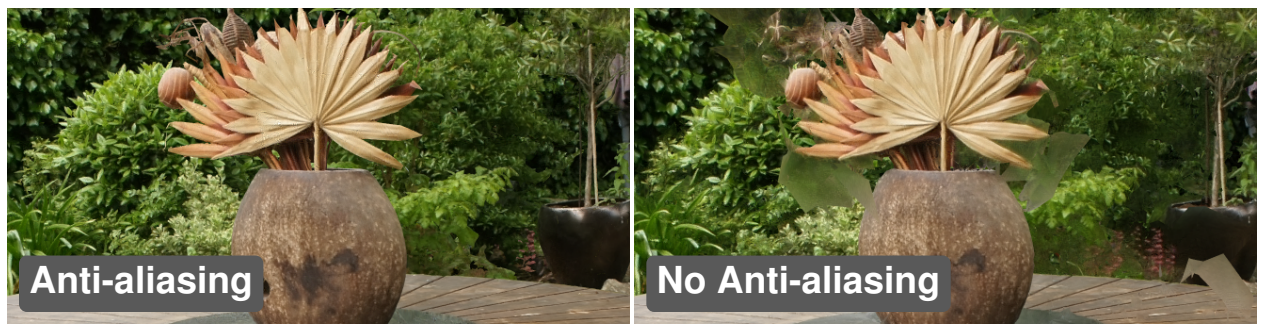

\subsection{Ablations}
Tab.~\ref{tab:ablation} reports ablations averaged over all nine Mip-NeRF 360 scenes.
Anti-aliasing has the largest impact: without it, silhouette edges of fine structures are deformed (Fig.~\ref{fig:bb}), PSNR drops by $4.4$\,dB, and training can diverge in regions with fine structures.
While the shells let the surface move towards the target geometry early in training, anti-aliasing is needed to refine fine details once only a nearly opaque surface remains.
Removing densification reduces PSNR by $0.9$\,dB, removing the prefiltered hash grid lookup by $0.5$\,dB, and training with a single shell from the start by $0.6$\,dB. Removing depth-of-field modeling costs only $0.05$\,dB but decreases training stability: two of nine scenes then converge with visible artifacts.
Starting with two instead of four shells reduces PSNR by $0.3$\,dB, while six shells give no measurable improvement but increase training time by about $19\%$; we therefore use four shells.
Baking the hash grid into the UV texture changes PSNR by only $0.05$\,dB, and LPIPS is slightly lower after baking ($0.216$ vs.\ $0.220$).

\begin{table}[tbp]
    \centering
    \caption{Losses/gains in quality/training time on Mip-NeRF 360. Training time includes texture unwrapping and fine-tuning.} 
    \resizebox{\linewidth}{!}{%
    \begin{tabular}{l|cccc}
         & PSNR$\uparrow$ & $\Delta$PSNR & LPIPS$\downarrow$ & Time [min]$\downarrow$ \\
        \hline
        Full & 25.88 & & 0.216 & 63.0 \\
        \hline
        $-$Anti-Aliasing & 21.47 & $-4.41$ & 0.344 & 50.4 \\
        $-$Densification & 24.97 & $-0.91$ & 0.248 & 49.1 \\
        $-$DoF & 25.83 & $-0.05$ & 0.215 & 63.4 \\
        $-$Prefiltered lookup & 25.39 & $-0.49$ & 0.238 & 48.7 \\
        \hline
        Single shell & 25.25 & $-0.63$ & 0.237 & 50.1 \\
        2 initial shells & 25.59 & $-0.29$ & 0.228 & 60.3 \\
        6 initial shells & 25.92 & $+0.04$ & 0.216 & 74.9 \\
        \hline
        $-$Baking & 25.93 & $+0.05$ & 0.220 & 53.0 \\
    \end{tabular}}
    \label{tab:ablation}
\end{table}

\section{Conclusion and limitations}
To our knowledge, GTR-Mesh is the first method that optimizes a watertight, textured triangle mesh end-to-end with a photometric loss on large unbounded scenes.
It produces a more compact shared-vertex mesh than other watertight methods, and can be rendered efficiently using hardware rasterization. 

Since GTR-Mesh is optimized only for photometric quality, the resulting triangulation can be irregular, with sliver triangles in regions of complex curvature (Fig.~\ref{fig:mesh}). Mesh-quality regularizers such as TetWeave~\cite{binninger2025tetweave} or standard post-processing remeshing could address this.
As an opaque surface representation, GTR-Mesh cannot represent semi-transparent or volumetric materials, and subpixel structures such as thin grass blades are harder to reproduce with opaque triangles than with semi-transparent Gaussians. This likely contributes to the remaining gap to 3DGS.
Appearance is optimized in a volumetric hash grid, which can slow convergence due to hash collisions and requires baking into a texture after training. Optimizing the texture directly would avoid both issues and reduce overall processing time.
Almost a third of the reconstruction time is spent on the repeated Delaunay tetrahedralization, which runs on the CPU because the available GPU implementations were not robust enough for our point clouds. In the future, we will consider fast progressive re-tetrahedralization to significantly accelerate training times.
Finally, learning material parameters instead of baked radiance can facilitate relighting and physically based rendering.

\beforereferences

{\small
\bibliographystyle{ieeenat_fullname}
\bibliography{main}
}

\begingroup
\renewcommand{\bibliographystyle}[1]{}
\renewcommand{\bibliography}[1]{}

\maketitlesupplementary

\appendix

\noindent This supplementary material is organized as follows.
\begin{itemize}
    \item[\ref{sec:suppl_impl}] Implementation details: hyperparameters, differentiable rasterizer, and texture export.
    \item[\ref{sec:suppl_shell_methods}] Relation to Meshtryoshka and Mesh Splatting by Zhang~\etal.
    \item[\ref{sec:suppl_eval}] Evaluation protocol and baseline numbers.
    \item[\ref{sec:suppl_quant}] Additional quantitative results: per-scene results and DTU per-scan geometry.
    \item[\ref{sec:suppl_validity}] Mesh validity: watertightness, vertex-manifoldness, and self-intersections.
    \item[\ref{sec:suppl_runtime}] Runtime, memory, and storage, including the comparison with \texttt{nvdiffrast}.
    \item[\ref{sec:suppl_ablations}] Additional ablations.
    \item[\ref{sec:suppl_visualizations}] Additional qualitative results.
    \item[\ref{sec:suppl_applications}] Downstream applications.
\end{itemize}

\section{Implementation Details}
\label{sec:suppl_impl}
\subsection{Hyperparameters}
\label{sec:suppl_hyperparameters}

\paragraph{Shell levels and temperature schedule.}
The opacity of a shell at SDF level $\tau$ is $\alpha = \sigma(-T\tau)$ with temperature $T$, as in the main paper. The outermost shells are placed at the SDF levels where the opacity at the initial temperature $T=4$ equals a minimum of $0.05$ and a maximum of $0.95$, and the remaining shells are spaced linearly in SDF between them. For four shells, this results in initial opacities of $0.05$, $0.27$, $0.73$, and $0.95$.

The temperature is kept at $T = 4$ until iteration 2,000, raised linearly to $T = 40$ at iteration 2,450, after which it stays constant.
Shells with an opacity below $0.01$, or behind a shell with an opacity above $0.99$, are not rendered.

From iteration 2,185 on, only one shell is rendered, which forms the final mesh; the full schedule is shown in Fig.~\ref{fig:suppl_temperature_schedule}. With four shells, the levels are $\tau \approx \pm 0.736$ and $\pm 0.245$, and the remaining opaque shell is the one at $\tau \approx -0.245$, not the zero level set. Since $s$ is not a distance (Sec.~3.2 of the main paper), this offset has no geometric meaning; the optimization places this level set on the observed surface.
We observe that the shells converge quickly to the scene surfaces, within about 500 iterations (Fig.~\ref{fig:suppl_training_progress}); the temperature ramp-up merely provides a smoother transition to a single opaque shell.
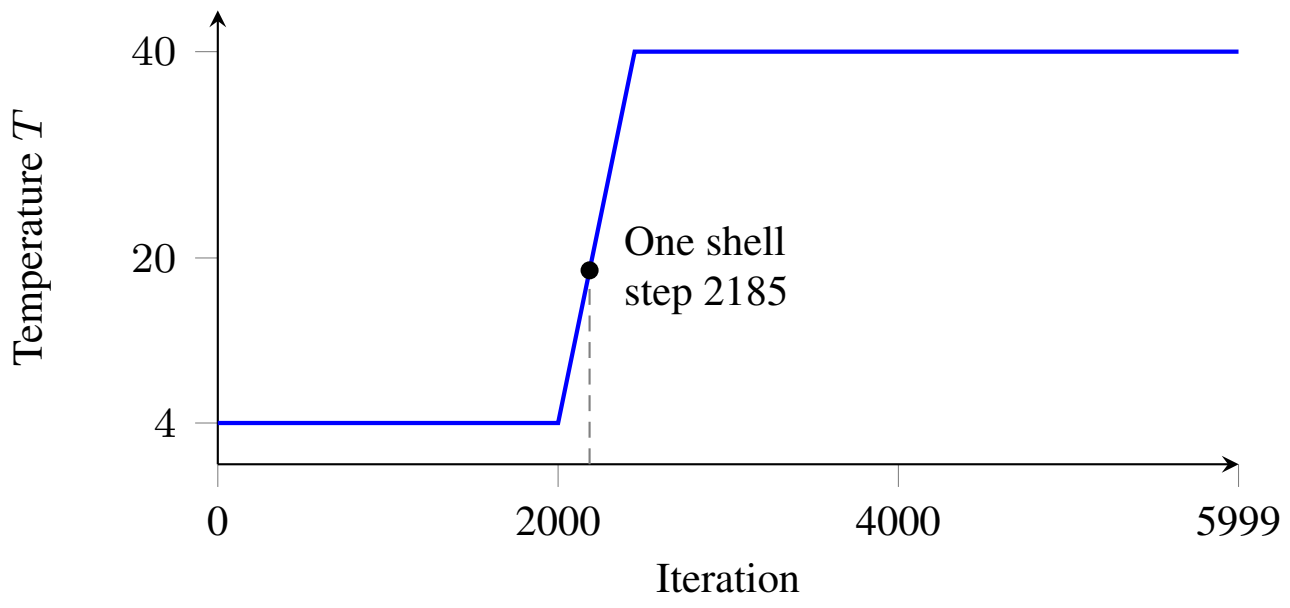
\begin{figure}[ht]
    \centering
    \begin{tikzpicture}
\begin{axis}[
  width=\linewidth, height=0.55\linewidth,
  xmin=0, xmax=6000, ymin=0, ymax=44,
  xlabel={Iteration}, ylabel={Temperature $T$},
  xtick={0,2000,4000,5999}, xticklabels={0,2000,4000,5999},
  ytick={4,20,40}, scaled ticks=false,
  axis lines=left, tick align=outside,
  tick label style={font=\footnotesize}, label style={font=\footnotesize},
]
  \addplot[blue, thick, no marks] coordinates {
    (0,4) (2000,4) (2450,40) (5999,40)
  };
  \draw[gray, densely dashed] (axis cs:2185,0) -- (axis cs:2185,18.8);
  \fill (axis cs:2185,18.8) circle[radius=1.7pt];
  \node[anchor=west, align=left, font=\footnotesize, xshift=3pt]
    at (axis cs:2185,18.8) {One shell\\step 2185};
\end{axis}
\end{tikzpicture}
    \caption{Temperature schedule. From iteration 2,185 on, only one shell is rendered.}
    \label{fig:suppl_temperature_schedule}
\end{figure}
\begin{figure*}[tbp]
    \centering
    \setlength{\tabcolsep}{1pt}
    \begin{tabular}{@{}ccc@{}}
        \includegraphics[width=0.32\textwidth]{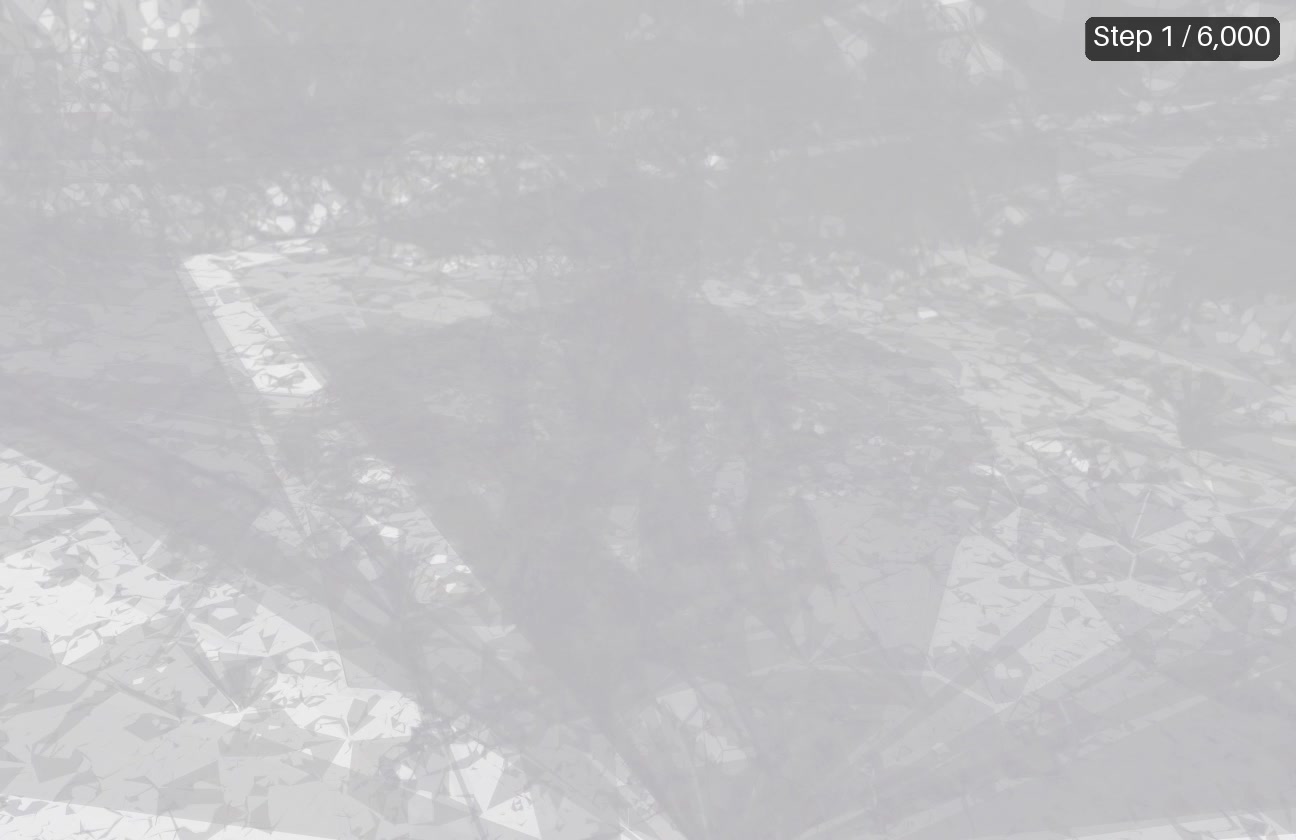} &
        \includegraphics[width=0.32\textwidth]{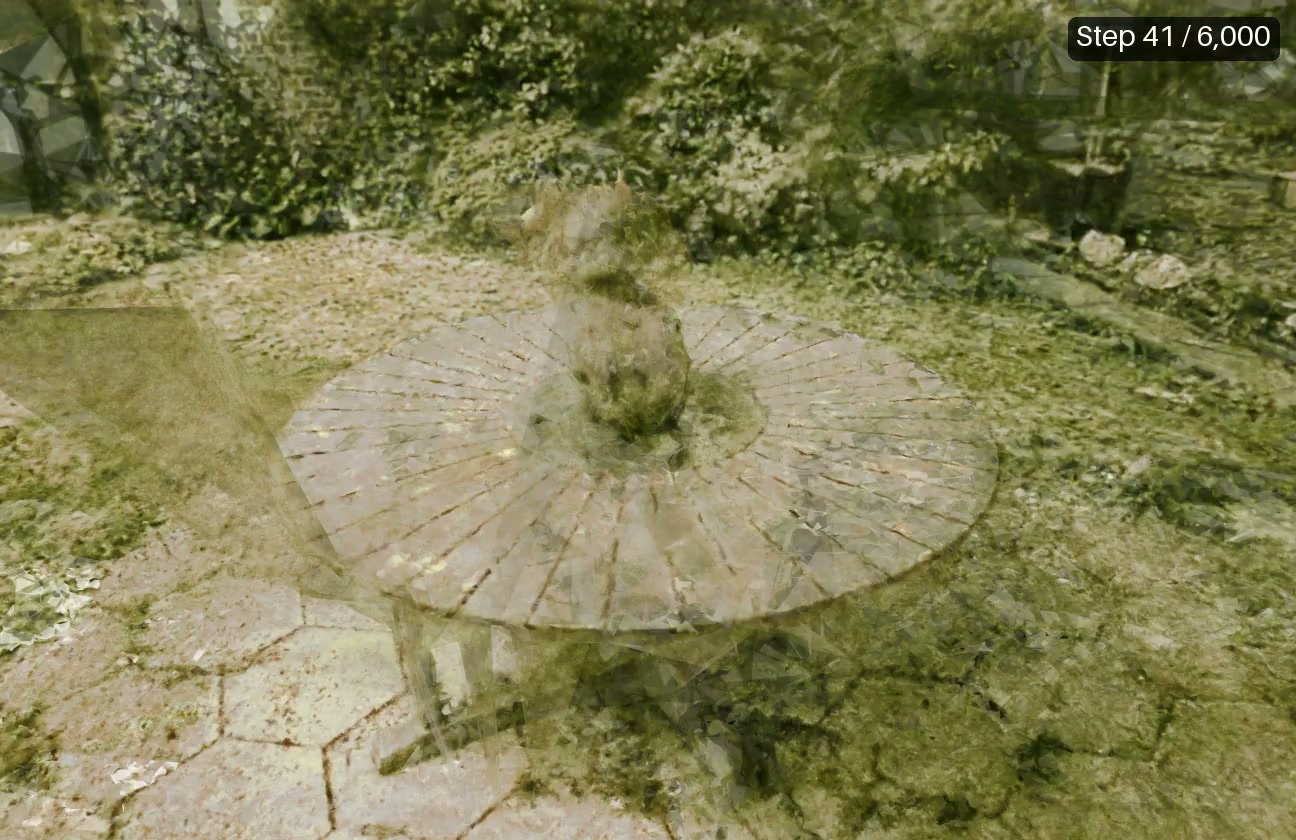} &
        \includegraphics[width=0.32\textwidth]{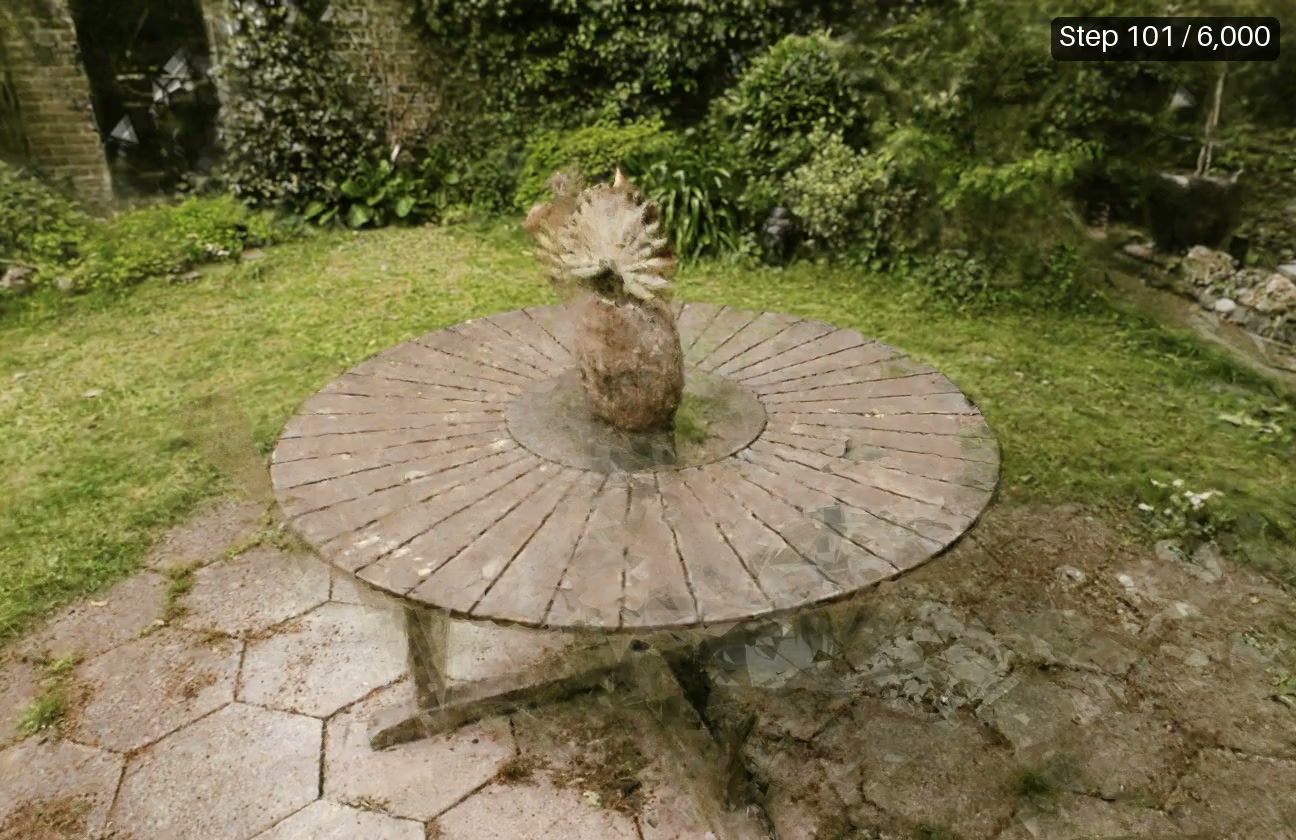} \\
        \small Step 1 & \small Step 41 & \small Step 101 \\[3pt]
        \includegraphics[width=0.32\textwidth]{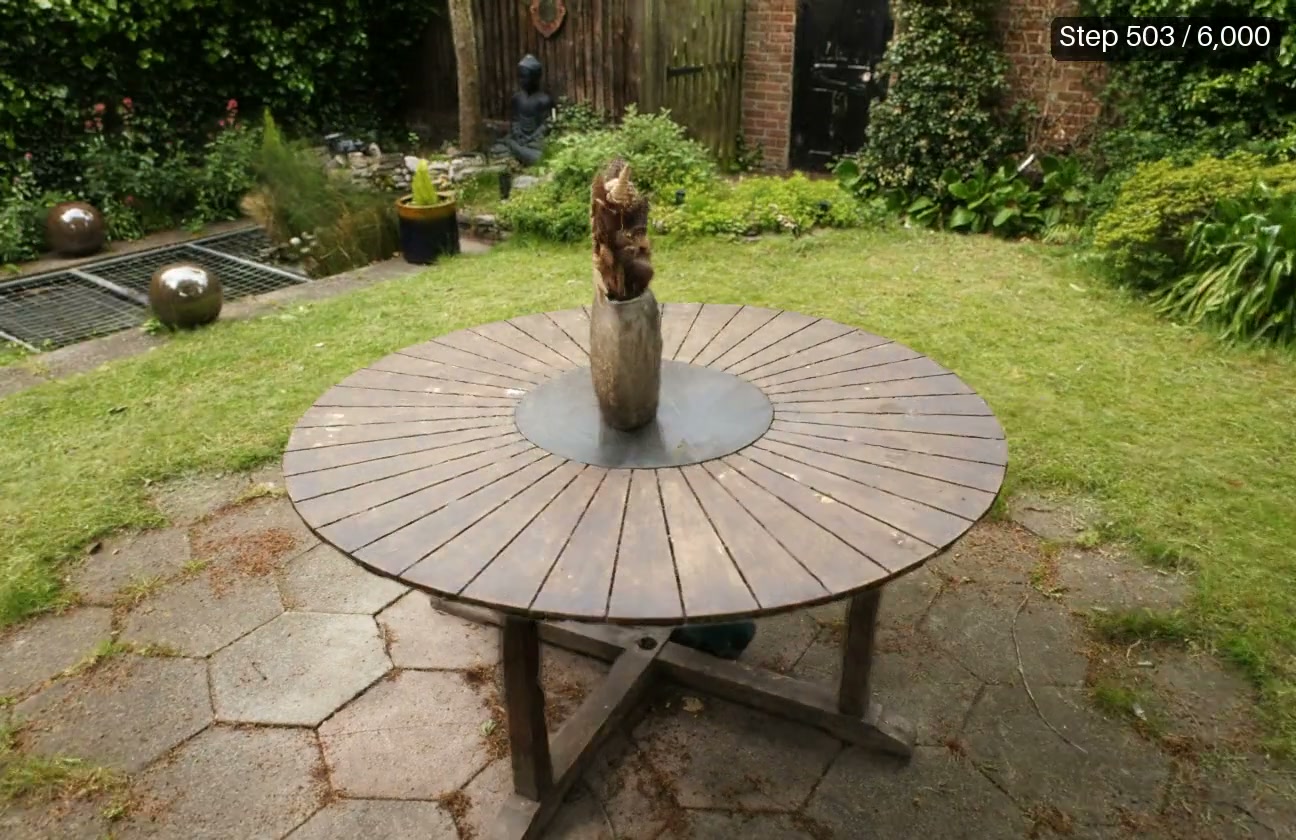} &
        \includegraphics[width=0.32\textwidth]{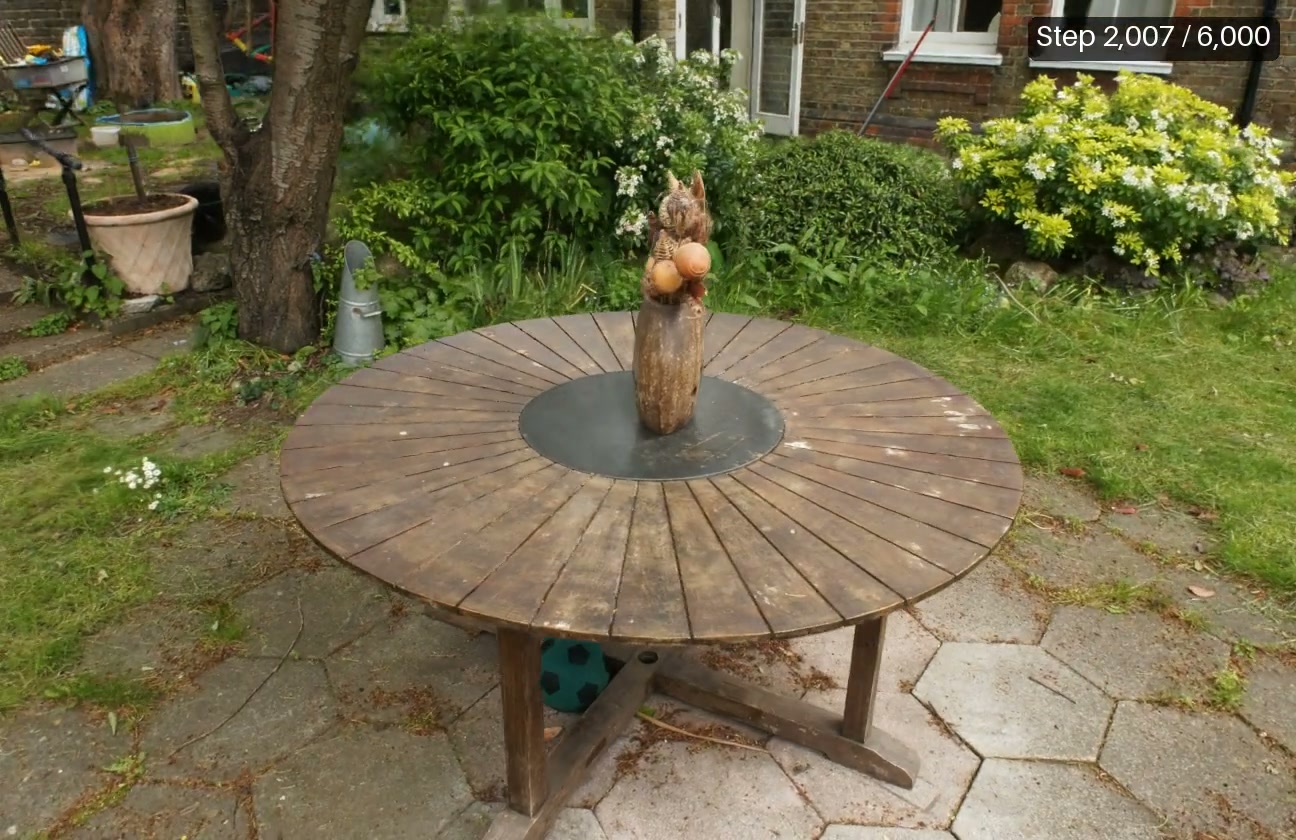} &
        \includegraphics[width=0.32\textwidth]{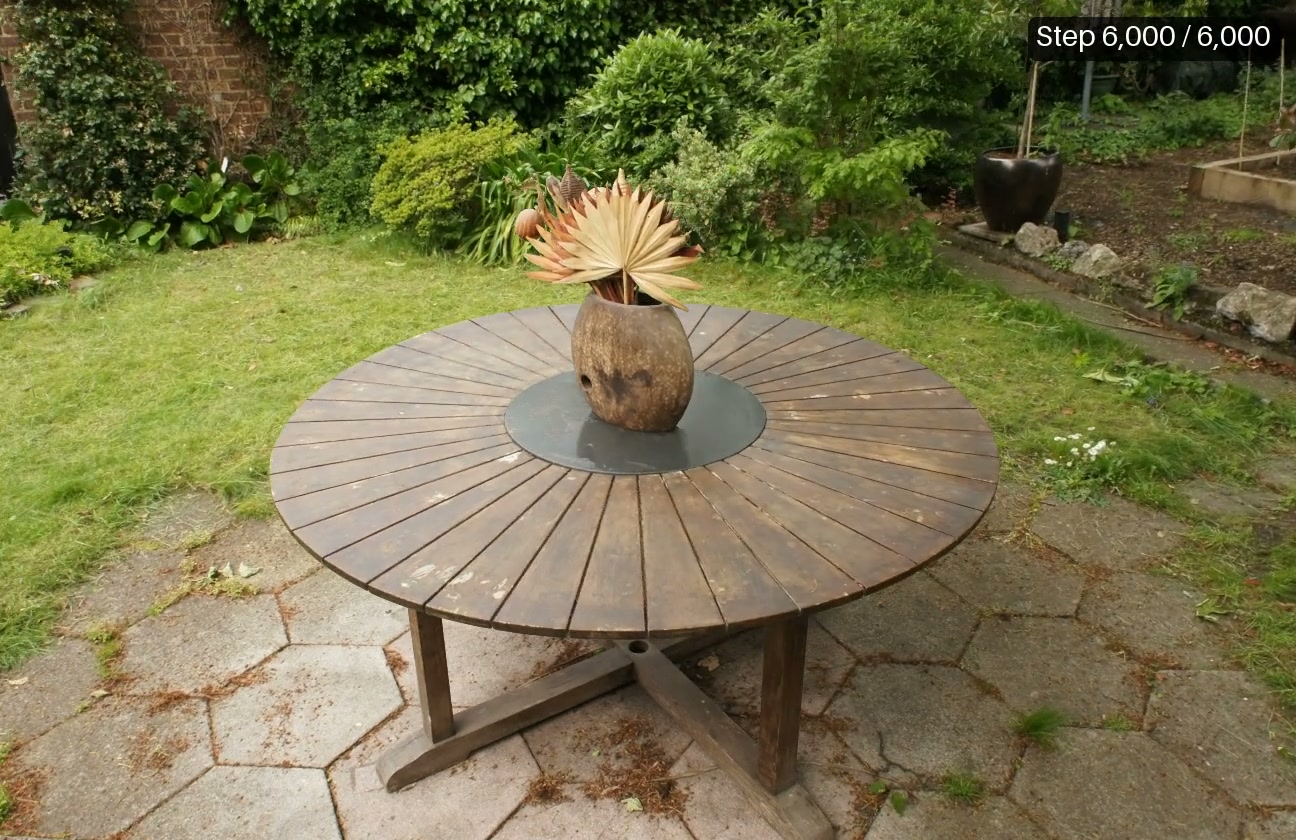} \\
        \small Step 503 & \small Step 2,007 & \small Step 6,000
    \end{tabular}
    \caption{\textbf{Shell evolution during training} on \textit{garden}. Renderings during training after 1, 41, 101, 503, 2,007, and 6,000 of the 6,000 iterations. The image at step 503 is still rendered with four semi-transparent shells, but has already converged to one almost opaque surface. Reconstructing accurate view-dependent colors requires an order of magnitude more steps.}
    \label{fig:suppl_training_progress}
\end{figure*}

\paragraph{Initialization and densification.}
We use all COLMAP SfM points and add one support point per SfM point $\mathbf p$ at $\mathbf p + \mathbf u\,(0.8\,d_{\text{nn}}(\mathbf p) + 0.1)$, where $\mathbf u \sim \mathcal U[0,1)^3$ and $d_{\text{nn}}$ is the distance to the nearest other SfM point. SfM points are initialized with an SDF value of zero and support points with a positive value. Since all initial SDF values are non-negative, only the two outer, low-opacity shells ($\tau > 0$) exist at initialization; the inner, more opaque shells only appear once the optimization drives SDF values below zero. The number of support points is capped at $10^6$. Densification is performed at iterations 1,000, 2,000, 3,000, and 4,000 using the accumulated SDF-space gradient with a threshold of $3\cdot10^{-6}$. The Delaunay tetrahedralization is recomputed every 10 iterations.

\paragraph{Appearance.}
The color hash grid uses 6 levels with a base resolution of 256 and a capacity of $2^{22}$ entries per level. During training, the background is filled with a random color so that the model cannot rely on a fixed background color to explain the image; for validation, we use a black background. We use no dedicated background model: distant content such as the sky is represented by the same mesh as the rest of the scene, and the convex-hull constraint (Sec.~3 of the main paper) keeps the mesh closed at the boundary of the point cloud.

\paragraph{Depth of field.}
We model defocus following DoF-Gaussian~\cite{shen2025dofgaussiancontrollabledepthoffield3d} with autofocus. The aperture scale is initialized to 0.01 and optimized with a constant learning rate of $10^{-3}$, the focus is estimated from a $9\times9$ patch, and the blur radius is limited to 3\% of the image size.

\paragraph{Optimization.}
Each step uses a single training image (batch size 1). We use Adam with $\epsilon = 10^{-15}$ and clip the gradient norm to 1. The loss weights are $0.8$ for $\mathcal{L}_1$, $0.2$ for SSIM, and $0.001$ for an L2 regularization of the hash grid features. The initial learning rates are $3\cdot10^{-3}$ for the support positions, $3\cdot10^{-2}$ for the SDF values, $5\cdot10^{-2}$ for the hash grid features, and $10^{-2}$ for the color MLP. The learning rate of the support positions decays exponentially over the whole training to $1\%$ of its initial value. All other learning rates are kept constant until iteration 4,000 and then decay exponentially to $10\%$ of their initial value at the last iteration.

\subsection{Differentiable Mesh Rendering}
\label{sec:suppl_rasterizer}
This section gives implementation details of the CUDA rasterizer. Fragment colors are queried from the hash grid inside the renderer, since evaluating them in a separate pass would be too costly for many layers.

\paragraph{Binning and sorting.}
Triangles are first binned into $16\times16$ screen-space tiles with a scanline rasterization of the triangle at tile resolution. A counting pass and a prefix sum allocate the per-tile lists, which a second pass fills. Each tile list is then refined into per-pixel fragment lists with the same count--prefix-sum--fill scheme. Each entry stores the triangle index, and the fragments of a pixel are sorted by the perspective-correct depth at the pixel with a segmented stable CUDA sort, which yields the strict front-to-back order required for compositing.

\paragraph{Visibility pass.}
Each pixel traverses its sorted fragments sequentially while maintaining the $m\times m$ occupancy mask ($m=8$, i.e., one 64-bit word per pixel). For each fragment, the covered pixel fraction $\kappa_i\in[0,1]$ is computed analytically from the triangle--pixel overlap, and the shell opacity $\alpha_i$ is interpolated from the per-vertex opacities. The effective opacity is $\hat\alpha_i = \kappa_i\alpha_i$. This pass assigns each fragment to a layer and stops as soon as the remaining transmittance falls below $1/256$. Fragments with an empty mask are discarded. Only the fragments that are retained are written to a compact list together with their layer index and $\hat\alpha_i$.

\paragraph{Shading and compositing.}
The color $c_i$ of every retained fragment is evaluated once, in parallel over the compact list, with the hash grid and the MLP at six samples on a hexagon around the pixel center, rather than once per subpixel. A fragment that covers only unoccupied subpixels is accumulated additively into the current layer,
\begin{equation}
C_L \leftarrow C_L + \hat{\alpha}_i c_i,
\qquad
A_L \leftarrow A_L + \hat{\alpha}_i,
\end{equation}
which is valid because the supports of such fragments are approximately disjoint. When a fragment overlaps occupied subpixels, the current layer is composited into the running pixel state,
\begin{equation}
C \leftarrow C + T\,C_L,
\qquad
T \leftarrow T(1-A_L),
\end{equation}
where $T$ is the accumulated transmittance, and a new layer with an empty mask is started. This decomposition avoids over-attenuation at shared edges while retaining anti-aliasing at partially covered pixels.

\paragraph{Backward pass.}
During training, the color and transmittance state of every layer is stored in the forward pass. The gradients of all fragments can therefore be evaluated in parallel over the compact fragment list, which improves thread coherence and GPU occupancy, instead of repeating the sequential traversal.

\paragraph{Final mesh.}
Once reconstruction is complete, only a single opaque shell remains. The connected mesh is UV-unwrapped and the hash grid is baked into a texture (Sec.~\ref{sec:suppl_texture}), so that it can be rendered with standard hardware rasterization without sorting.

\subsection{Delaunay Tetrahedralization}
\label{sec:suppl_delaunay}
\paragraph{Repeated triangulation.}
The SDF is defined on the Delaunay tetrahedralization of the point cloud, with values interpolated barycentrically inside each tetrahedron. Because the points move during optimization, the tetrahedralization has to be recomputed regularly so that the connectivity follows the points. As this is slow and expensive, we do not retriangulate in every step but only every 10 iterations. In between, the connectivity is kept fixed while the points move, so the tetrahedralization is only approximately Delaunay. We did not observe a decrease in quality when reducing the retriangulation frequency from every step to every 10 steps.

\paragraph{Flips after retriangulation.}
Because the connectivity changes at every retriangulation, the interpolated SDF field changes with it. The SDF values of the points are optimized for the previous connectivity, so tetrahedra can flip and the extracted shells can briefly show artifacts such as long, thin sheets of triangles (Fig.~\ref{fig:suppl_triangulation_artifacts}). We occasionally observe this in early training, when the points still move substantially between retriangulations. The optimizer quickly recovers from these artifacts within a few iterations, and we did not observe any lasting effect on the reconstruction. In the last iteration, we recompute the tetrahedralization once more, so that the final mesh is extracted from a valid Delaunay tetrahedralization without flipped tetrahedra. At this point, the points have converged and barely move between retriangulations, so the final mesh does not show these artifacts.

\begin{figure}[ht]
    \centering
    \includegraphics[width=0.85\linewidth]{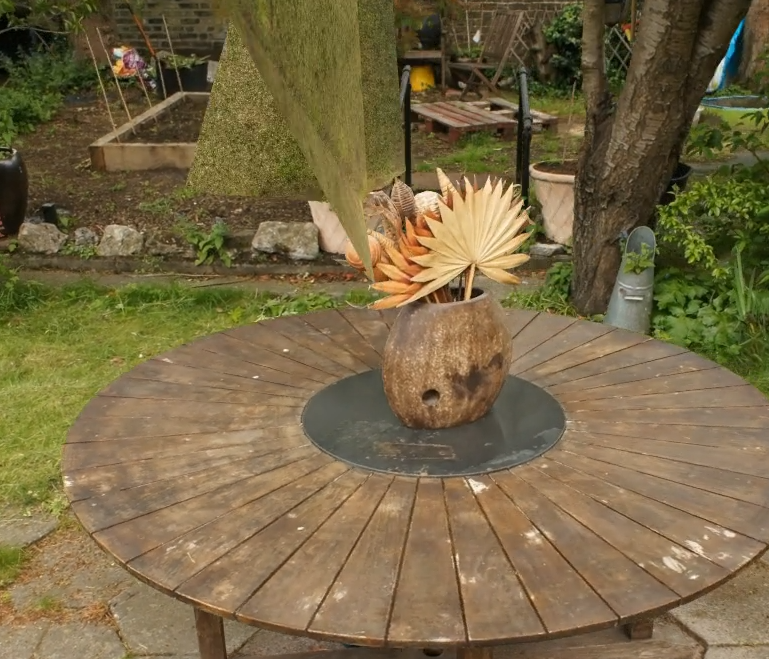}
    \caption{Artifact after a retriangulation in early training (\textit{garden}): the changed connectivity flips tetrahedra, which produces a sheet of triangles in the rendering (top). The optimizer quickly removes it again.}
    \label{fig:suppl_triangulation_artifacts}
\end{figure}

\paragraph{Triangulation speed.}
Tetrahedralization is a significant part of the reconstruction time. We therefore tested multiple Delaunay implementations, on the CPU (geogram~\cite{Levy2015geogram} and SciPy~\cite{2020SciPy-NMeth}) and on the GPU (gDel3D~\cite{cao2014gpu} and the implementation of Radiant Foam~\cite{govindarajan2025radiant}). We found geogram, which we use in all experiments, to be both the fastest and the most robust: gDel3D was slower than geogram, and the Radiant Foam implementation was not robust on our point clouds and caused errors. Degenerate point configurations that cause a triangulation error in geogram are a very rare occurrence. If such an error occurs, we add small noise ($10^{-8}$) to the point positions and retry until a triangulation is found. Developing or integrating a robust and fast GPU Delaunay triangulation is future work and could greatly reduce the reconstruction time.

\subsection{Texture Export}
\label{sec:suppl_texture}
The UV atlas is computed in the following steps:
\begin{itemize}
    \item \textbf{Texel budget.} We rasterize the mesh from every training camera at $2\times$ the image resolution and record, for each triangle, the largest number of pixels it covers as the frontmost surface in a single view. Counts are expressed in source-image pixels, so triangles that are visible only at subpixel scale receive a small positive demand. Held-out views are not used. Triangles that are not visible in any training view are \emph{unseen} and receive no atlas area. This defines the requested texel density of the triangle, with one texel per source pixel.
    \item \textbf{Unwrapping.} We unwrap the complete mesh in a single call with Smart UV Project~\cite{smartUV}, a Python implementation of Blender's projective unwrapper, which produces semantically meaningful UV islands.
    \item \textbf{Packing.} The islands are scaled to the maximum requested density of their triangles and packed with xatlas~\cite{young2018xatlas} with a one-texel padding. The atlas size is chosen automatically: charts are packed at full density, and each side of the canvas is rounded up to the next power of two to allow a complete GPU mipmap chain. The size is capped at $8192\times8192$; if the charts do not fit at full density, all densities are scaled down uniformly until they do. Unseen triangles are not part of the packing. They reuse the UV coordinates of the nearest observed chart, so that all triangles of the mesh are textured.
    \item \textbf{Initialization and fine-tuning.} Each texel is initialized by querying the hash grid at its world position, using the prefiltered lookup with the footprint of the texel. Texels next to the islands are filled by dilating the neighboring samples. Texture and SH coefficients are then fine-tuned for 1,500 steps on the training views with a differentiable rasterizer~\cite{Laine2020diffrast} using mipmapped texture lookups, matching standard renderers. We currently use \texttt{nvdiffrast} for this step, since our rasterizer does not yet support textures with mipmapping. Integrating mipmapped texture lookups into our rasterizer is future work; we do not expect it to change the quality of the results. Images are rendered with $2\times$ supersampling and compared with the ground-truth images. We use the same photometric loss as in the scene training, with constant learning rates of $0.001$ for the texture atlas, and $0.003$ for the spherical harmonics, and Adam with $\epsilon=10^{-12}$.
\end{itemize}
The entire export process takes about ten minutes on average for the Mip-NeRF 360 scenes (Tab.~\ref{tab:suppl_time_baking}). Only 1:14 minutes of that are spent in fine-tuning; the remaining time is spent in the preceding unwrapping, packing, and texture initialization steps and in writing the glTF file.

\section{Relation to Meshtryoshka and Mesh Splatting}
\label{sec:suppl_shell_methods}
Meshtryoshka~\cite{charatan2026meshtryoshka} and Mesh Splatting by Zhang~\etal~\cite{zhang2026meshsplatting} are the methods closest to ours. Both render several semi-transparent shells around the surface and use a stop-gradient so that the photometric loss reaches the geometry. The main differences lie in how the shells are built and how they relate to the final mesh.

\paragraph{Meshtryoshka.}
Meshtryoshka stores an SDF on a sparse voxel grid and extracts five shells with Marching Cubes. Each shell has a fixed transmittance that is chosen as a hyperparameter, so its opacity does not change during training.

The exported mesh is the zero level set. During training, this shell is rendered with an opacity of about 0.44, so it is never optimized as an opaque surface. The authors report that this costs little quality (24.30 vs.\ 24.34\,dB PSNR on Mip-NeRF 360).

The shells are rendered with a non-differentiable rasterizer at $3\times$ supersampled resolution. The geometry therefore receives gradients only through the shell opacities. Appearance is stored as spherical harmonics at the grid vertices. The authors report a training time of about four hours on a single NVIDIA H200 GPU.

\paragraph{Mesh Splatting.}
Zhang~\etal optimize an explicit base mesh. It is first extracted with DMTet~\cite{shen2021deep} on a $128^3$ grid and then refined with continuous remeshing~\cite{werner2022continuous}. The layers are created by offsetting the base vertices along their normals, within a fixed band of $\pm$10\,cm.

The layers only serve rendering; the exported mesh is the base mesh. Since the layers are offset surfaces and not level sets of a common field, they can overlap in concave regions. The fixed band also provides no gradients where the base mesh is far from the true surface, which the authors note for distant background. This likely makes scaling to unbounded scenes difficult, since these scenes contain large, distant regions where a coarse initial mesh is often far from the true surface.

The method additionally uses shading and monocular normal supervision. Quantitative results are reported only for object-centric scenes, and no texture export is described.

\paragraph{GTR-Mesh.}
Our shells differ from both approaches in three ways.

First, the opacity of our shells is annealed instead of fixed. The shells lie at fixed SDF levels, but their opacity $\sigma(-T\tau)$ depends on the temperature $T$. Training therefore starts with soft, volume-like shells and ends with a single opaque shell, which remains from iteration 2,185 of 6,000 on (Sec.~\ref{sec:suppl_hyperparameters}). This shell is the exported mesh. It is not a separate base surface, as in Mesh Splatting, and not a shell that was rendered semi-transparently during all of training, as in Meshtryoshka.

Second, the SDF is defined on the Delaunay tetrahedralization of a point cloud. The points are optimized, added where the SDF gradient indicates missing resolution, and removed where they do not contribute. Shell vertices therefore move with the points and are not tied to the edges of a fixed voxel grid, and resolution is added locally instead of by uniform subdivision.

Third, the SDF is constrained on the convex hull of the points. This keeps all shells closed in unbounded scenes without a separate background model.

Our differentiable rasterizer complements this parameterization. Once a shell is opaque, its opacity gradient $-T\alpha(1-\alpha)$ becomes very small. The analytic coverage gradients (Sec.~3.4 of the main paper) do not have this problem. They continue to refine the silhouettes and fine structures of the final mesh during the remaining two thirds of training.

\section{Evaluation Protocol and Baseline Numbers}
\label{sec:suppl_eval}
\paragraph{Protocol.}
Following 3DGS~\cite{kerbl_3d_2023}, we use every 8th image of each scene for testing.
Mip-NeRF 360~\cite{barron2022mipnerf360} images are downscaled by a factor of 4 for outdoor and 2 for indoor scenes.
On Tanks \& Temples~\cite{knapitsch_tanks_2017}, we evaluate the \textit{train} and \textit{truck} scenes.
We report PSNR, SSIM, and LPIPS (VGG)~\cite{zhang2018unreasonable}, computed on the exported UV-textured mesh.

\paragraph{Baselines in Table~1 of the main paper.}
\begin{itemize}
    \item \textbf{3DGS}~\cite{kerbl_3d_2023}, \textbf{Binary Opacity Grids}~\cite{reiser2024binary}, \textbf{MILo}~\cite{guedon2025milo}, and \textbf{Meshtryoshka}~\cite{charatan2026meshtryoshka}: numbers as reported in the respective publications. For MILo and Meshtryoshka, we use the results of the extracted meshes (mesh-based novel view synthesis), not of the underlying Gaussians or soft shells. Binary Opacity Grids reports outdoor and indoor scenes separately; we average them weighted by the number of scenes (5 outdoor, 4 indoor). Its face count is reported for the outdoor scenes only.
    \item \textbf{Blobs to Spokes}~\cite{gomez2026gaussianwrapping}: Mip-NeRF 360 numbers as reported in the publication (mesh-based novel view synthesis). For Tanks \& Temples, we ran the published code on the \textit{train} and \textit{truck} scenes, evaluated the extracted meshes with their evaluation protocol, and report these numbers for both scenes. The publication does not report numbers for the \textit{truck} scene.
    \item \textbf{MeshSplatting}~\cite{held2026mesh}: we ran the published code on all scenes, using bicubic image downscaling to match the other methods. Hence, both the Mip-NeRF 360 and the Tanks \& Temples numbers of MeshSplatting in Tab.~1 of the main paper come from our rerun. Compared to the published numbers, the rerun improves the averages on Mip-NeRF 360 (24.78~dB PSNR, 0.310 LPIPS, 0.728 SSIM) by 0.09~dB in PSNR, 0.009 in LPIPS, and 0.013 in SSIM, and on Tanks \& Temples (20.52~dB, 0.287, 0.745) by 0.05~dB, 0.016, and 0.015, respectively.
\end{itemize}

\paragraph{Baselines in Table~2 of the main paper.}
All DTU~\cite{jensen2014large} Chamfer distances are taken from the respective publications. We use the standard 15 DTU scans from the processed IDR/NeuS release, and train at full $1600\times 1200$ resolution with foreground masks for 6000 steps, with the same schedule and parameters as the Mip-Nerf reconstructions. Without running colmap, we initialize from 30,000 randomly placed points, and cap the densification to 100,000 supports. Unseen geomtry is underconstrained and tends to lie close to the surface, which reduces accuracy, so we remove unseen triangles before evaluation. For Triangle Splatting~\cite{held2025triangle}, we use the numbers reported by MeshSplatting~\cite{held2026mesh}. For Blobs to Spokes, we report the published numbers of the 30k-iteration variant.

\paragraph{Reproducing Blobs to Spokes.}
We could not reproduce the DTU numbers that Blobs to Spokes~\cite{gomez2026gaussianwrapping} reports, and the paper does not report numbers for the \textit{truck} scene of Tanks \& Temples. Neither the published code nor the paper describes how these datasets were reconstructed. We therefore take the numbers from the paper where we can (Mip-NeRF 360 and DTU), and report our own rerun for Tanks \& Temples. For qualitative results, we use the meshes published by the authors with per-vertex color. We have reached out to the authors to clarify the reconstruction settings and will update the Tanks \& Temples numbers if they differ.

\section{Additional Quantitative Results}
\label{sec:suppl_quant}
\subsection{Per-Scene Novel View Synthesis}
\label{sec:suppl_per_scene}
Tab.~\ref{tab:suppl_per_scene} lists our per-scene results. All metrics are computed on the exported mesh (GLB) with the baked texture. Mip-NeRF 360 indoor scenes are evaluated at half resolution and outdoor scenes at quarter resolution; LPIPS uses the VGG network. Tab.~\ref{tab:suppl_indoor_outdoor} compares the outdoor and indoor scenes separately against the baselines that report this split.

\begin{table}[ht]
    \centering
    \caption{Per-scene results of our method on Mip-NeRF 360 and Tanks \& Temples. $|V|$ denotes the number of vertices of the extracted mesh. }
    \resizebox{\linewidth}{!}{%
    \begin{tabular}{@{}ll|ccc|c@{}}
        & Scene & PSNR$\uparrow$ & SSIM$\uparrow$ & LPIPS$\downarrow$ & $|V|$ \\
        \hline
        \multirow{9}{*}{\rotatebox{90}{Mip-NeRF 360}}
        & Bicycle  & 23.963 & 0.7160 & 0.2329 & 2,789,148 \\
        & Flowers  & 20.613 & 0.5684 & 0.3024 & 2,831,545 \\
        & Garden   & 26.381 & 0.8356 & 0.1260 & 3,122,562 \\
        & Stump    & 25.119 & 0.7314 & 0.2531 & 2,861,813 \\
        & Treehill & 22.219 & 0.6322 & 0.2828 & 2,875,615 \\
        & Bonsai   & 29.293 & 0.9215 & 0.1892 & 3,095,747 \\
        & Counter  & 26.845 & 0.8802 & 0.2072 & 3,122,607 \\
        & Kitchen  & 29.163 & 0.9088 & 0.1446 & 3,299,052 \\
        & Room     & 29.303 & 0.9024 & 0.2098 & 3,030,013 \\
        \cline{2-6}
        & \textit{Average} & 25.878 & 0.7885 & 0.2164 & 3,003,122 \\
        \hline
        \multirow{3}{*}{\rotatebox{90}{T\&T}}
        & Train    & 19.726 & 0.7608 & 0.2222 & 2,766,323 \\
        & Truck    & 24.093 & 0.8660 & 0.1126 & 2,997,555 \\
        \cline{2-6}
        & \textit{Average} & 21.910 & 0.8134 & 0.1674 & 2,881,939 \\
    \end{tabular}}
    \label{tab:suppl_per_scene}
\end{table}

\begin{table}[ht]
    \centering
    \resizebox{\linewidth}{!}{%
    \begin{tabular}{@{}l|ccc|ccc@{}}
        & \multicolumn{3}{c|}{Outdoor} & \multicolumn{3}{c}{Indoor} \\
        Method & PSNR$\uparrow$ & LPIPS$\downarrow$ & SSIM$\uparrow$ & PSNR$\uparrow$ & LPIPS$\downarrow$ & SSIM$\uparrow$ \\
        \hline
        \textit{3DGS}~\cite{kerbl_3d_2023} & \textit{24.64} & \textit{0.234} & \textit{0.731} & \textit{30.41} & \textit{0.189} & \textit{0.920} \\
        \hline
        NeRF2Mesh~\cite{tang2023nerf2mesh}$^\ddagger$ & 22.36 & 0.478 & 0.493 & - & - & - \\
        Binary Opacity Grids~\cite{reiser2024binary} & \textbf{23.94} & 0.263 & 0.680 & 27.71 & 0.227 & 0.873 \\
        MeshSplatting~\cite{held2026mesh}$^\ast$ & 22.71 & 0.338 & 0.637 & 27.57 & 0.254 & 0.870 \\
        Ours & 23.66 & \textbf{0.239} & \textbf{0.697} & \textbf{28.65} & \textbf{0.188} & \textbf{0.903} \\
    \end{tabular}}
    \caption{Results on the outdoor (5 scenes) and indoor (4 scenes) scenes of Mip-NeRF 360~\cite{barron2022mipnerf360} for the methods that report this split; best mesh-based result in bold, 3DGS as non-mesh reference. MILo, Blobs to Spokes, Meshtryoshka, and Triangle Splatting report their mesh-based results only as an average over all scenes. Baseline numbers are taken from the respective publications. $^\ast$Our rerun with bicubic image downscaling (see Sec.~\ref{sec:suppl_eval}). $^\ddagger$Reported only on 3 outdoor scenes.}
    \label{tab:suppl_indoor_outdoor}
\end{table}

\subsection{DTU Geometry}
\label{sec:suppl_dtu}
Tab.~\ref{tab:suppl_dtu} compares our per-scan Chamfer distance with the baselines, and Tab.~\ref{tab:suppl_dtu_ours} lists our accuracy, completeness, and mesh size per scan. Fig.~\ref{fig:suppl_dtu_all} shows qualitative results for all scans. Our method does not handle scans 65, 97, and 110 well, which we attribute to strong specular highlights. A better directional color model or additional regularization could improve the results.

\begin{table}[ht]
    \caption{Per-scene Chamfer distance [mm] on DTU~\cite{jensen2014large}, compared to NeuS~\cite{wang2021neus} (without mask supervision), Triangle Splatting~\cite{held2025triangle}, MeshSplatting~\cite{held2026mesh}, MILo~\cite{guedon2025milo}, Zhang~\etal~\cite{zhang2026meshsplatting}, and Blobs to Spokes~\cite{gomez2026gaussianwrapping}. Baseline numbers are taken from the respective publications (Triangle Splatting from MeshSplatting).}
    \centering
    \resizebox{\linewidth}{!}{%
    \begin{tabular}{@{}l|ccccccc@{}}
        Scan & NeuS & TriSplat & MeshSplat & MILo & Zhang & B2S (30k) & Ours \\
        \hline
        24 & 1.00 & 0.98 & 0.77 & 0.43 & 0.46 & 0.40 & 0.44 \\
        37 & 1.37 & 1.07 & 0.72 & 0.74 & 0.73 & 0.51 & 0.55 \\
        40 & 0.93 & 1.07 & 0.74 & 0.34 & 0.49 & 0.27 & 0.30 \\
        55 & 0.43 & 0.51 & 0.60 & 0.37 & 0.43 & 0.32 & 0.36 \\
        63 & 1.10 & 1.67 & 0.89 & 0.80 & 0.77 & 0.81 & 0.88 \\
        65 & 0.65 & 1.44 & 1.00 & 0.74 & 0.82 & 0.46 & 0.89 \\
        69 & 0.57 & 1.17 & 0.81 & 0.70 & 0.65 & 0.45 & 0.57 \\
        83 & 1.48 & 1.32 & 1.09 & 1.21 & 1.03 & 1.04 & 1.04 \\
        97 & 1.09 & 1.75 & 1.19 & 1.22 & 0.95 & 0.65 & 0.93 \\
        105 & 0.83 & 0.98 & 0.58 & 0.66 & 0.52 & 0.52 & 0.51 \\
        106 & 0.52 & 0.96 & 0.68 & 0.62 & 0.58 & 0.33 & 0.48 \\
        110 & 1.20 & 1.11 & 0.93 & 0.80 & 0.59 & 0.59 & 1.21 \\
        114 & 0.35 & 0.56 & 0.63 & 0.37 & 0.37 & 0.32 & 0.35 \\
        118 & 0.49 & 0.93 & 0.66 & 0.76 & 0.44 & 0.33 & 0.40 \\
        122 & 0.54 & 0.72 & 0.59 & 0.48 & 0.42 & 0.34 & 0.38 \\
        \hline
        \textit{Mean} & 0.84 & 1.06 & 0.79 & 0.68 & 0.62 & 0.49 & 0.62 \\
    \end{tabular}}
    \label{tab:suppl_dtu}
\end{table}

\begin{table}[ht]
    \caption{Our per-scene geometric accuracy on DTU~\cite{jensen2014large}: Chamfer distance, accuracy, and completeness in mm, and number of mesh vertices.}
    \centering
    \small
    \begin{tabular}{@{}l|ccc|r@{}}
        Scan & Chamfer$\downarrow$ & Acc.$\downarrow$ & Comp.$\downarrow$ & $|V|$ \\
        \hline
        24 & 0.441 & 0.416 & 0.467 & 189{,}348 \\
        37 & 0.551 & 0.633 & 0.469 & 190{,}019 \\
        40 & 0.305 & 0.303 & 0.307 & 189{,}878 \\
        55 & 0.359 & 0.355 & 0.364 & 165{,}962 \\
        63 & 0.880 & 1.295 & 0.465 & 102{,}717 \\
        65 & 0.889 & 0.890 & 0.888 & 106{,}455 \\
        69 & 0.566 & 0.667 & 0.464 & 147{,}002 \\
        83 & 1.039 & 0.732 & 1.346 & 83{,}851 \\
        97 & 0.928 & 1.099 & 0.758 & 160{,}125 \\
        105 & 0.513 & 0.465 & 0.561 & 133{,}290 \\
        106 & 0.480 & 0.445 & 0.514 & 113{,}630 \\
        110 & 1.206 & 1.625 & 0.787 & 146{,}299 \\
        114 & 0.348 & 0.367 & 0.329 & 123{,}131 \\
        118 & 0.402 & 0.390 & 0.415 & 109{,}499 \\
        122 & 0.376 & 0.380 & 0.372 & 99{,}576 \\
        \hline
        \textit{Mean} & 0.619 & 0.671 & 0.567 & 137{,}385 \\
    \end{tabular}
    \label{tab:suppl_dtu_ours}
\end{table}

\section{Mesh Validity}
\label{sec:suppl_validity}
\label{sec:suppl_validity_check}
We check the exported meshes with Open3D~\cite{Zhou2018} after merging duplicate vertices (\texttt{remove\_duplicated\_vertices}). A mesh is \emph{watertight} if every edge is shared by exactly two faces, so that it has neither boundary edges (holes) nor non-manifold edges; we test this with \texttt{get\_non\_manifold\_edges} with boundary edges not allowed. It is \emph{vertex-manifold} if the faces around each vertex form a single fan, which excludes vertices where two surfaces touch at a point; we test this with \texttt{get\_non\_manifold\_vertices}. For \emph{self-intersections}, Open3D provides \texttt{get\_self\_intersecting\_triangles}, which compares all pairs of triangles and does not scale to our meshes with millions of faces. We therefore use a custom GPU implementation that tests triangle pairs for intersection, ignoring pairs that share a vertex and coplanar overlaps. All checks pass for our meshes.

\section{Runtime, Memory, and Storage}
\label{sec:suppl_runtime}
\subsection{Training and Storage}
\label{sec:suppl_training_storage}
\paragraph{Training.} We train all scenes on an NVIDIA RTX 5090. On Mip-NeRF 360, training takes 63.0 minutes on average, including UV unwrapping and texture fine-tuning. Tab.~\ref{tab:suppl_time_training} breaks the training time down by step, and Tab.~\ref{tab:suppl_time_baking} does the same for baking. The repeated Delaunay tetrahedralization, which runs on the CPU, takes almost a third of the training time. Baking the texture adds about 10 minutes.

\paragraph{Storage.} On Mip-NeRF 360, an exported glTF file takes 991.5\,MB on average. Of this, 163.4\,MB is the texture atlas, 752.1\,MB are vertex attributes, and 76.1\,MB are triangles. The vertex attributes include 115.4\,MB of duplicated vertices: glTF cannot store attributes that differ across a UV seam, so vertices on seams have to be duplicated. Formats such as USD could store this more efficiently, but they cannot be loaded by all tools, so we use glTF.

\begin{table}[ht]
    \centering
    \caption{Mean reconstruction time over the nine
    Mip-NeRF~360 scenes for 6,000 training steps,
    on an NVIDIA GeForce RTX~5090
    and an Intel Core Ultra~9~285K. GLB export is excluded here but broken down in Tab.~\ref{tab:suppl_time_baking}.}
    \resizebox{\linewidth}{!}{%
    \begin{tabular}{@{}lrr@{}}
        Step & Time [min:s] & Share [\%] \\
        \hline
        Delaunay rebuilds  & 15:50 & 29.87 \\
        Marching tetrahedra (forward)                    &  0:13 &  0.41 \\
        Rendering, forward pass                          &  5:44 & 10.82 \\
        Full training backward pass                      & 20:07 & 37.96 \\
        Optimizer step                                   &  1:32 &  2.89 \\
        Other training and evaluation work               &  9:34 & 18.02 \\
        \hline
        \textit{Total}                                   & 53:00 & 100.00 \\
    \end{tabular}}
    \label{tab:suppl_time_training}
\end{table}

\begin{table}[ht]
    \centering
    \caption{GLB baking time breakdown by step, averaged over the nine Mip-NeRF 360 scenes on an RTX~5090.}
    \resizebox{\linewidth}{!}{%
    \begin{tabular}{@{}lcc@{}}
        Step & Time [min:s] & Share [\%] \\
        \hline
        UV unwrapping & 0:57 & 9.5 \\
        Reprojection of sliver triangles & 2:29 & 24.8 \\
        Atlas packing & 1:14 & 12.3 \\
        Texel budget and texture initialization & 3:13 & 32.1 \\
        Texture and SH fine-tuning & 1:14 & 12.3 \\
        Mesh export (glTF) & 0:54 & 9.0 \\
        \hline
        \textit{Total} & 10:01 & 100.0 \\
    \end{tabular}}
    \label{tab:suppl_time_baking}
\end{table}

\subsection{Real-Time Rendering}
\label{sec:suppl_rendering}
The exported mesh is rendered with the standard hardware rasterization pipeline and does not require the differentiable rasterizer used during training. We implemented renderers for Metal and Vulkan. Metal is used on Apple devices such as the MacBook Air M3, and Vulkan on desktop GPUs such as the NVIDIA RTX A5000. Both renderers use task and mesh shaders. The mesh is split into clusters of at most 64 vertices and 126 triangles, and a task shader culls clusters against the view frustum and their normal cone. For each visible cluster, the mesh shader evaluates the view-dependent color of its points from the spherical harmonics coefficients, so this is done only for visible clusters instead of for every vertex of the mesh. The fragment shader then needs only a single RGBA texture lookup, and combines it with the interpolated per-vertex colors. We render with $4\times$ MSAA. Tab.~\ref{tab:suppl_fps} lists the frame rates.

\begin{table}[ht]
    \centering
    \caption{Rendering speed (FPS) of our exported textured mesh (GTR) and 3DGS per scene on a MacBook Air M3 and an NVIDIA RTX A5000. 3DGS is rendered with the hardware rasterizer of Niedermayr~\etal~\cite{Niedermayr_2024_CVPR}. All methods are rendered at the training resolution, our mesh with $4\times$ MSAA.}
    \begin{tabular}{@{}ll|cc|cc@{}}
        & & \multicolumn{2}{c|}{MacBook Air M3} & \multicolumn{2}{c}{RTX A5000} \\
        & Scene & GTR & 3DGS & GTR & 3DGS \\
        \hline
        \multirow{9}{*}{\rotatebox{90}{Mip-NeRF 360}}
        & Bicycle  & 223 & 32 & 1608 & 192 \\
        & Flowers  & 200 & 40 & 1057 & 248 \\
        & Garden   & 157 & 22 & 852 & 155 \\
        & Stump    & 324 & 36 & 1501 & 241 \\
        & Treehill & 218 & 33 & 1111 & 207 \\
        & Bonsai   & 150 & 62 & 781 & 293 \\
        & Counter  & 136 & 42 & 705 & 204 \\
        & Kitchen  & 99 & 31 & 514 & 176 \\
        & Room     & 303 & 53 & 1280 & 259 \\
        \cline{2-6}
        & \textit{Average} & 201 & 39 & 1045 & 219 \\
        \hline
        \multirow{3}{*}{\rotatebox{90}{T\&T}}
        & Train    & 174 & 34 & 909 & 162 \\
        & Truck    & 140 & 34 & 752 & 182 \\
        \cline{2-6}
        & \textit{Average} & 157 & 34 & 831 & 172 \\
    \end{tabular}
    \label{tab:suppl_fps}
\end{table}

\subsection{Comparison with nvdiffrast}
\label{sec:suppl_nvdiffrast}
We compare our renderer (referred to as Ours in the tables) to nvdiffrast with depth peeling on \emph{Chair} from the NeRF Synthetic dataset~\cite{mildenhall_nerf_2021}.
Both renderers use the same hash-grid appearance model, with multisample mip filtering disabled throughout for a fair comparison.
Under identical training settings, our renderer achieves a PSNR of $33.97$\,dB compared to $32.47$\,dB for nvdiffrast.
On the same checkpoint and view with four soft shells, our renderer is $2.8\times$ faster for the combined forward and backward passes ($66$ vs.\ $184$\,ms on an RTX~4500 Ada), and nvdiffrast retains $3.3\times$ as much renderer state for backward ($1{,}478$ vs.\ $448$\,MiB). These are the speed and memory factors quoted in the introduction of the main paper. They depend on the scene and on the number of rendered shells; for the larger \emph{Garden} scene and averaged over the whole training, both renderers run at a similar speed, as detailed in the paragraph on speed and memory below.

\paragraph{Image formation.}
To investigate the quality gap, we render the same \emph{Garden} checkpoint from Mip-NeRF~360 with both renderers (Tab.~\ref{tab:garden-point-attenuation-aa}).
Without anti-aliasing, both renderers produce practically the same validation images. With anti-aliasing, nvdiffrast trails ours by $0.461$\,dB at native resolution.
Rendering nvdiffrast at $2\times$ resolution and downsampling matches the PSNR of our coverage-based anti-aliasing at $1\times$ ($26.18$ vs.\ $26.19$\,dB).

\begin{table}[t]
  \centering
  \caption{Validation metrics on Mip-NeRF 360 Garden using a checkpoint trained with our renderer without supersampling.
    Both renderers use point sampling with footprint-based attenuation of the hash-grid levels.
    Supersampling is per dimension before box-filter downsampling.
    Bold values compare ours at native resolution with $2\times$ nvdiffrast.}
  \label{tab:garden-point-attenuation-aa}
  \small
  \setlength{\tabcolsep}{3pt}
  \begin{tabular}{@{}cclrrr@{}}
    \toprule
    Supersampling & AA & Renderer & PSNR$\uparrow$ & SSIM$\uparrow$ & LPIPS$\downarrow$ \\
    \midrule
    $1\times$ & Off & Ours       & 25.674 & 0.8066 & 0.1567 \\
              &     & nvdiffrast & 25.671 & 0.8065 & 0.1568 \\
    \addlinespace[2pt]
              & On  & Ours       & \textbf{26.191} & \textbf{0.8234} & \textbf{0.1402} \\
              &     & nvdiffrast & 25.730 & 0.8077 & 0.1558 \\
    \midrule
    $2\times$ & Off & Ours       & 26.183 & 0.8253 & 0.1421 \\
              &     & nvdiffrast & 26.182 & 0.8253 & 0.1421 \\
    \addlinespace[2pt]
              & On  & Ours       & 26.310 & 0.8289 & 0.1391 \\
              &     & nvdiffrast & \textbf{26.183} & \textbf{0.8254} & \textbf{0.1419} \\
    \bottomrule
  \end{tabular}
\end{table}

\paragraph{Optimization.}
We next train Garden separately with each renderer and evaluate the fitted GLBs under the same settings (Tab.~\ref{tab:garden-glb-training-costs}).
At $1\times$, training with our renderer gives a PSNR of $26.32$\,dB compared to $25.72$\,dB for nvdiffrast.
Although rendering the fixed checkpoint with nvdiffrast at $2\times$ matches the PSNR of our renderer at $1\times$, training nvdiffrast at $2\times$ raises the fitted GLB PSNR by only $0.01$\,dB, to $25.73$\,dB.
That suggests that the geometry gradients do not get more accurate with higher resolution for nvdiffrast because geometry is still point-sampled before the screen-space antialiasing. The quality gap in optimization between nvdiffrast and our renderer does not close by rendering at higher resolution.

\paragraph{Speed and memory.}
With four shells at native resolution on Garden, the combined forward and backward passes take $335$\,ms per frame with our renderer and $381$\,ms with nvdiffrast.
When only one opaque shell is rendered, nvdiffrast is almost twice as fast as our renderer ($43.7$ vs.\ $80.4$\,ms per frame).
Averaged over training, a rendering forward and backward pass takes $170$\,ms with our renderer and $163$\,ms with nvdiffrast, with peak renderer memory of $5.7$ and $12.0$\,GiB, respectively.
At $2\times$, our renderer takes $571$\,ms per frame and uses $10.1$\,GiB. Training \emph{Garden} at $2\times$ with nvdiffrast would take an estimated $45$\,GiB of VRAM, which is more than we have access to. We implement a tiled renderer that reduces VRAM usage at the cost of performance to measure the feasibility of using nvdiffrast for training. Special care was taken to ensure the equivalence of the tiled renderer's output and gradients with the untiled version.

\begin{table}[t]
  \centering
  \caption{Fitted GLB quality and training cost on Mip-NeRF~360 Garden.
    Time is the mean forward and backward time combined over 6{,}000 steps on RTX~5090 GPUs.
    VRAM is the peak required only by the renderer.
    The $2\times$ nvdiffrast run renders the full image as tiles to reduce memory requirements. We estimate $580$\,ms/frame and $45$\,GiB for full-frame rendering.}
  \label{tab:garden-glb-training-costs}
  \small
  \setlength{\tabcolsep}{3pt}
  \begin{tabular}{@{}lrr@{\hspace{8pt}}rr@{}}
    \toprule
    & \multicolumn{2}{c}{$1\times$ training} & \multicolumn{2}{c}{$2\times$ training} \\
    \cmidrule(lr){2-3}\cmidrule(l){4-5}
    & Ours & nvdiffrast & Ours & nvdiffrast \\
    \midrule
    PSNR$\uparrow$  & 26.322 & 25.720 & 26.389 & 25.729 \\
    SSIM$\uparrow$  & 0.8343 & 0.8041 & 0.8366 & 0.8033 \\
    LPIPS$\downarrow$ & 0.1264 & 0.1525 & 0.1238 & 0.1527 \\
    \addlinespace[3pt]
    Time (ms/frame) & 170.34 & 162.92 & 571.23 & 2{,}709 \\
    VRAM (GiB)      & 5.7    & 12.0   & 10.1   & 3.6 \\
    \bottomrule
  \end{tabular}
\end{table}

\section{Additional Ablations}
\label{sec:suppl_ablations}
Tab.~\ref{tab:additional-ablation} complements the ablations in the main paper, again averaged over all nine Mip-NeRF 360 scenes. Removing either the hash grid or the spherical harmonics from the color model reduces PSNR by about 2\,dB, so both the view-independent and the view-dependent part are needed. Replacing the learned strength $w$ by the constant $w\equiv1$, i.e., adding the full view-dependent color everywhere, reduces PSNR by 0.65\,dB and increases LPIPS from 0.216 to 0.276. Training with a single shell from the start for 12k steps, i.e., twice the number of steps of our full model, still trails the full model by 0.56\,dB, so the benefit of multiple shells is not explained by a longer optimization alone. Retriangulating in every step instead of every 10 steps (Sec.~\ref{sec:suppl_delaunay}) does not measurably change the quality ($+0.02$\,dB) but increases the training time from 63.0 to 194.1 minutes. 
\begin{table}[tbp]
    \centering
    \caption{Additional ablations on Mip-NeRF 360. Training time includes texture unwrapping and fine-tuning.}
    \resizebox{\linewidth}{!}{%
    \begin{tabular}{l|cccc}
         & PSNR$\uparrow$ & $\Delta$PSNR & LPIPS$\downarrow$ & Time [min]$\downarrow$ \\
        \hline
        Full & 25.88 & & 0.216 & 63.0 \\
        \hline
        $-$Hash grid & 23.80 & $-2.08$ & 0.412 & 39.8 \\
        $-$SH & 23.91 & $-1.97$ & 0.260 & 50.5 \\
        $-$Learned strength ($w\equiv1$) & 25.23 & $-0.65$ & 0.276 & 54.7 \\
        \hline
        Single shell, 12k steps & 25.32 & $-0.56$ & 0.238 & 59.4 \\
        Retriangulate every step & 25.90 & $+0.02$ & 0.214 & 194.1 \\
    \end{tabular}}
    \label{tab:additional-ablation}
\end{table}

\section{Additional Qualitative Results}
\label{sec:suppl_visualizations}
We show additional qualitative results in Fig.~\ref{fig:suppl_bicycle_closeup}, Fig.~\ref{fig:suppl_overview_1} and Fig.~\ref{fig:suppl_overview_2}.
For the qualitative comparisons with Blobs to Spokes on Mip-NeRF 360, we use the meshes with per-vertex color that the authors publish, since we were unable to reproduce their published results with our own training runs (see Sec.~\ref{sec:suppl_eval}). The quantitative Mip-NeRF 360 numbers of Blobs to Spokes are the published ones (Sec.~\ref{sec:suppl_eval}). \\
Fig.~\ref{fig:suppl_bicycle_closeup} shows a closeup of the \emph{bicycle} scene, colored and in an untextured mesh view. Geometric details like the spokes are represented by geometry, while regions with high frequency details such as the markings on the tire or the flag on the bar are encoded only in the texture.
\begin{figure*}[tbp]
  \centering
  \includegraphics[width=\textwidth]{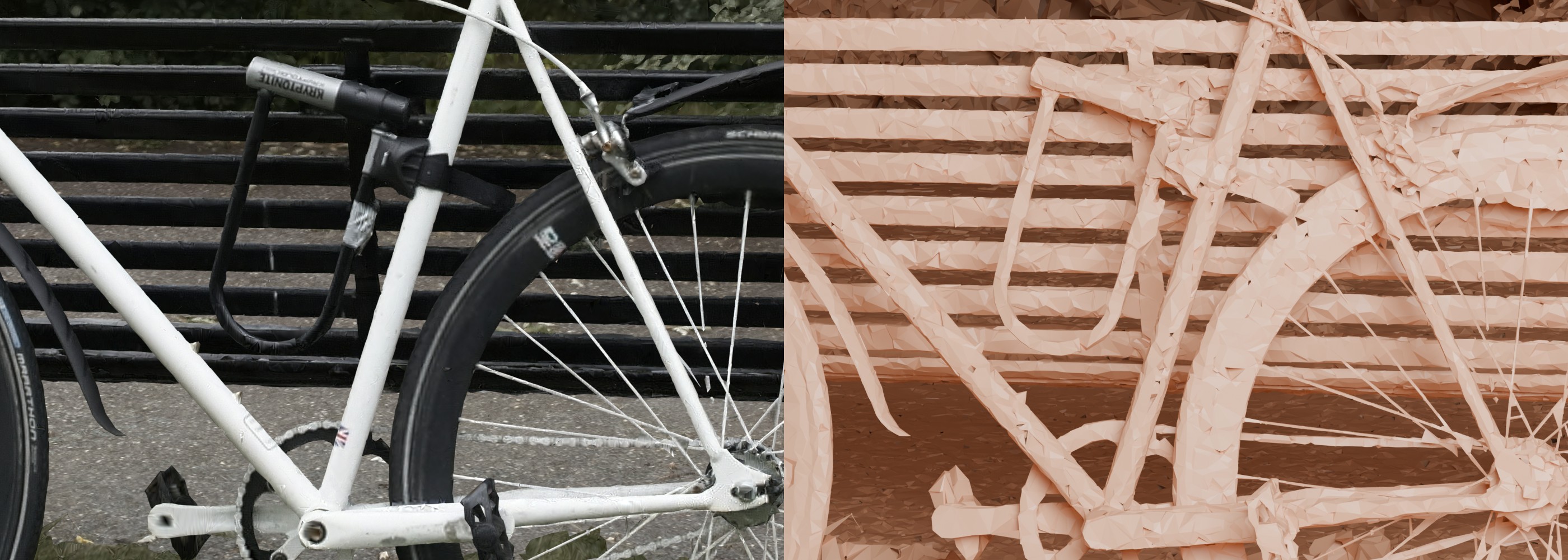}
  \caption{\textbf{Close-up of the \textit{bicycle} scene.} Rendered at 4$\times$ the training resolution. Left: textured mesh; right: the same mesh without texture. Texture details like the British flag low on the bar do not increase the vertex count.}
  \label{fig:suppl_bicycle_closeup}
\end{figure*}

We show closeups of mesh renders from GTR-Mesh, Blobs to Spokes, and MeshSplatting in Fig.~\ref{fig:suppl_overview_1}. Our method has significantly fewer artifacts and sharper textures.

\begin{figure*}[tbp]
    \centering
    \setlength{\tabcolsep}{1pt}
    \begin{tabular}{@{}ccc@{}}
        \small Ours & \small Blobs to Spokes & \small MeshSplatting \\
        \includegraphics[width=0.315\textwidth]{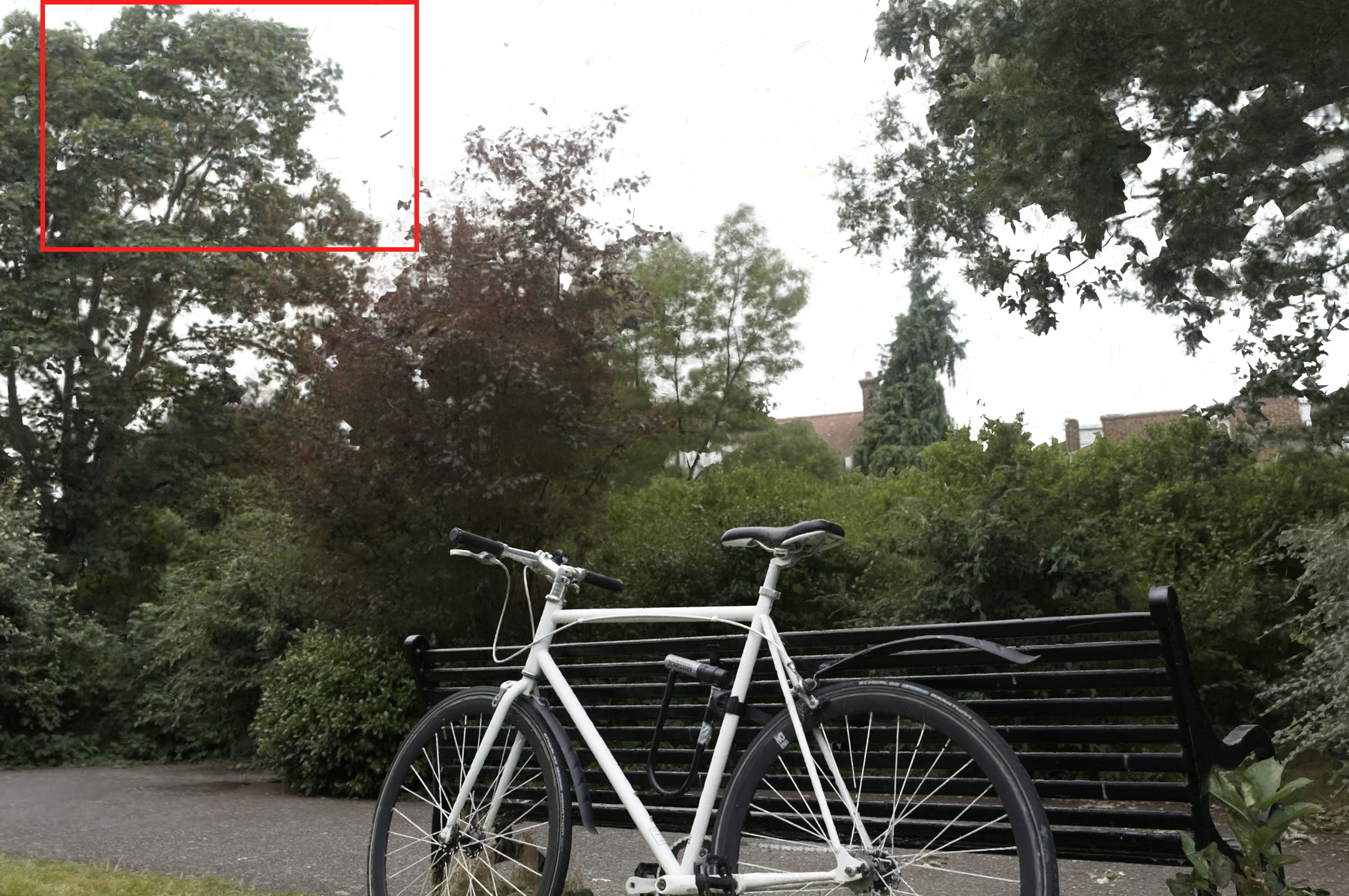} &
        \includegraphics[width=0.315\textwidth]{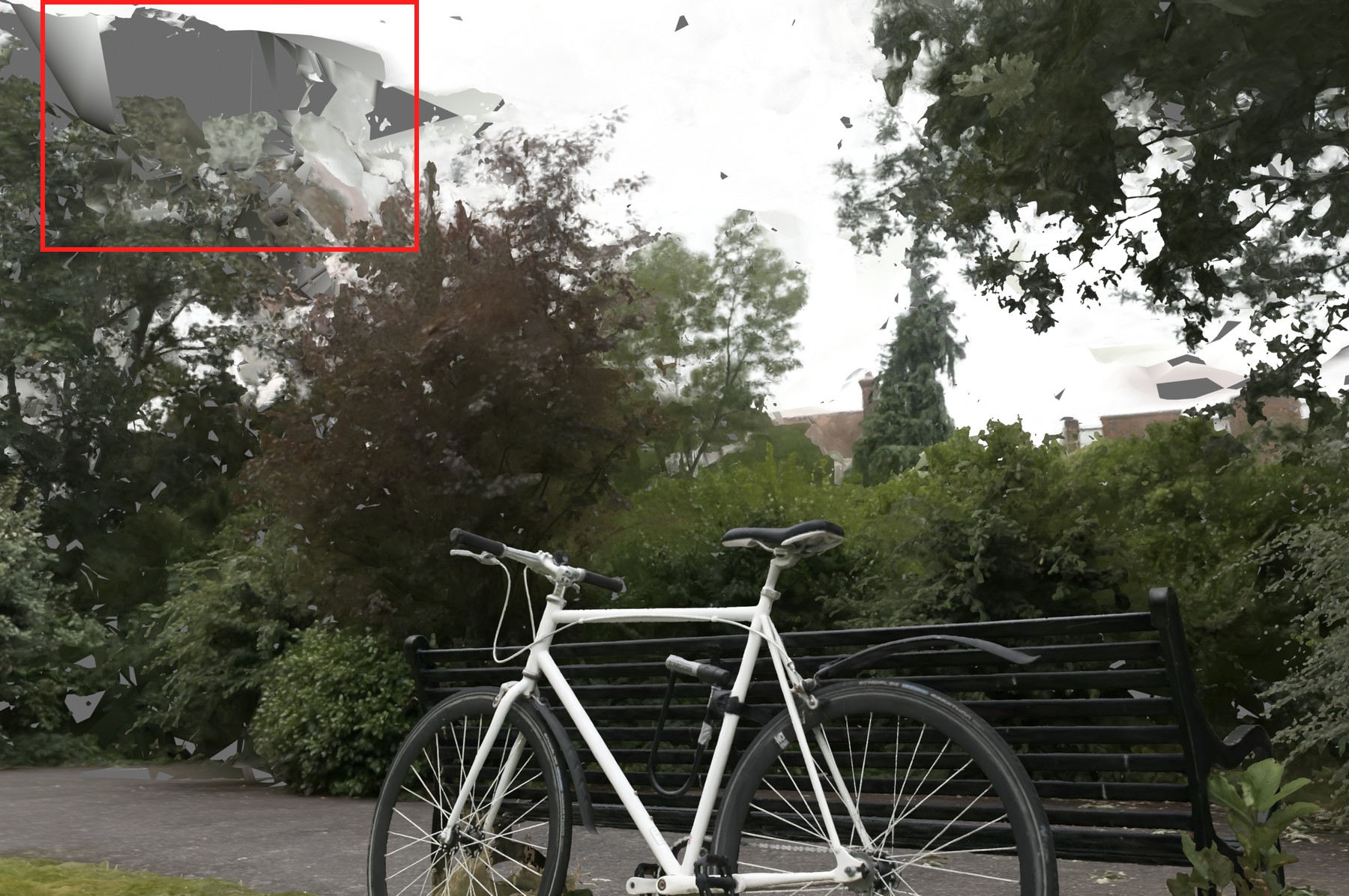} &
        \includegraphics[width=0.315\textwidth]{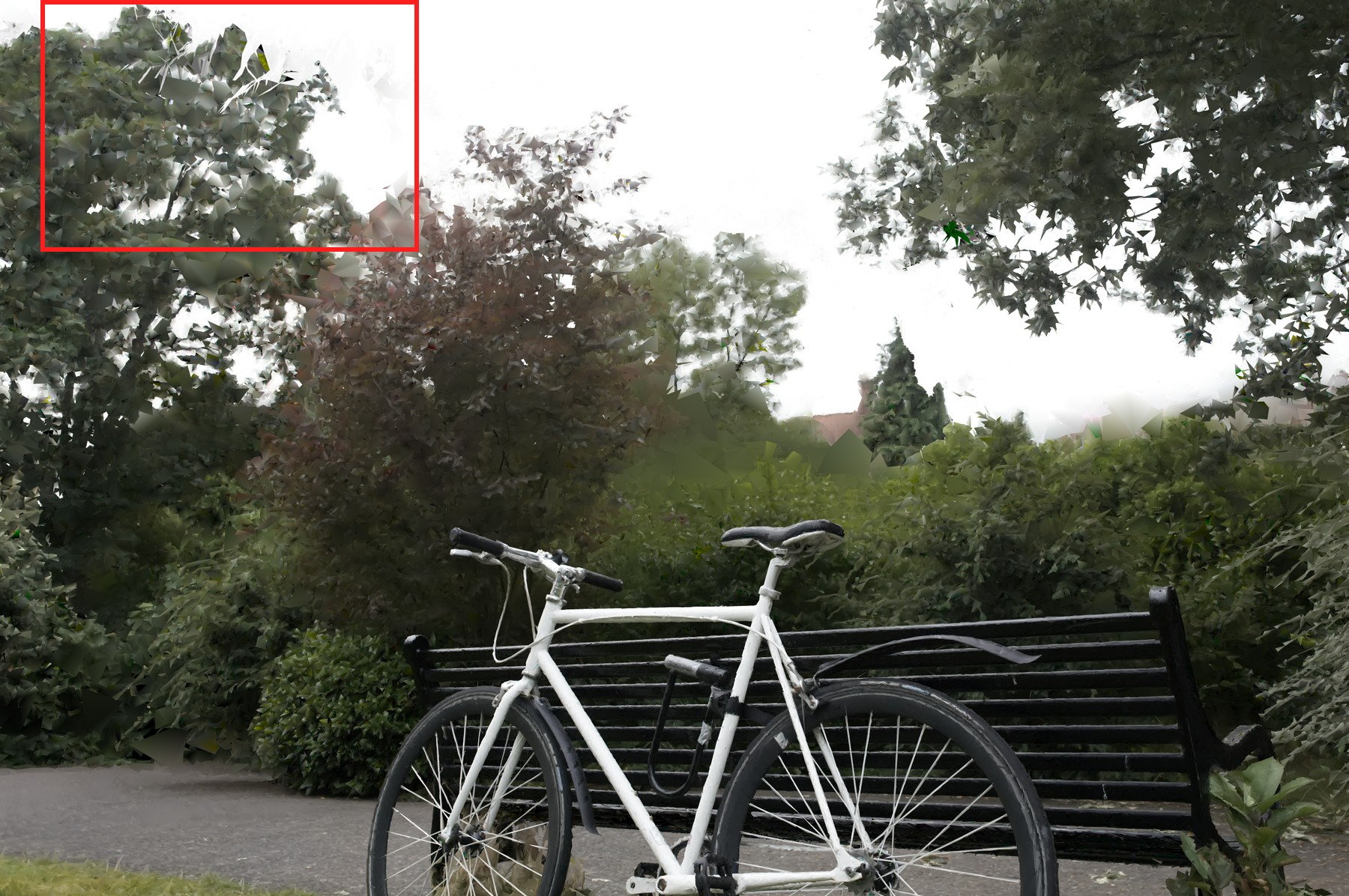} \\
        \includegraphics[width=0.315\textwidth]{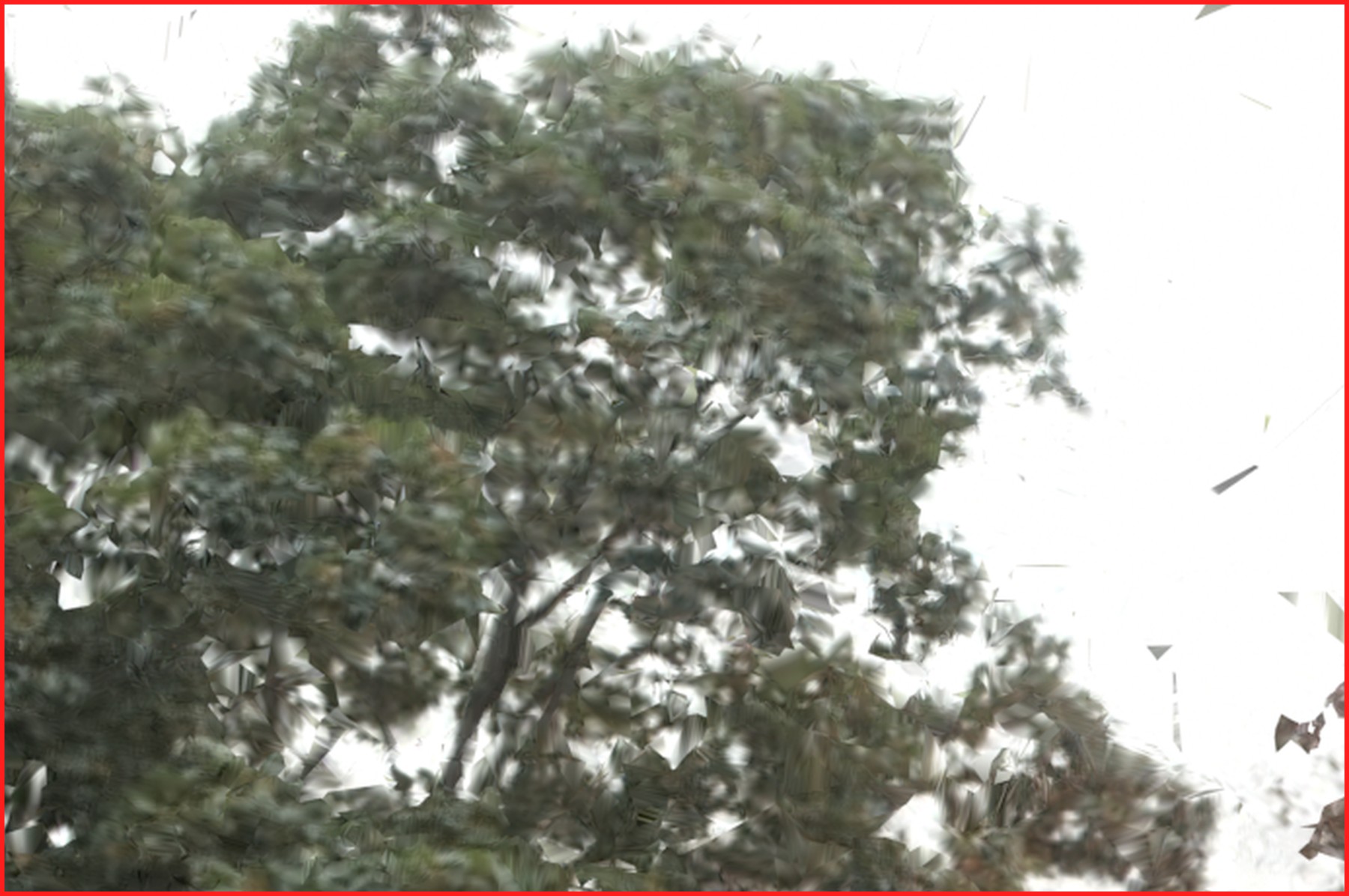} &
        \includegraphics[width=0.315\textwidth]{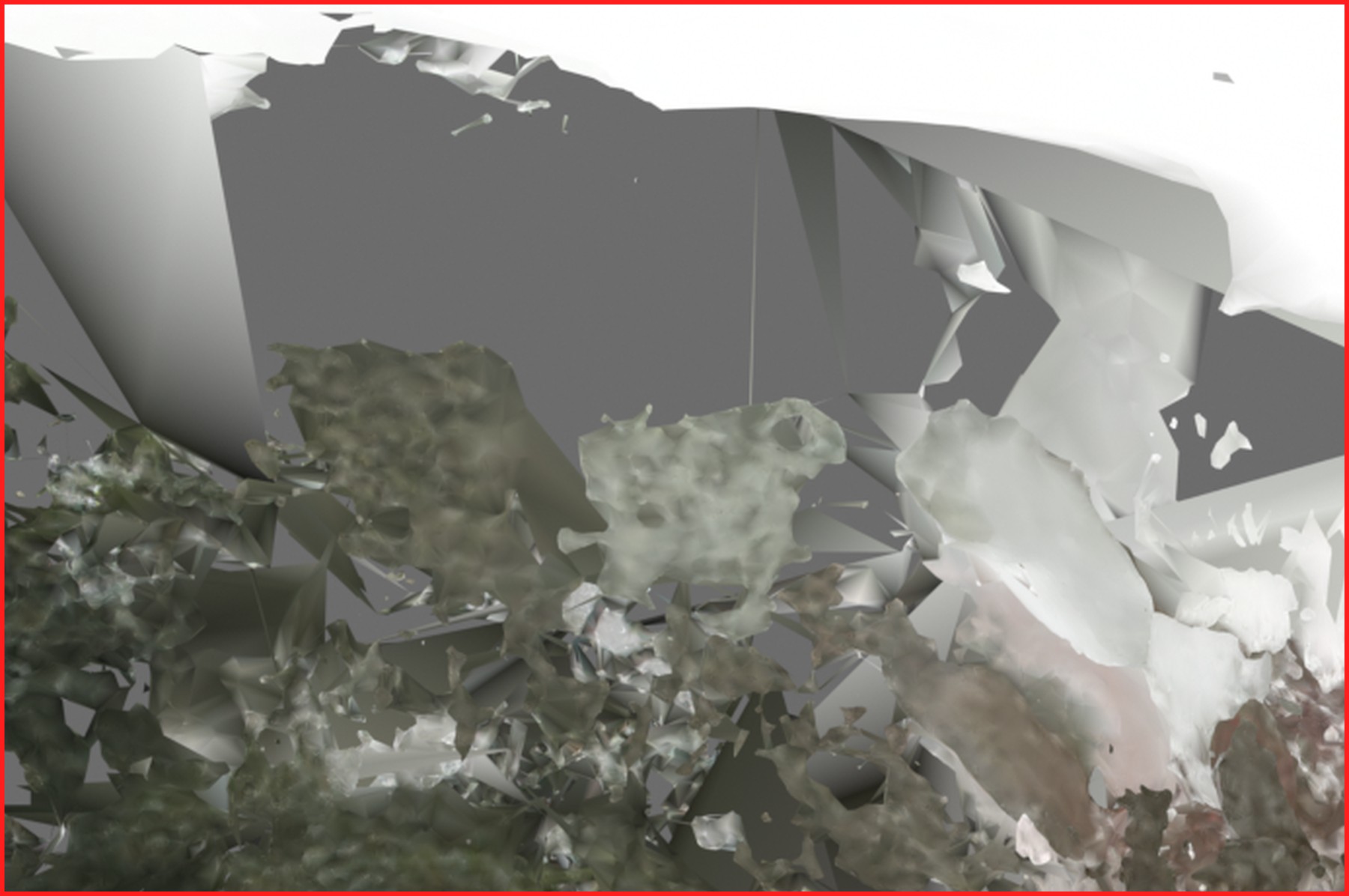} &
        \includegraphics[width=0.315\textwidth]{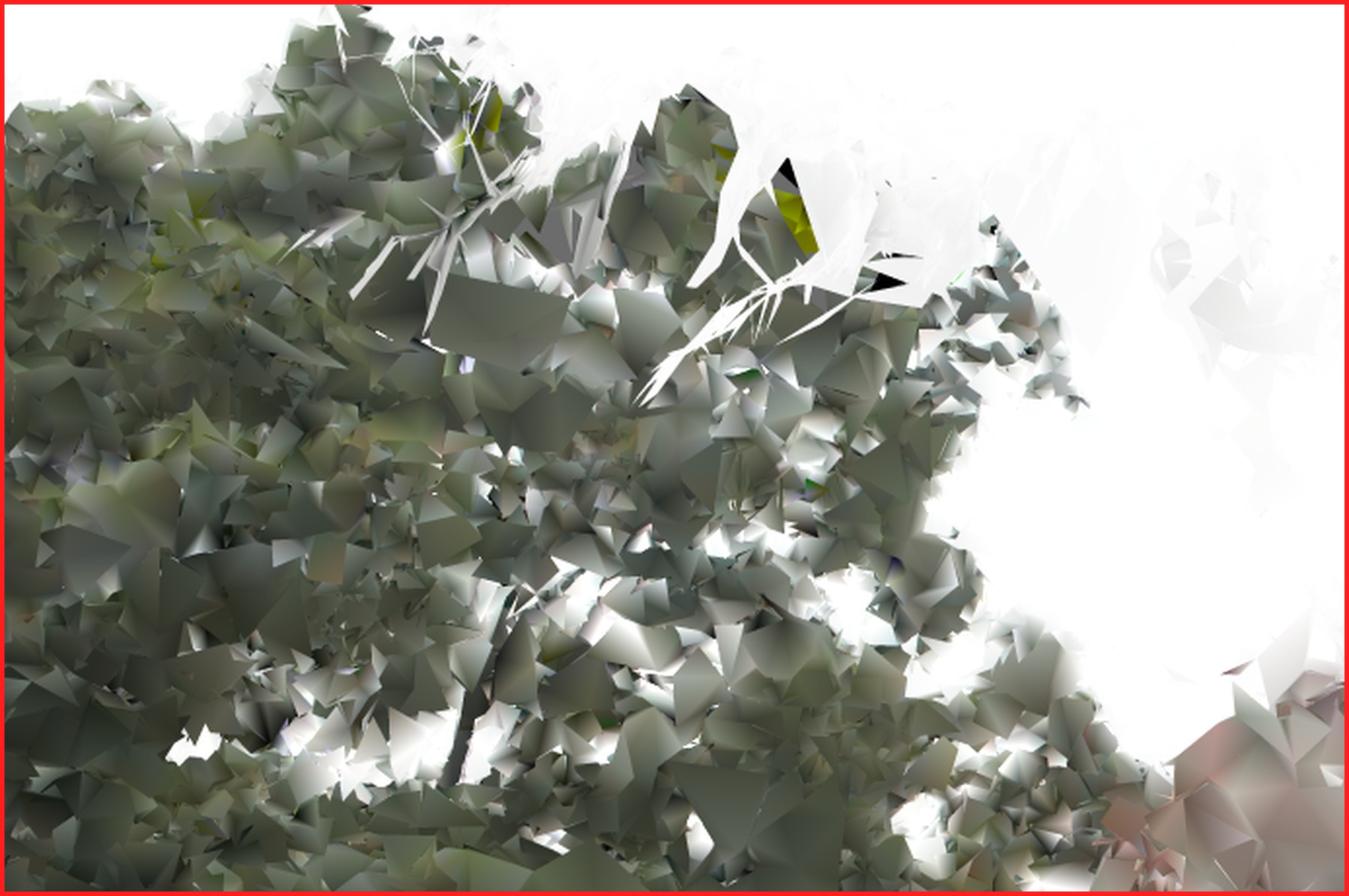} \\
        \includegraphics[width=0.315\textwidth]{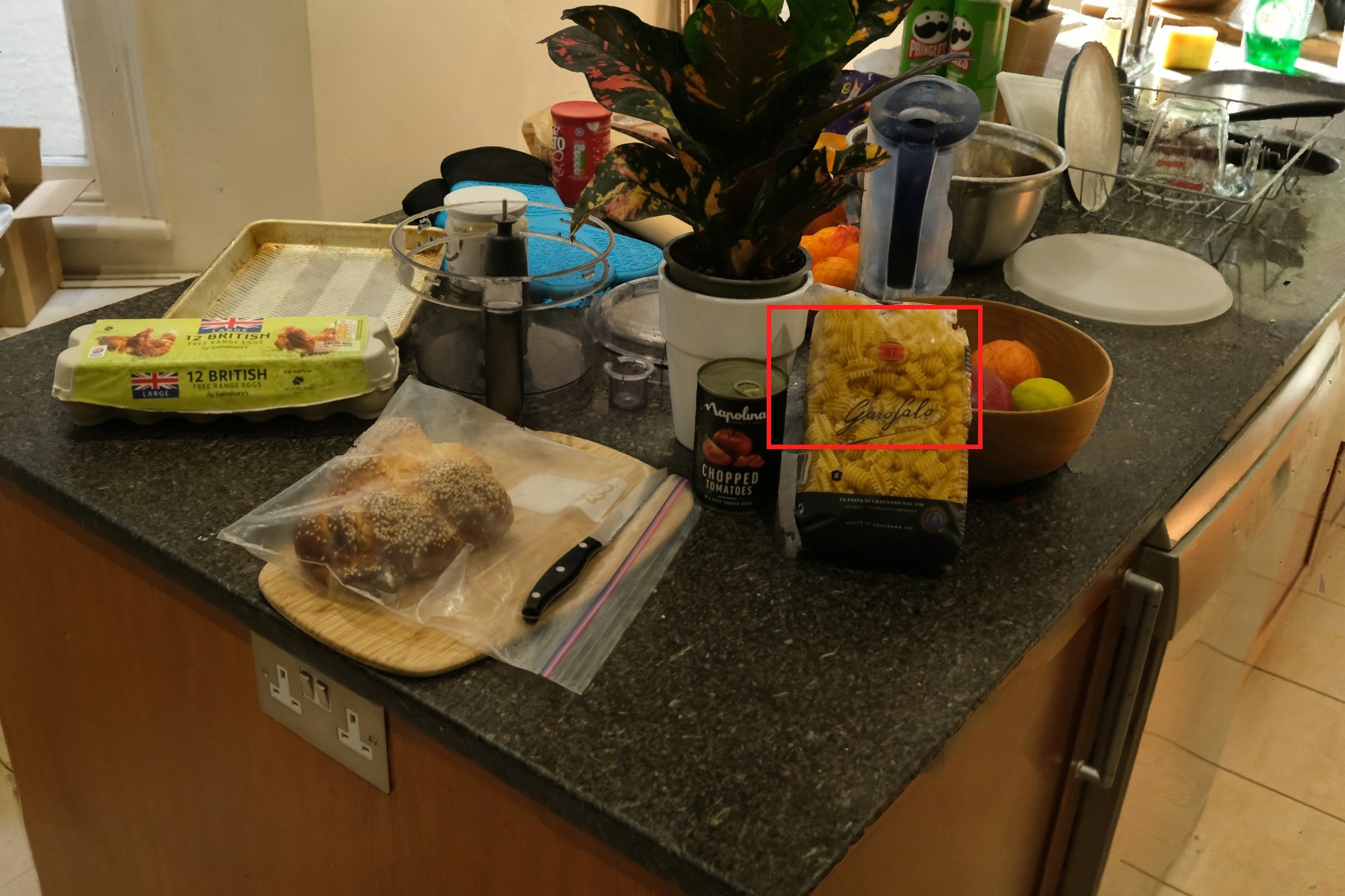} &
        \includegraphics[width=0.315\textwidth]{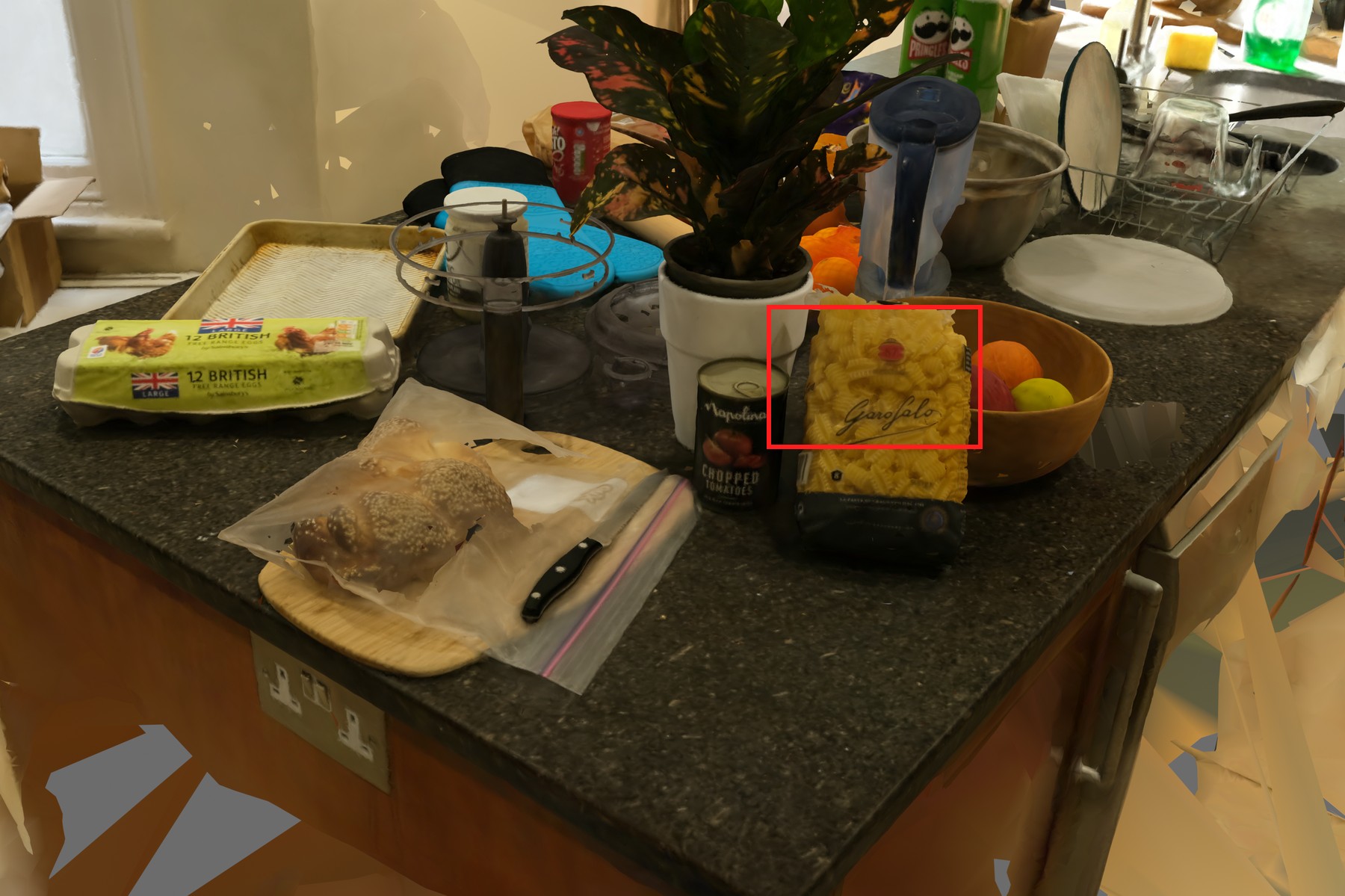} &
        \includegraphics[width=0.315\textwidth]{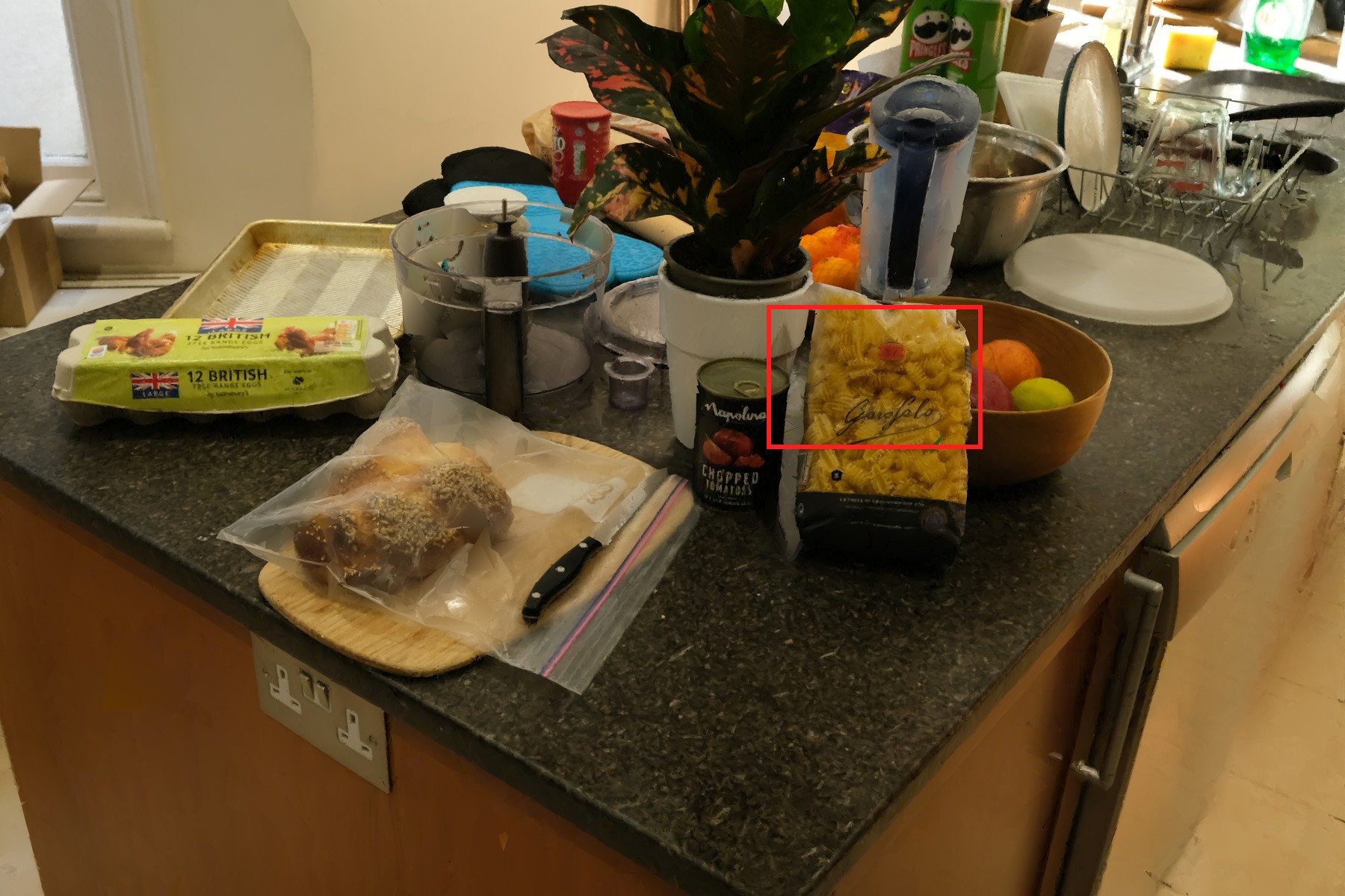} \\
        \includegraphics[width=0.315\textwidth]{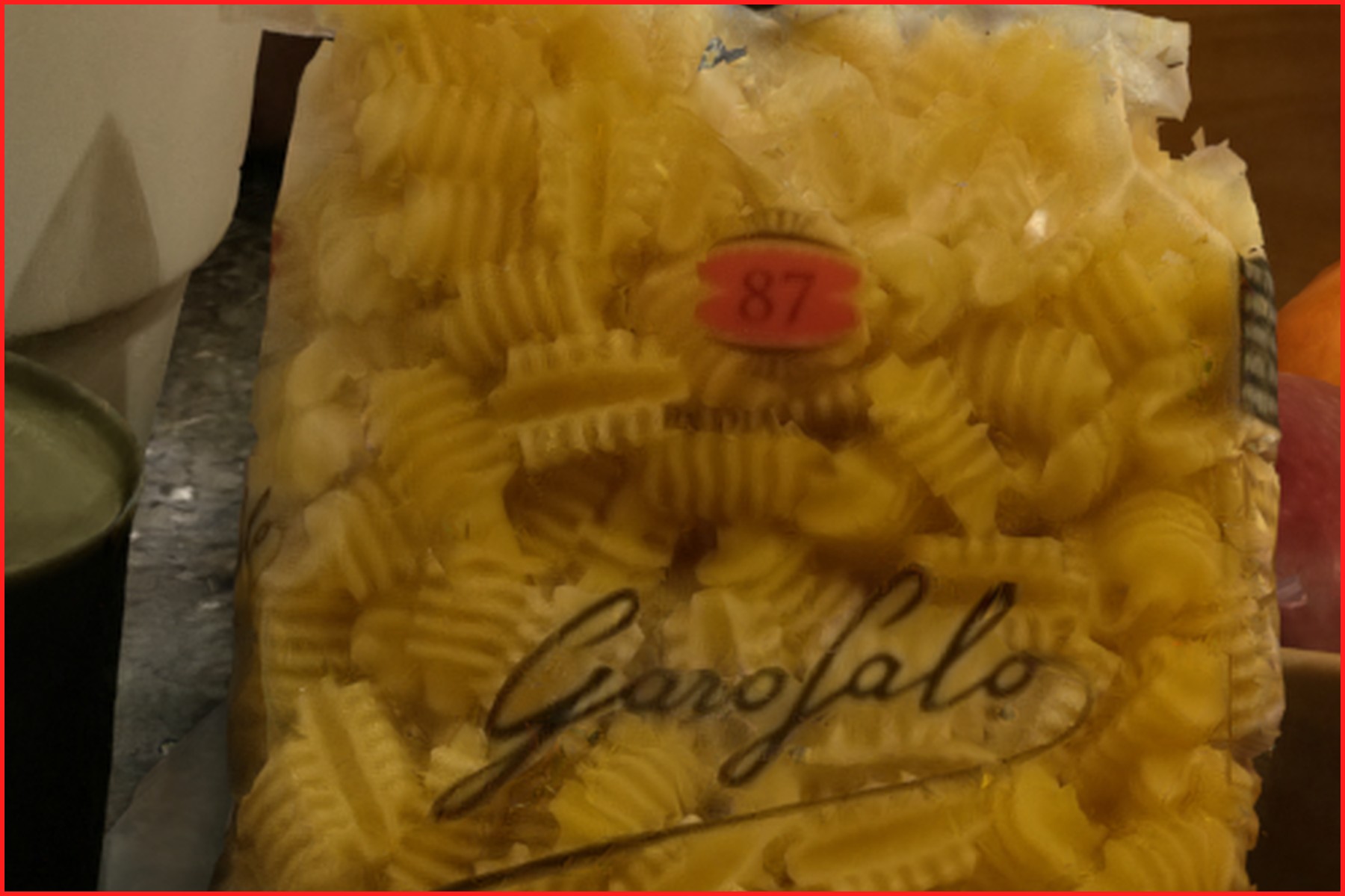} &
        \includegraphics[width=0.315\textwidth]{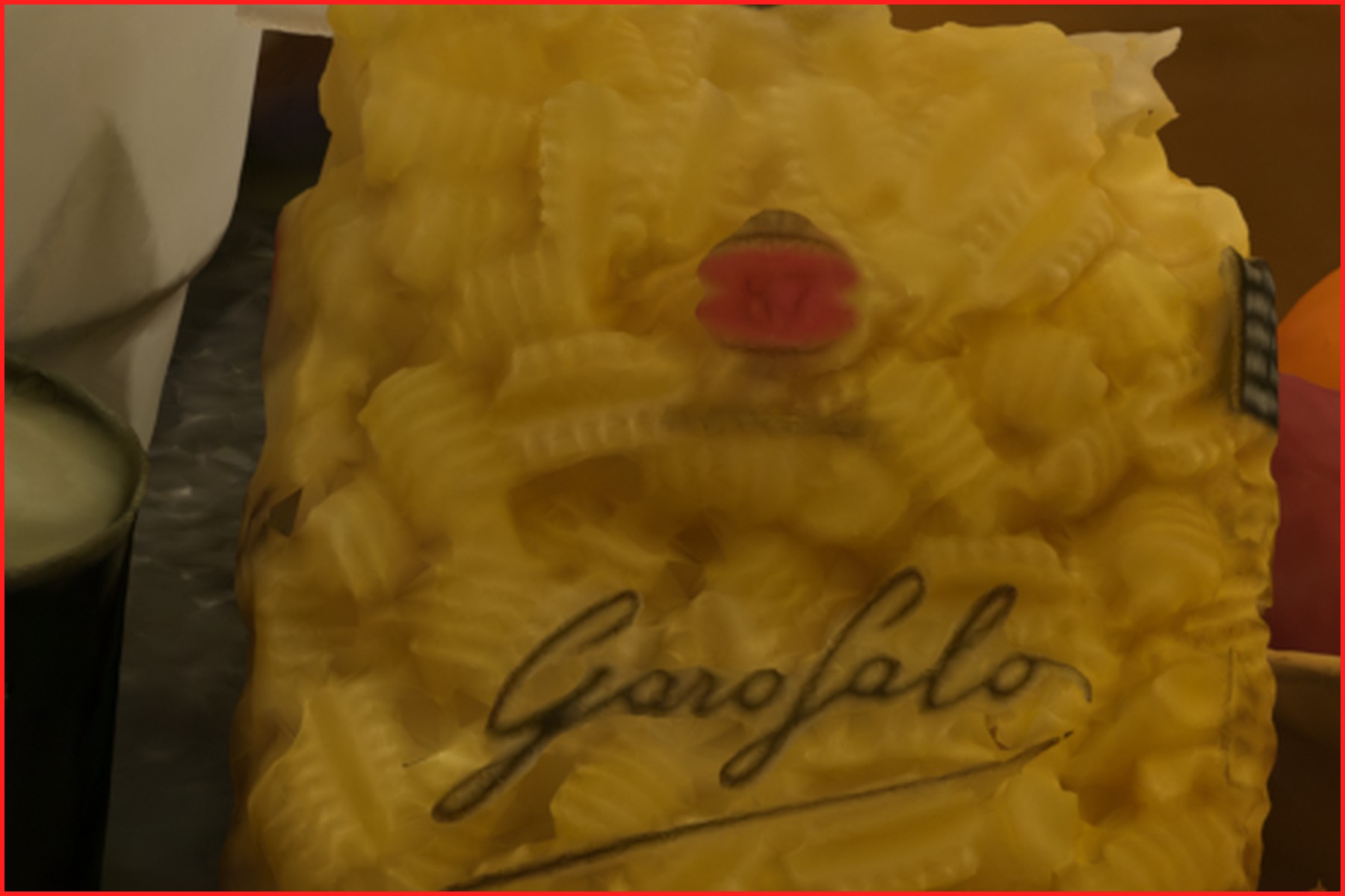} &
        \includegraphics[width=0.315\textwidth]{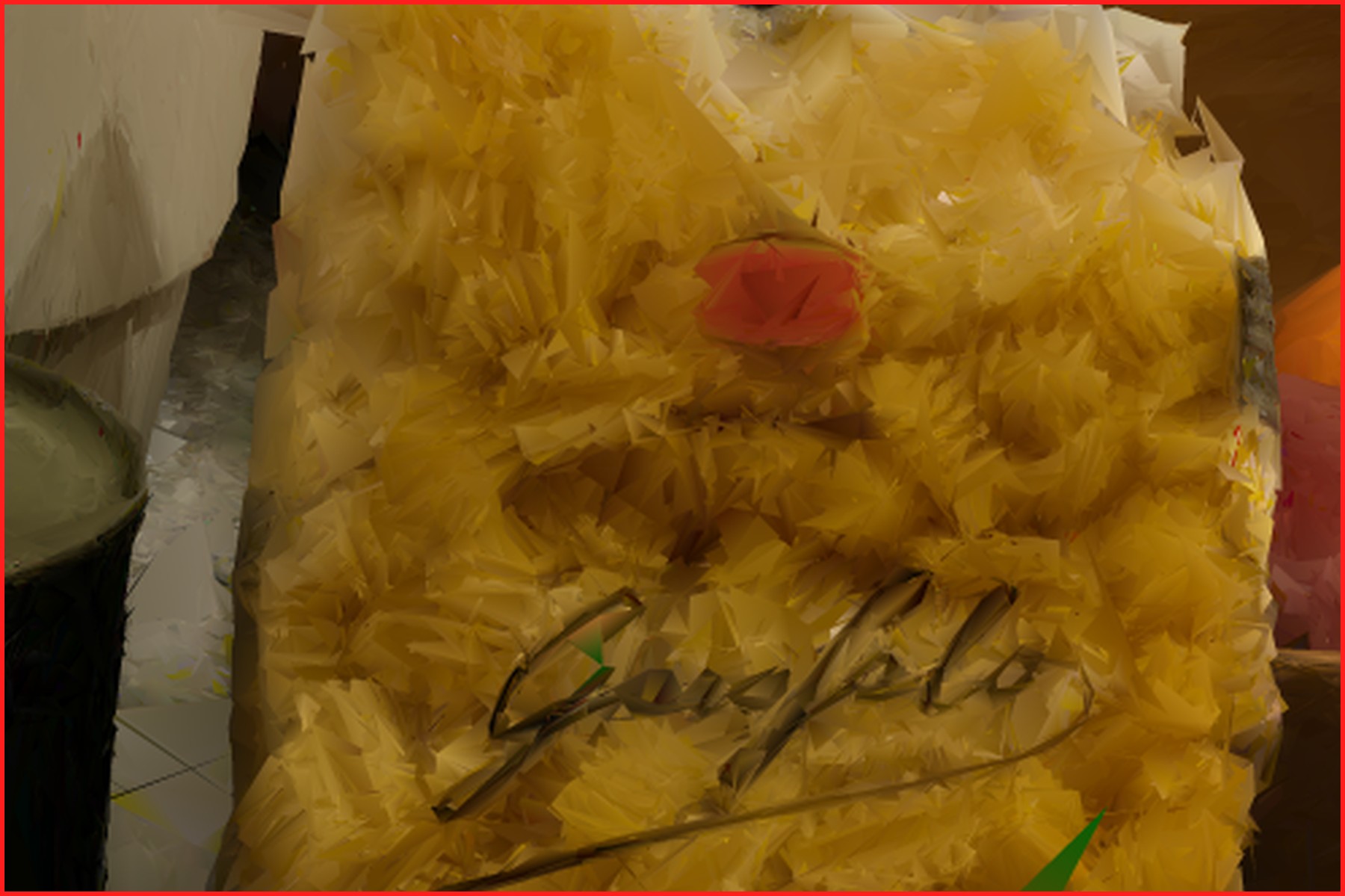} \\
    \end{tabular}
    \caption{\textbf{Qualitative comparison on Mip-NeRF 360} (\textit{bicycle}, rows 1--2; \textit{counter}, rows 3--4). For each scene, we show the textured mesh of our method, Blobs to Spokes~\cite{gomez2026gaussianwrapping}, and MeshSplatting~\cite{held2026mesh}, and below it the enlarged region marked by the red box.}
    \label{fig:suppl_overview_1}
\end{figure*}
We compare our meshes to Blobs to Spokes and MeshSplatting in Fig.~\ref{fig:suppl_overview_2}.
\begin{figure*}[tbp]
    \centering
    \setlength{\tabcolsep}{1pt}
    \begin{tabular}{@{}ccc@{}}
        \small Ours & \small Blobs to Spokes & \small MeshSplatting \\
        \includegraphics[width=0.315\textwidth]{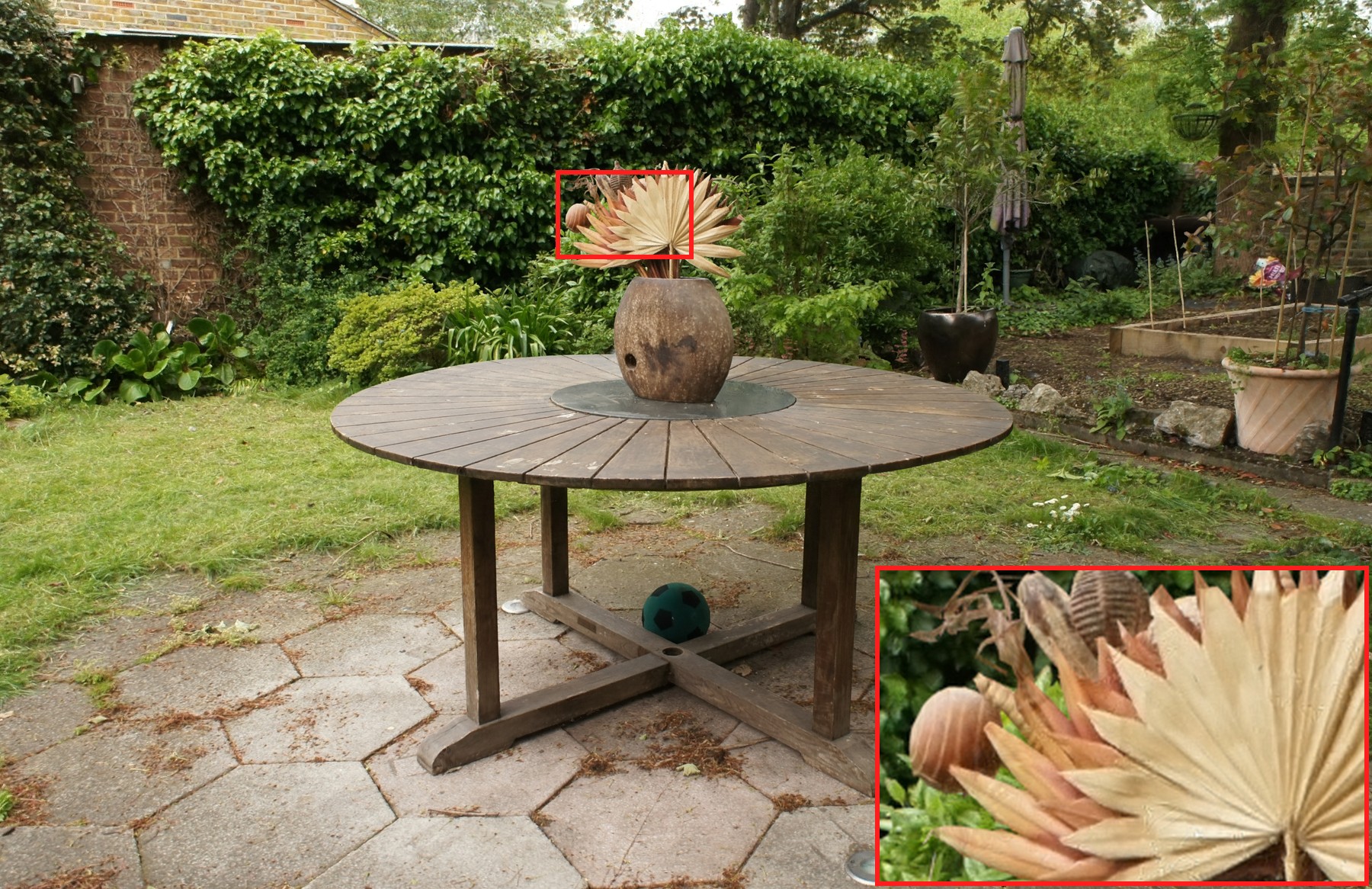} &
        \includegraphics[width=0.315\textwidth]{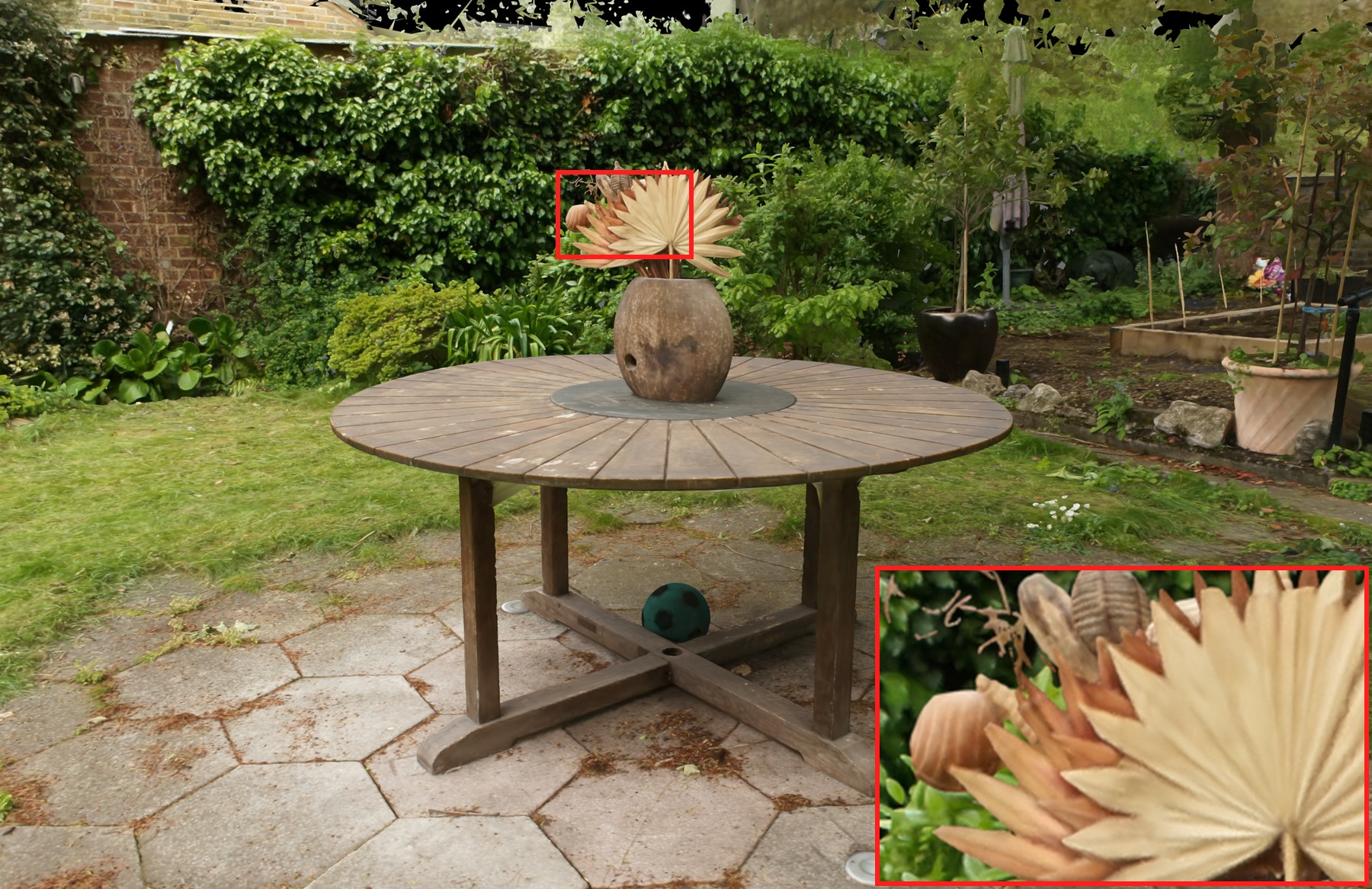} &
        \includegraphics[width=0.315\textwidth]{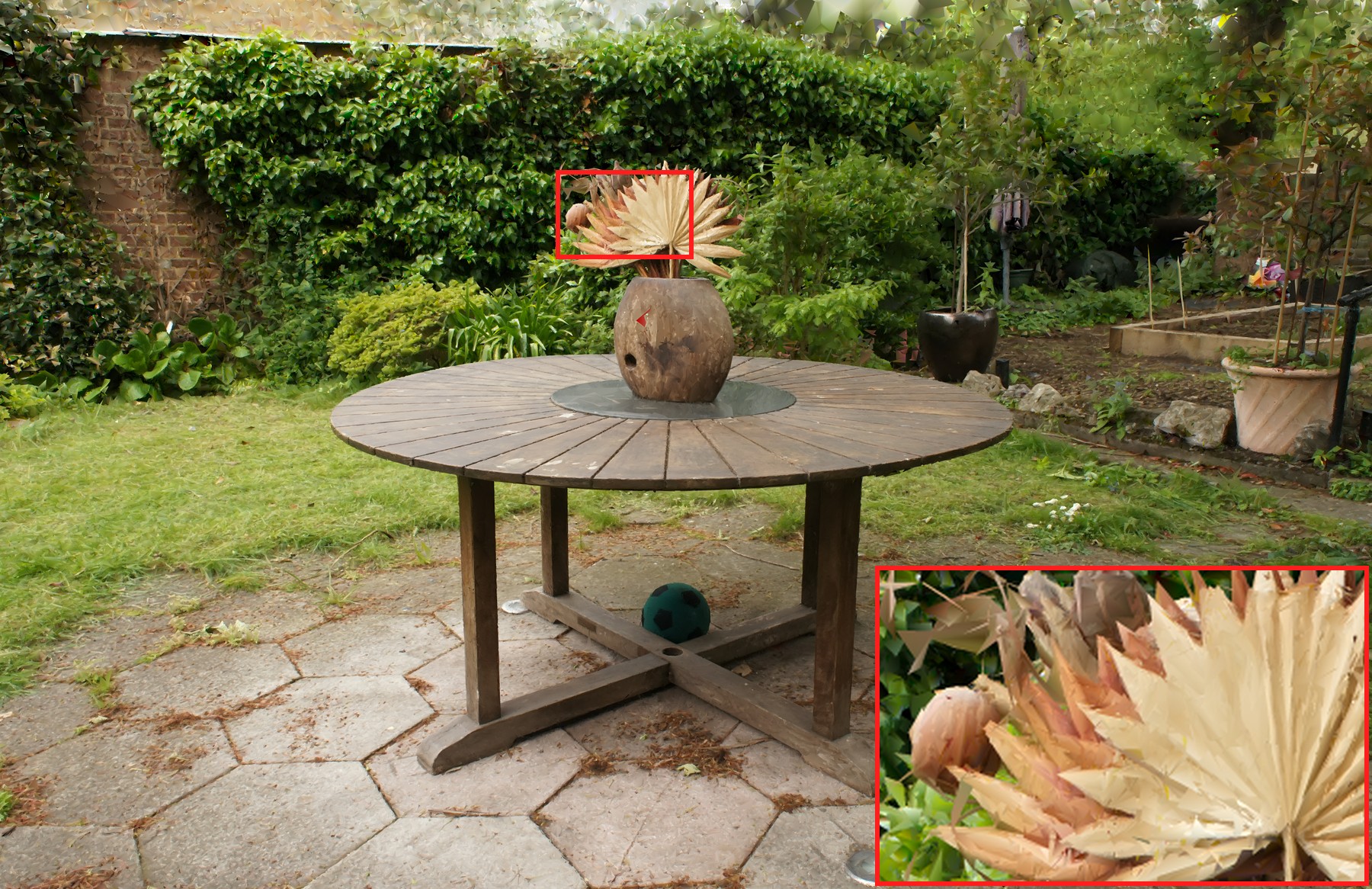} \\
        \includegraphics[width=0.315\textwidth]{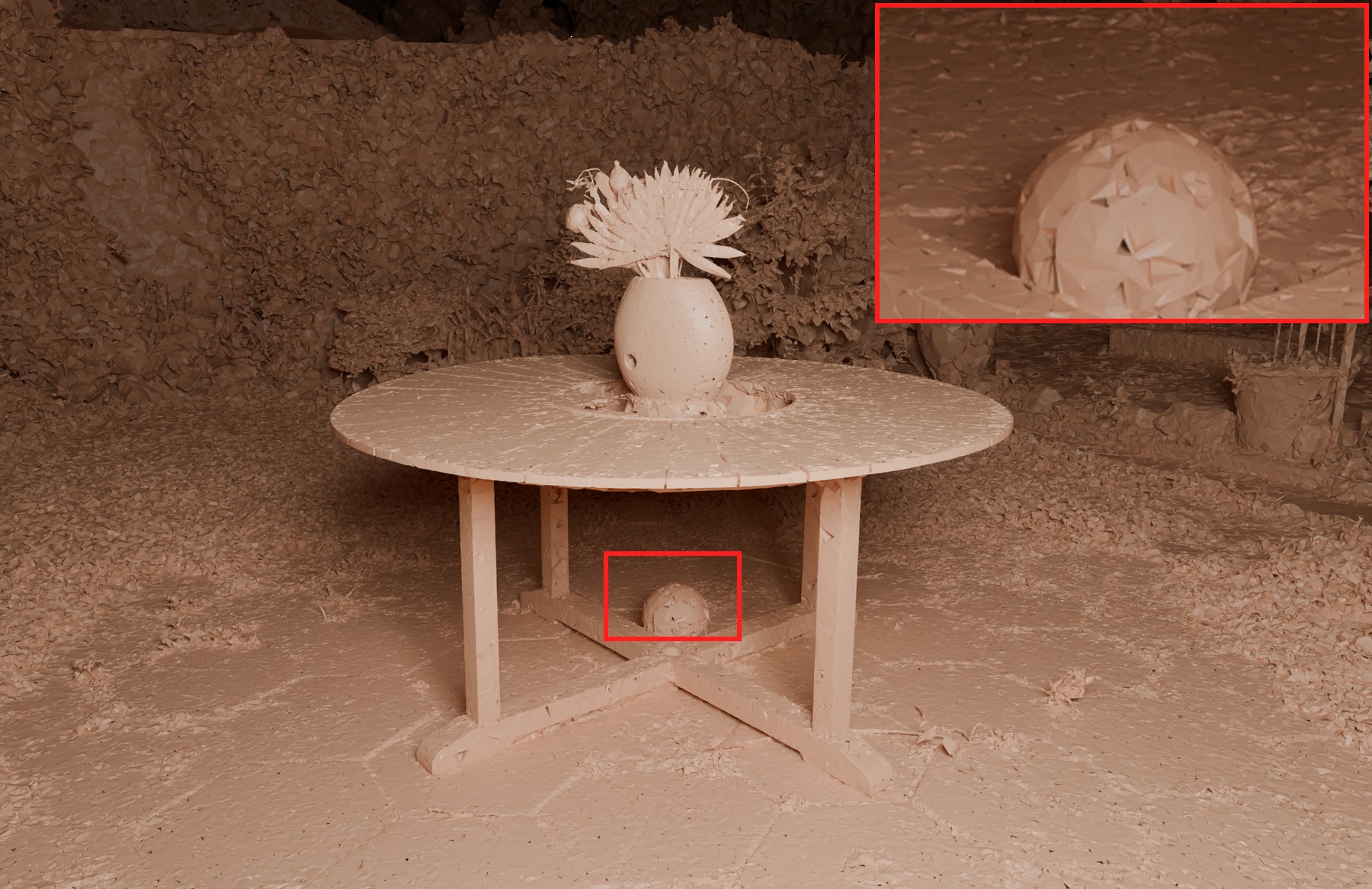} &
        \includegraphics[width=0.315\textwidth]{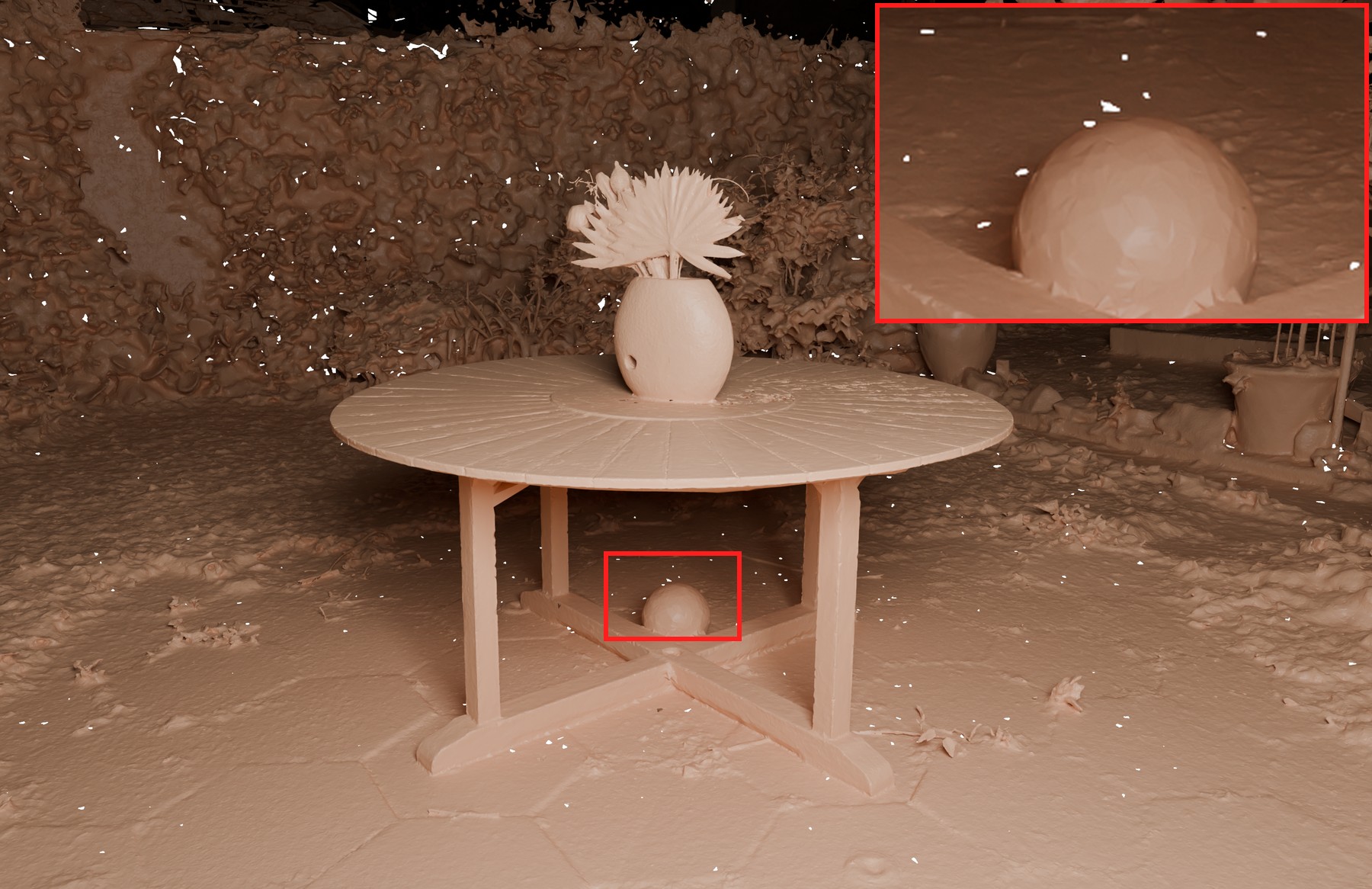} &
        \includegraphics[width=0.315\textwidth]{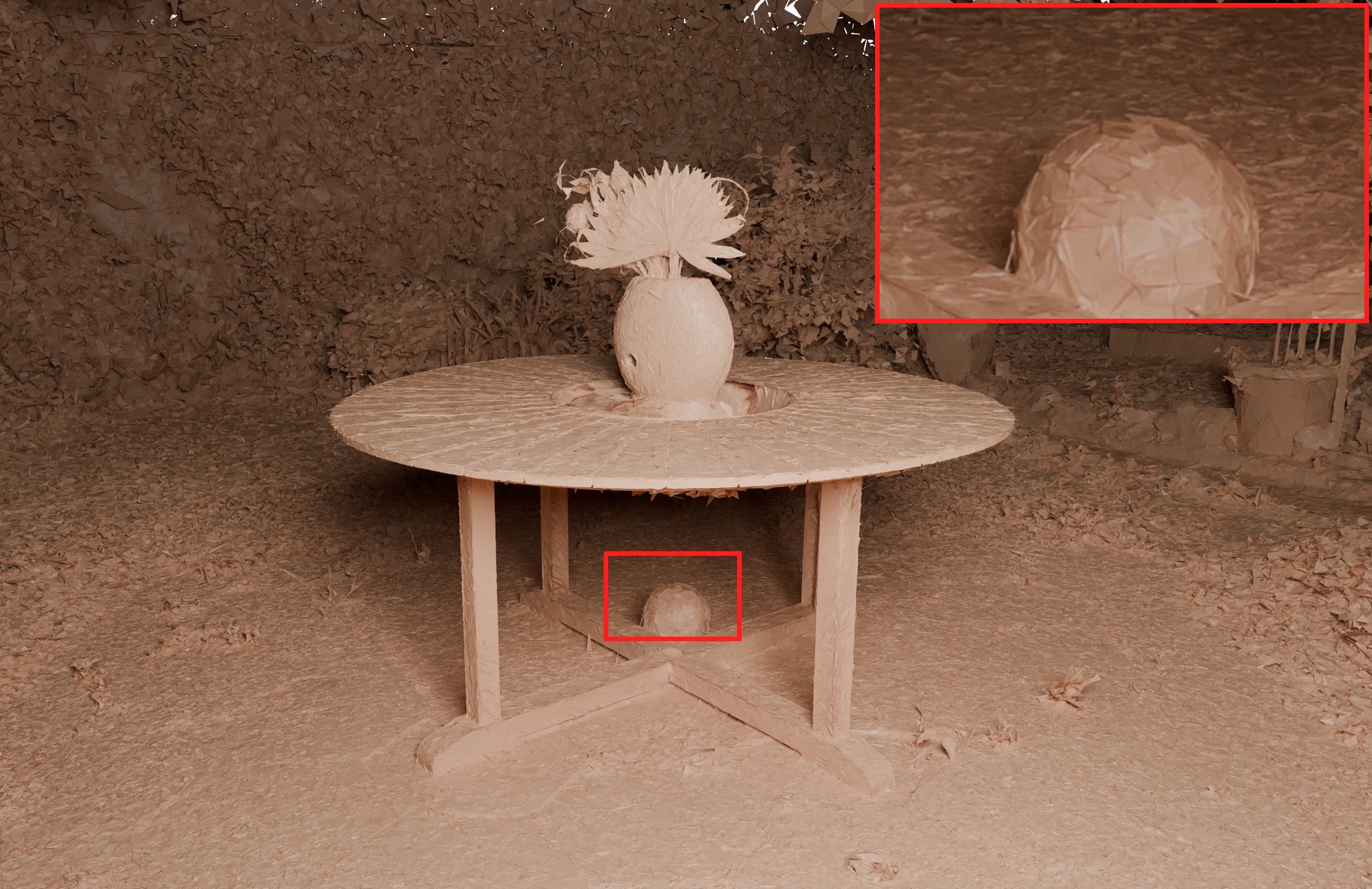} \\
        \includegraphics[width=0.315\textwidth]{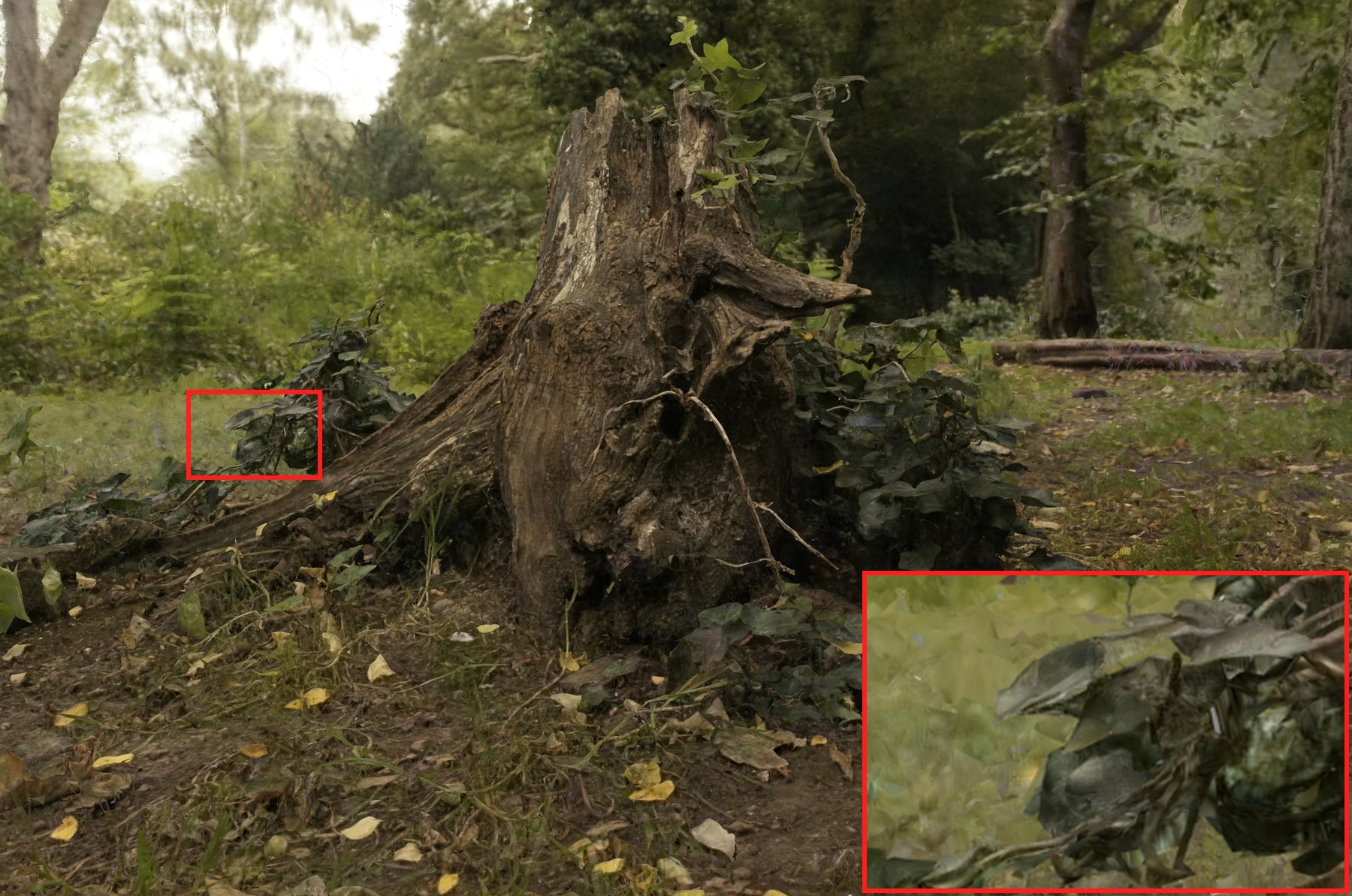} &
        \includegraphics[width=0.315\textwidth]{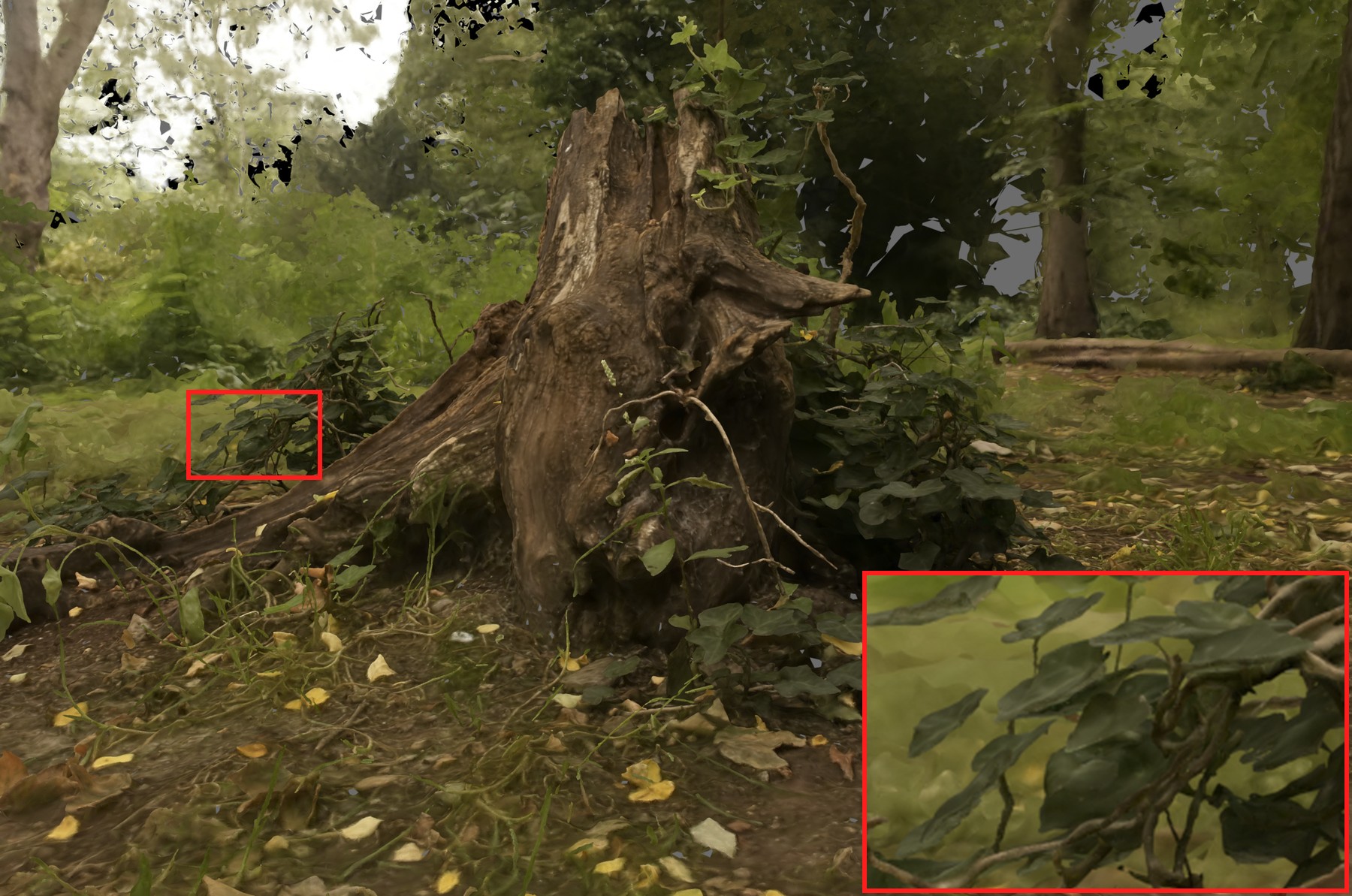} &
        \includegraphics[width=0.315\textwidth]{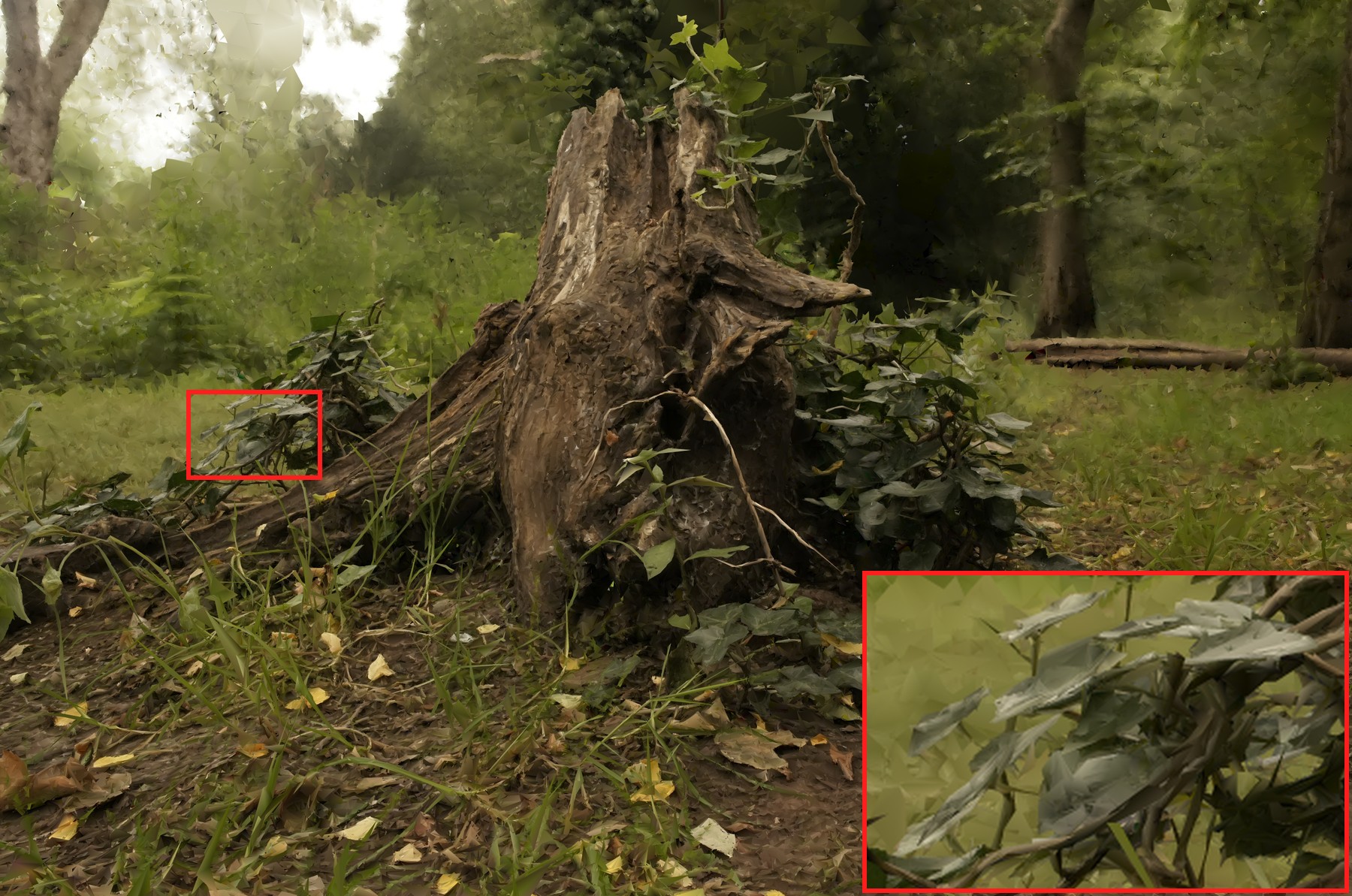} \\
        \includegraphics[width=0.315\textwidth]{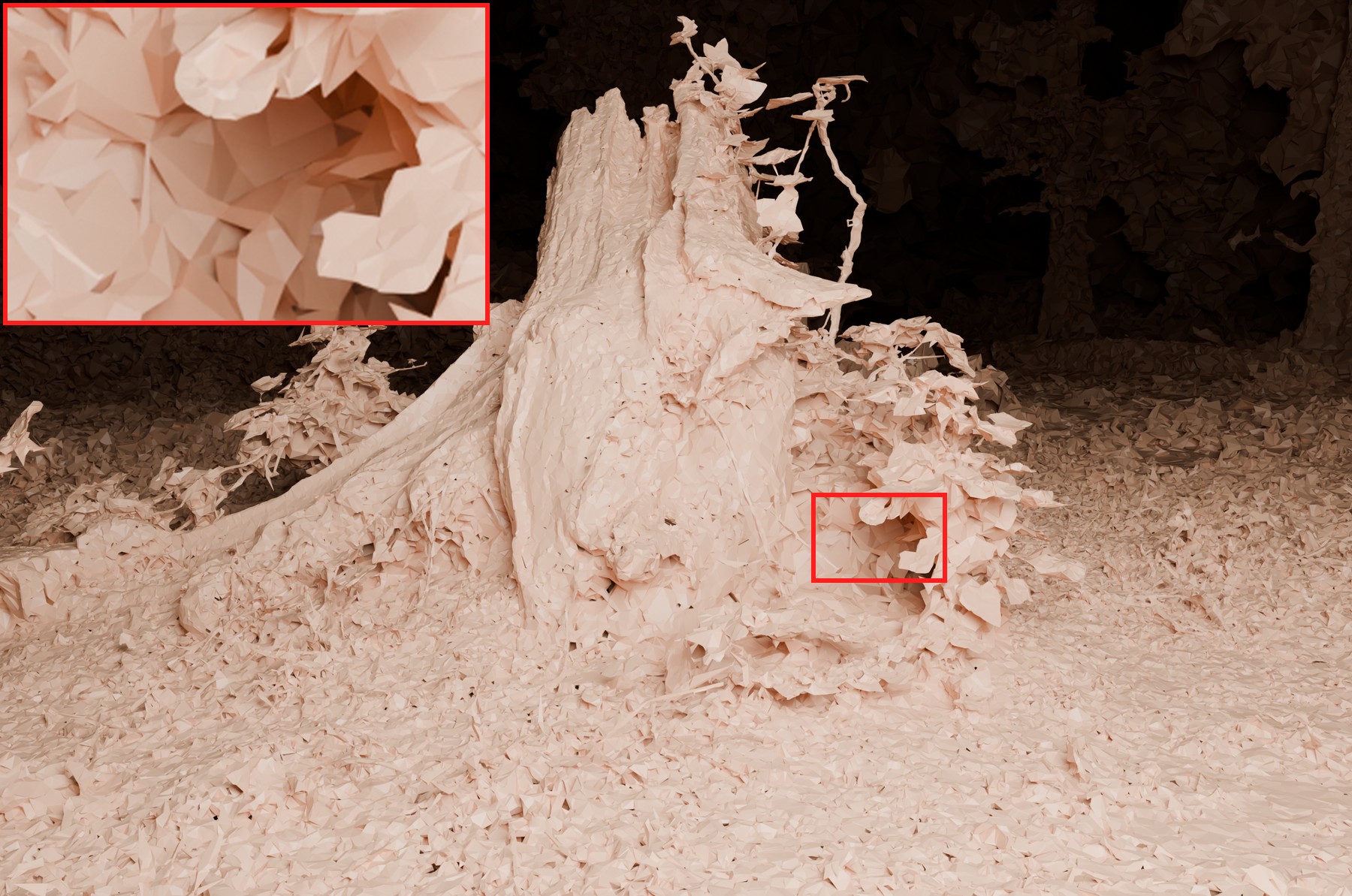} &
        \includegraphics[width=0.315\textwidth]{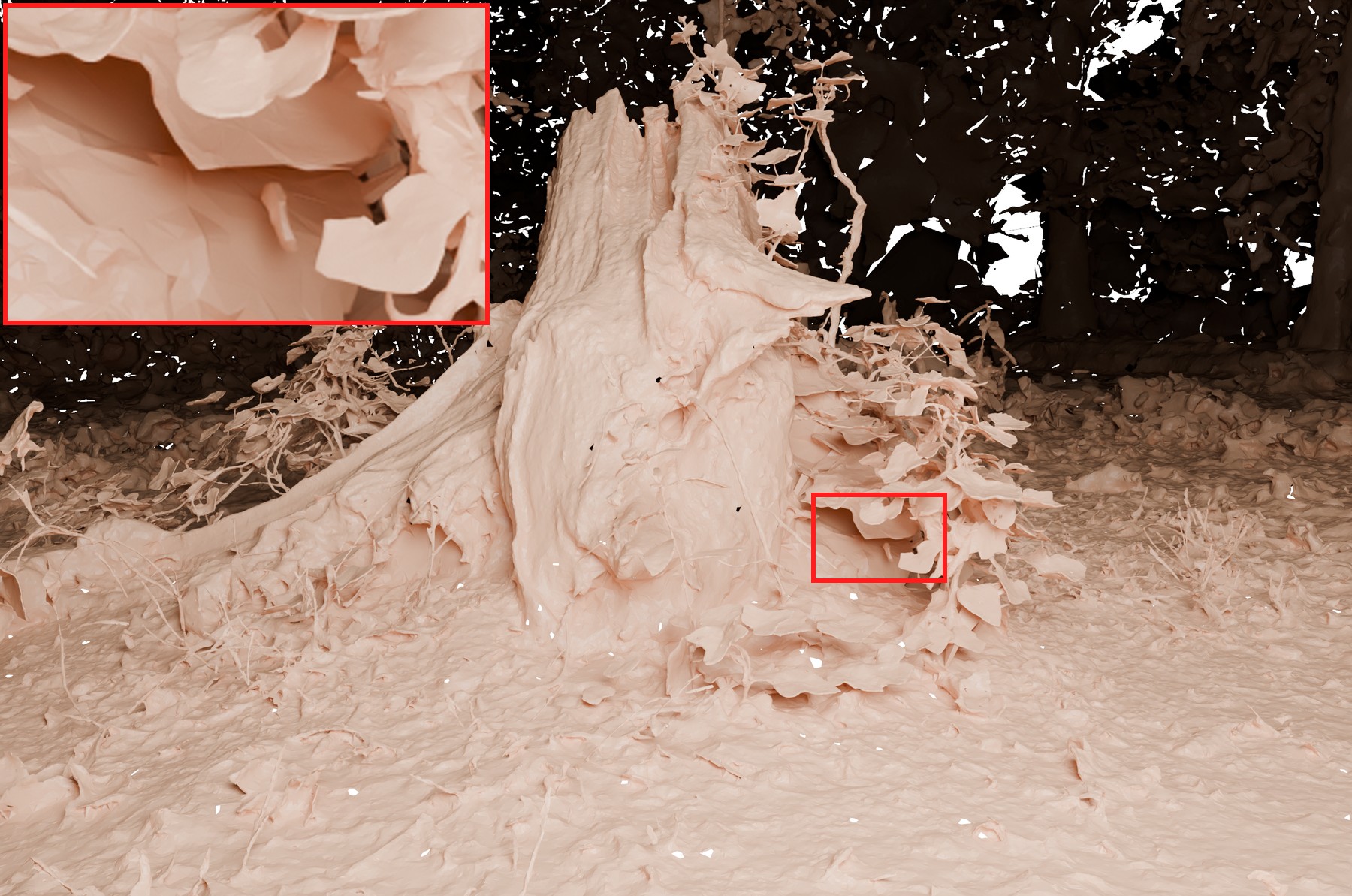} &
        \includegraphics[width=0.315\textwidth]{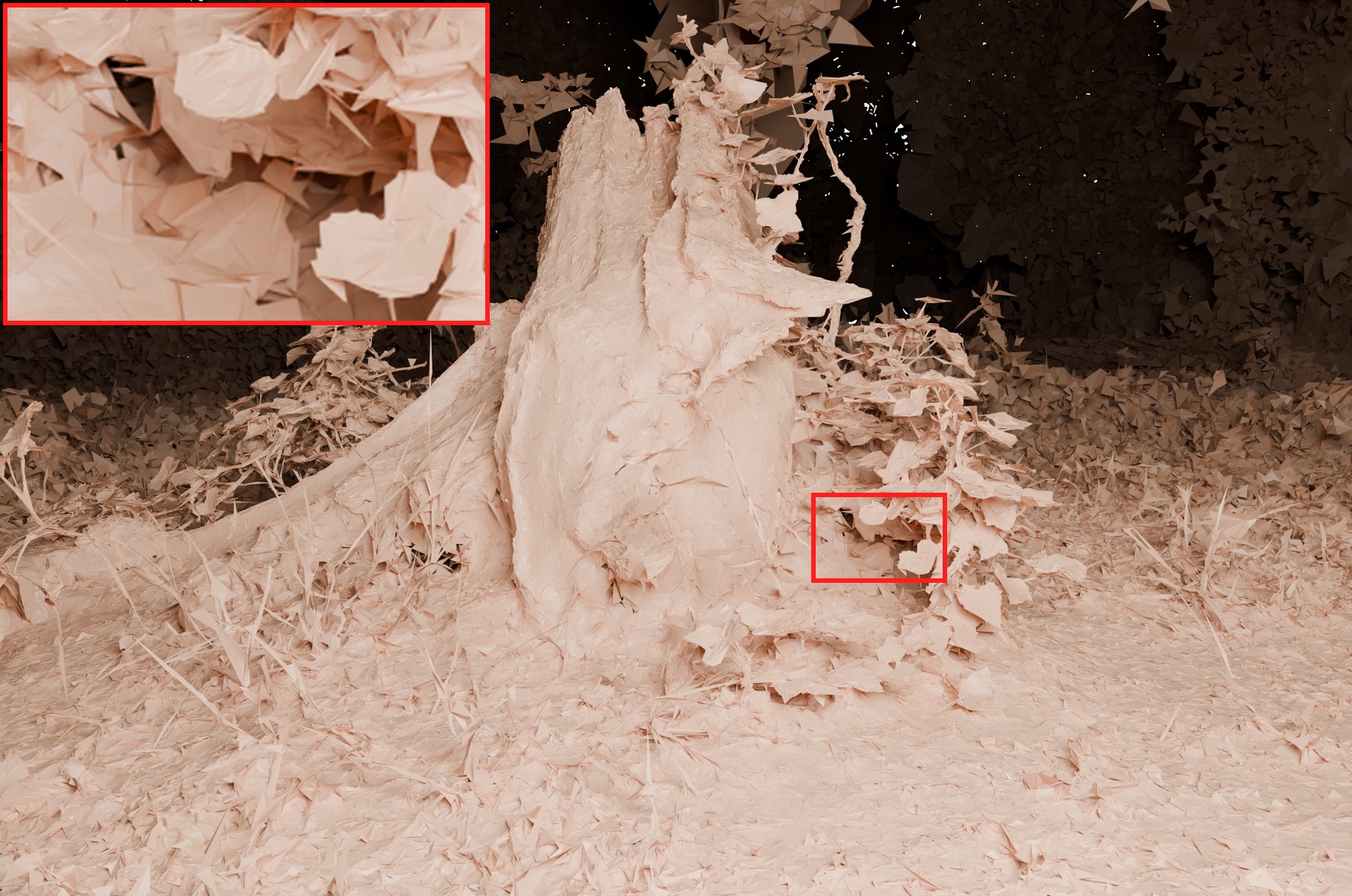} \\
    \end{tabular}
    \caption{\textbf{Qualitative comparison on Mip-NeRF 360} (\textit{garden}, rows 1--2; \textit{stump}, rows 3--4). For each scene, we show the textured mesh (rows 1 and 3) and the untextured mesh (rows 2 and 4) of our method, Blobs to Spokes~\cite{gomez2026gaussianwrapping}, and MeshSplatting~\cite{held2026mesh}. The red box marks the region shown enlarged in the inset of each image.}
    \label{fig:suppl_overview_2}
\end{figure*}

Fig.~\ref{fig:suppl_dtu_all} shows the textured mesh and the mesh geometry for all 15 DTU scans.
\begin{figure*}[t]
    \centering
    \setlength{\tabcolsep}{1pt}
    \renewcommand{\arraystretch}{0.9}
    \begin{tabular}{ccccc}
        \small 24 & \small 37 & \small 40 & \small 55 & \small 63 \\
        \includegraphics[width=0.19\textwidth]{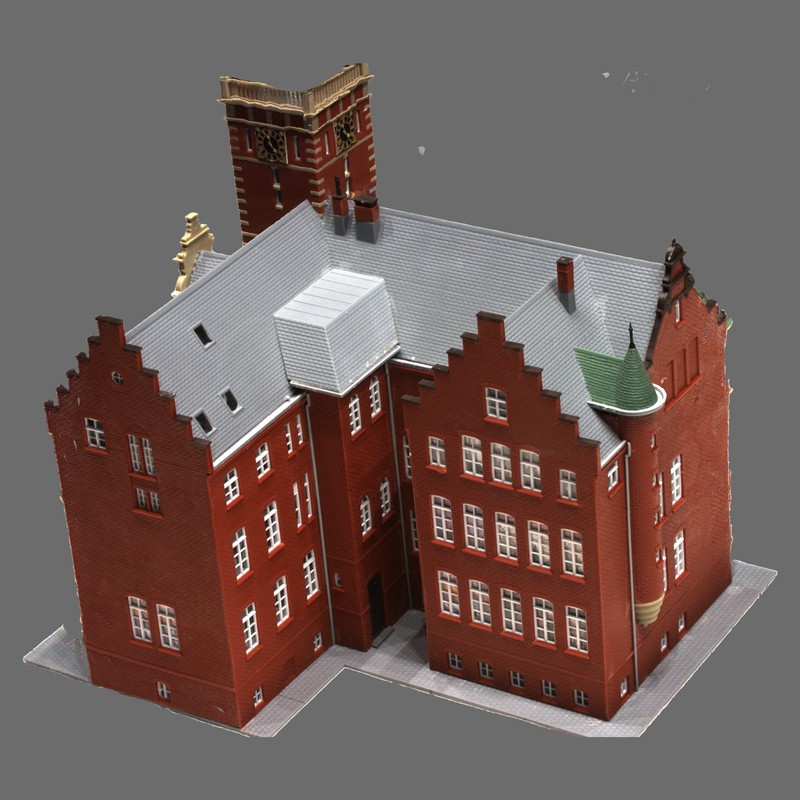} & \includegraphics[width=0.19\textwidth]{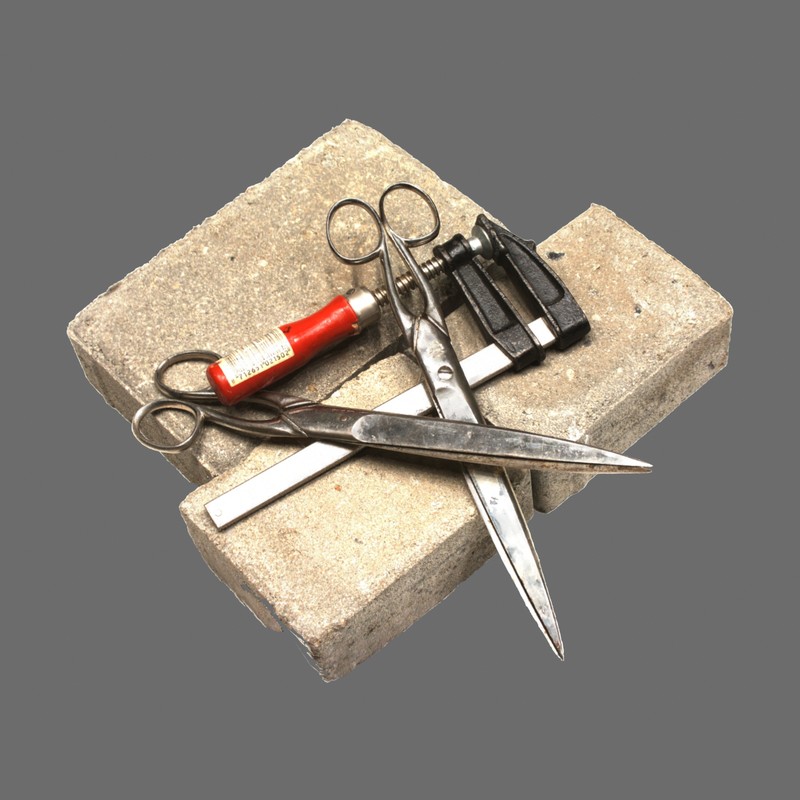} & \includegraphics[width=0.19\textwidth]{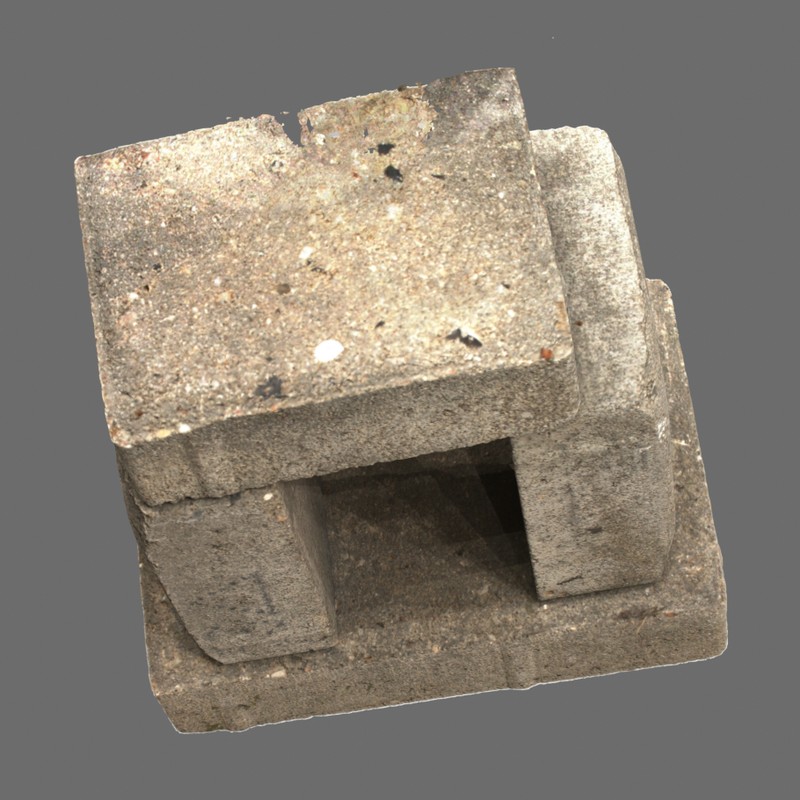} & \includegraphics[width=0.19\textwidth]{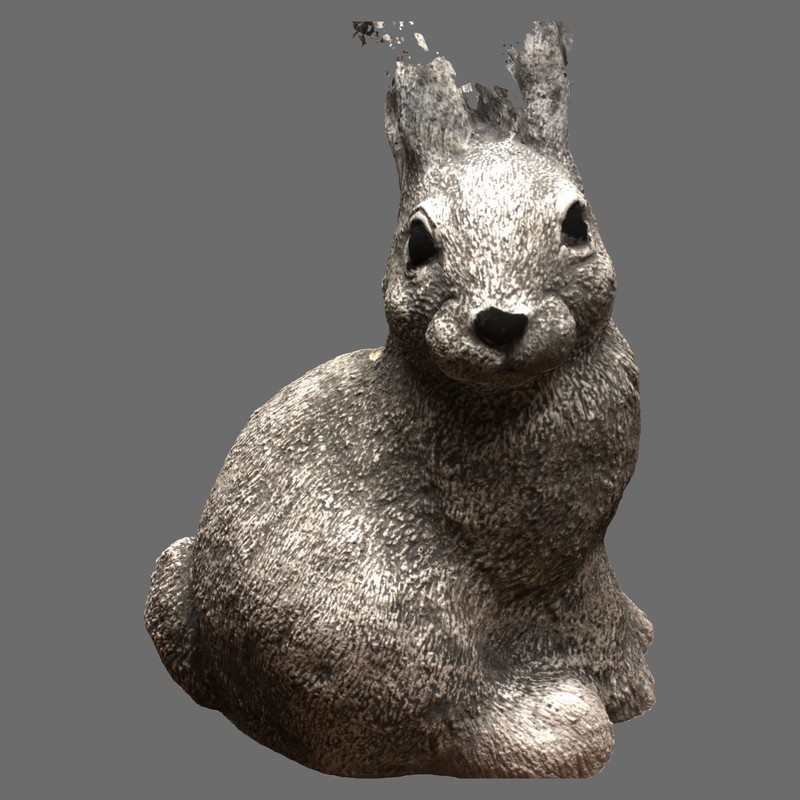} & \includegraphics[width=0.19\textwidth]{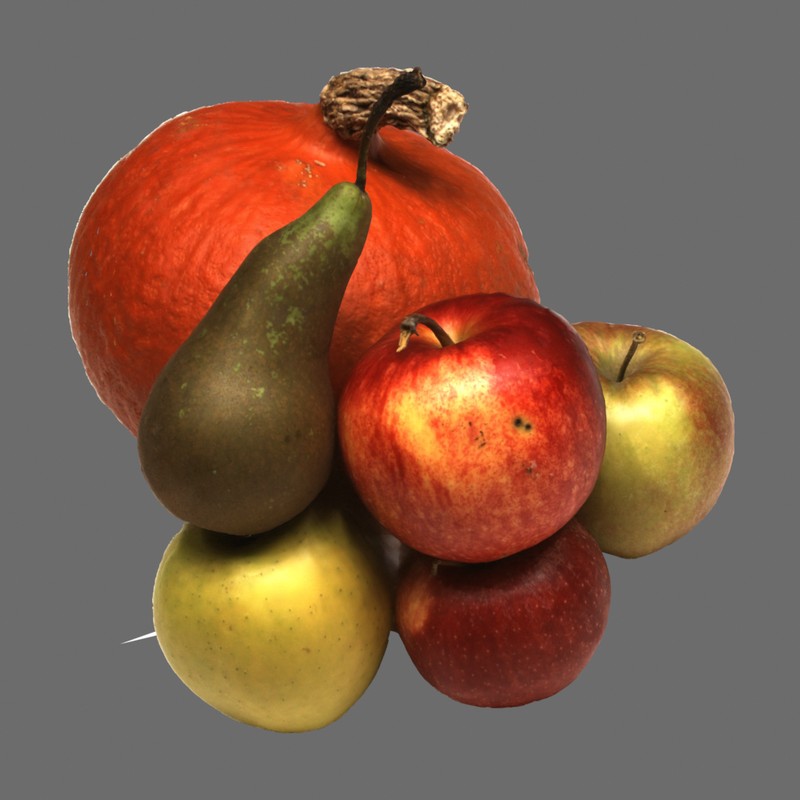} \\
        \includegraphics[width=0.19\textwidth]{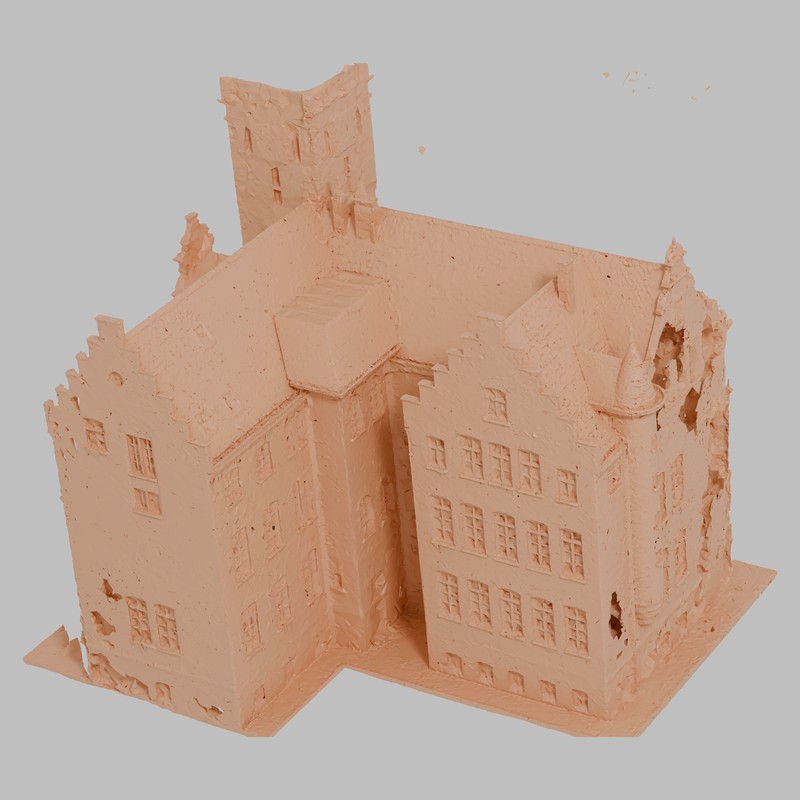} & \includegraphics[width=0.19\textwidth]{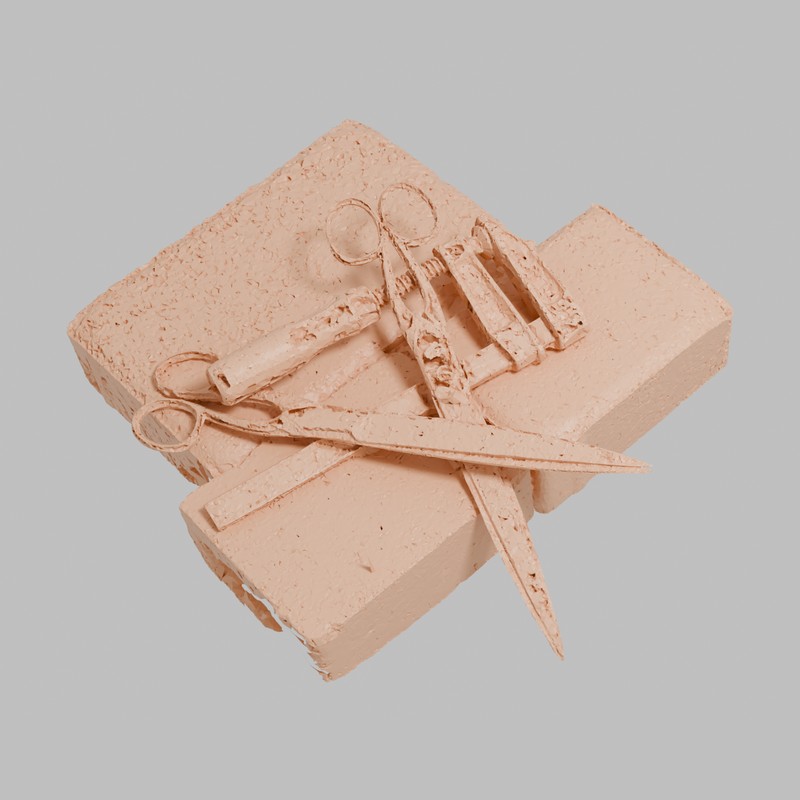} & \includegraphics[width=0.19\textwidth]{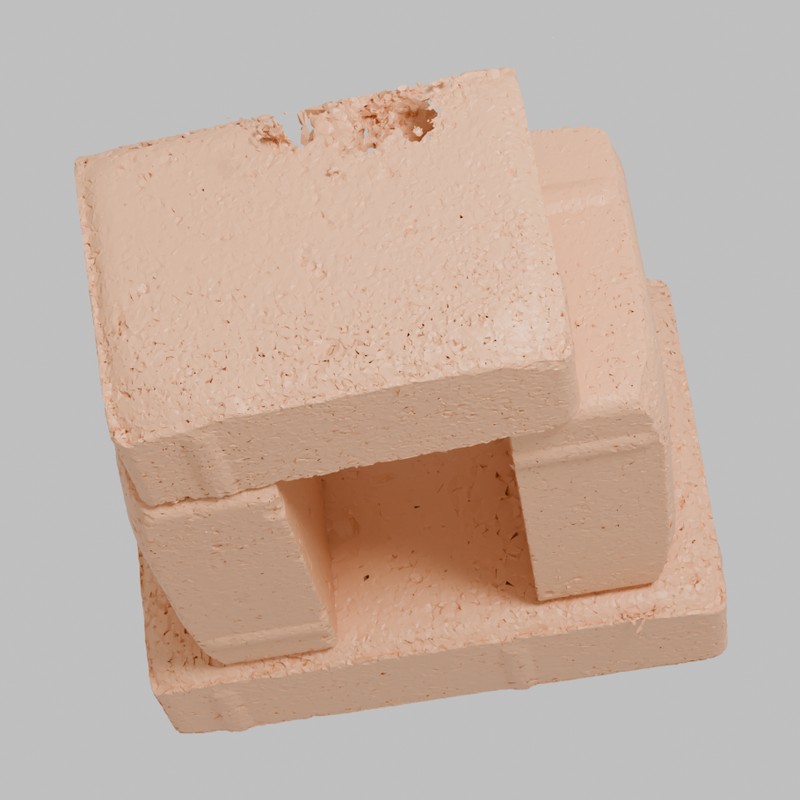} & \includegraphics[width=0.19\textwidth]{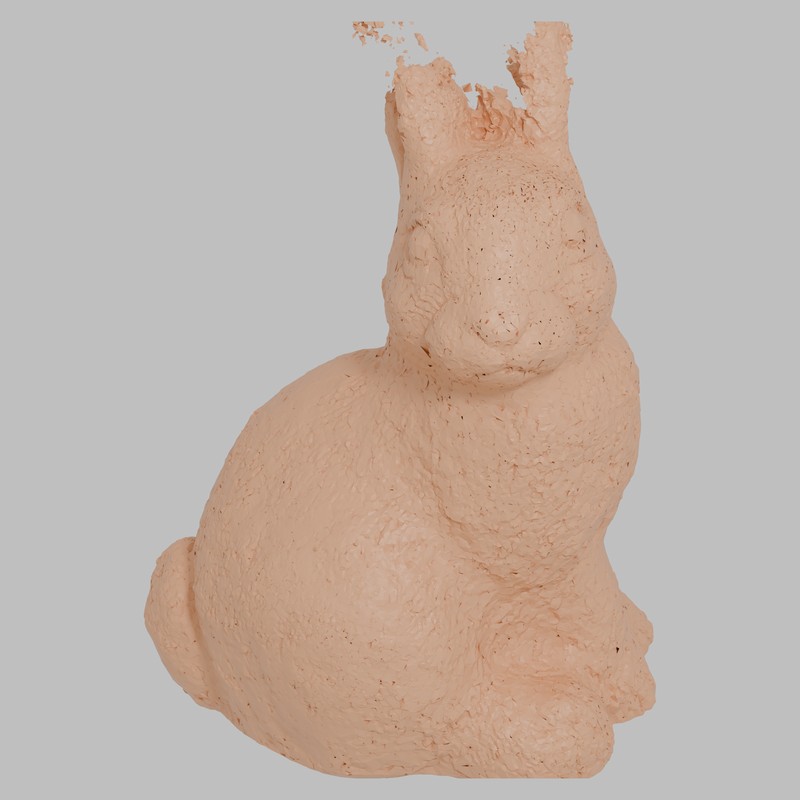} & \includegraphics[width=0.19\textwidth]{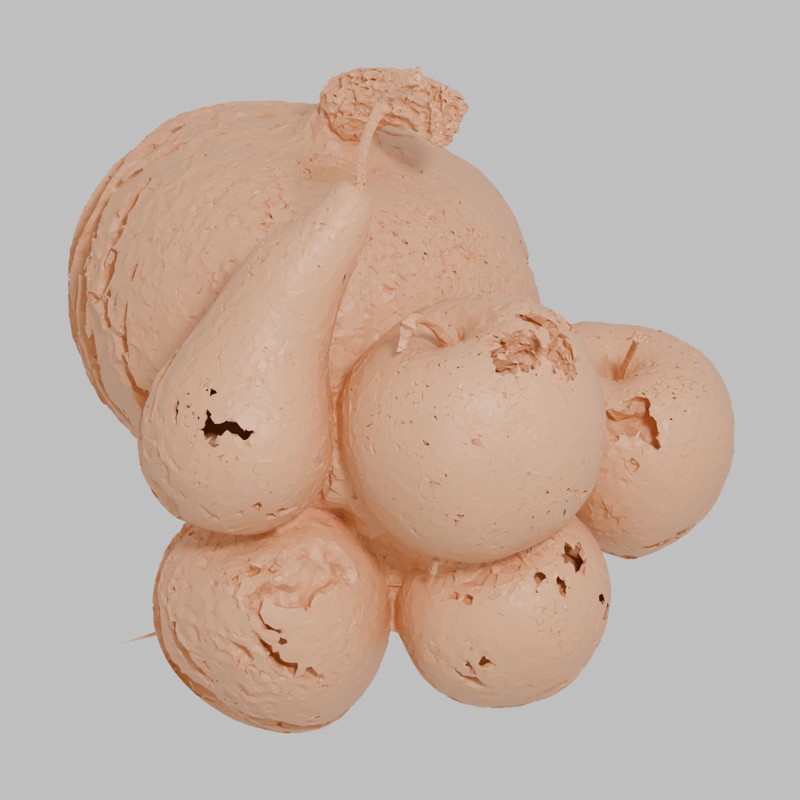} \\[4pt]
        \small 65 & \small 69 & \small 83 & \small 97 & \small 105 \\
        \includegraphics[width=0.19\textwidth]{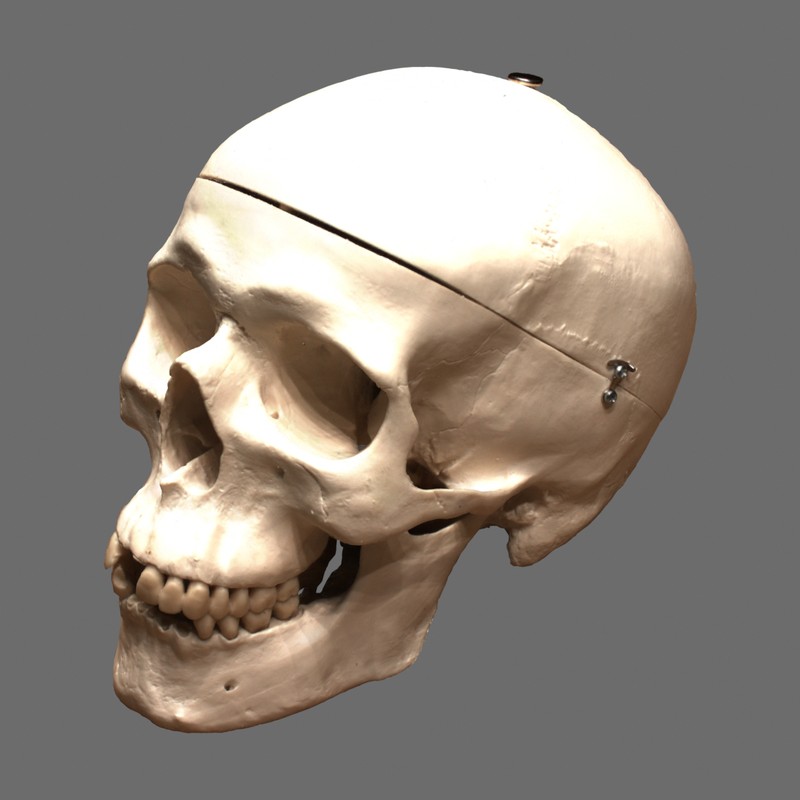} & \includegraphics[width=0.19\textwidth]{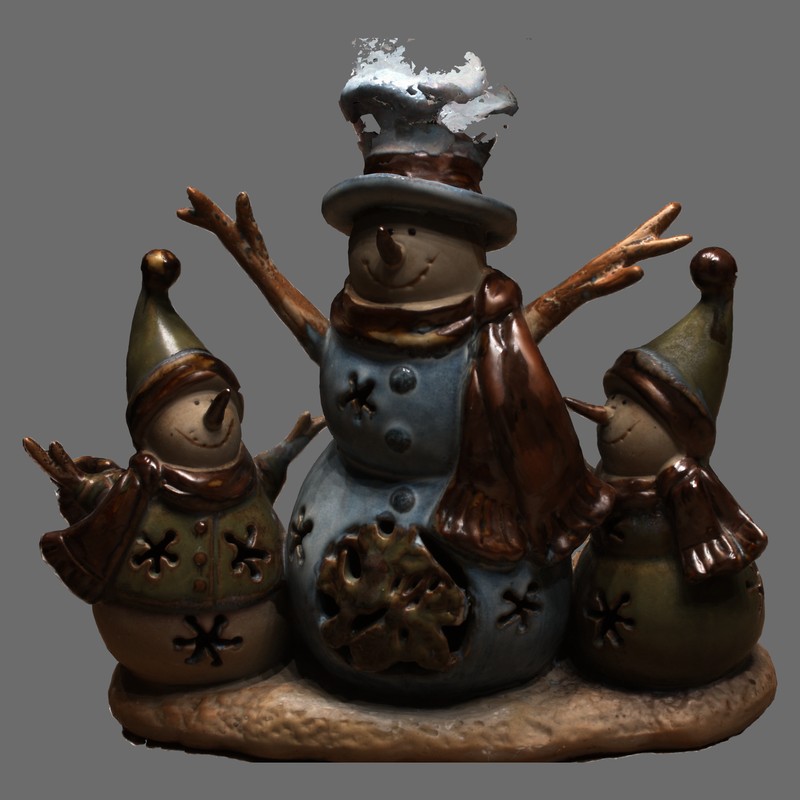} & \includegraphics[width=0.19\textwidth]{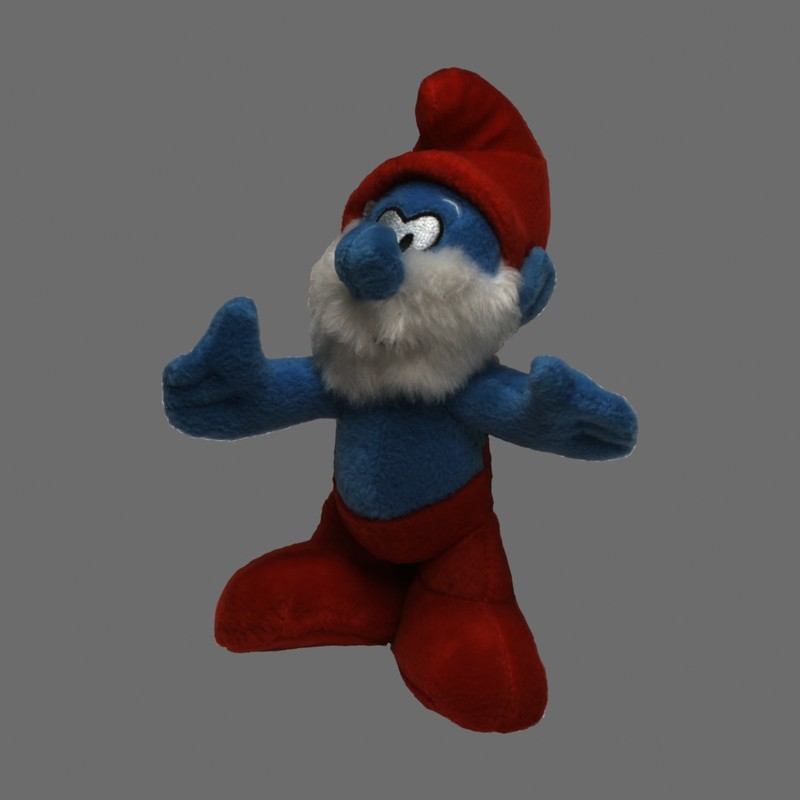} & \includegraphics[width=0.19\textwidth]{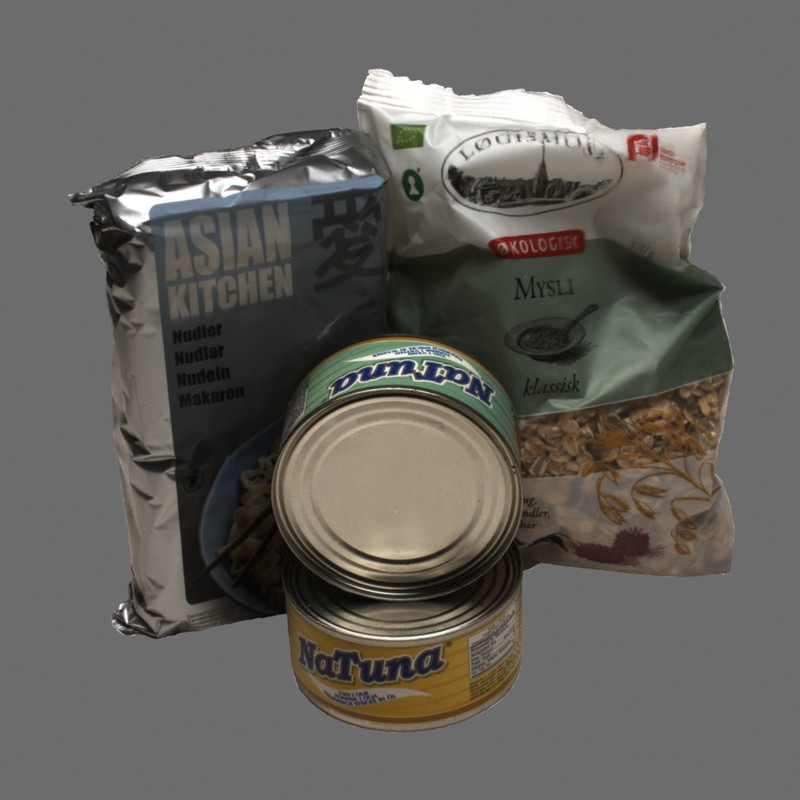} & \includegraphics[width=0.19\textwidth]{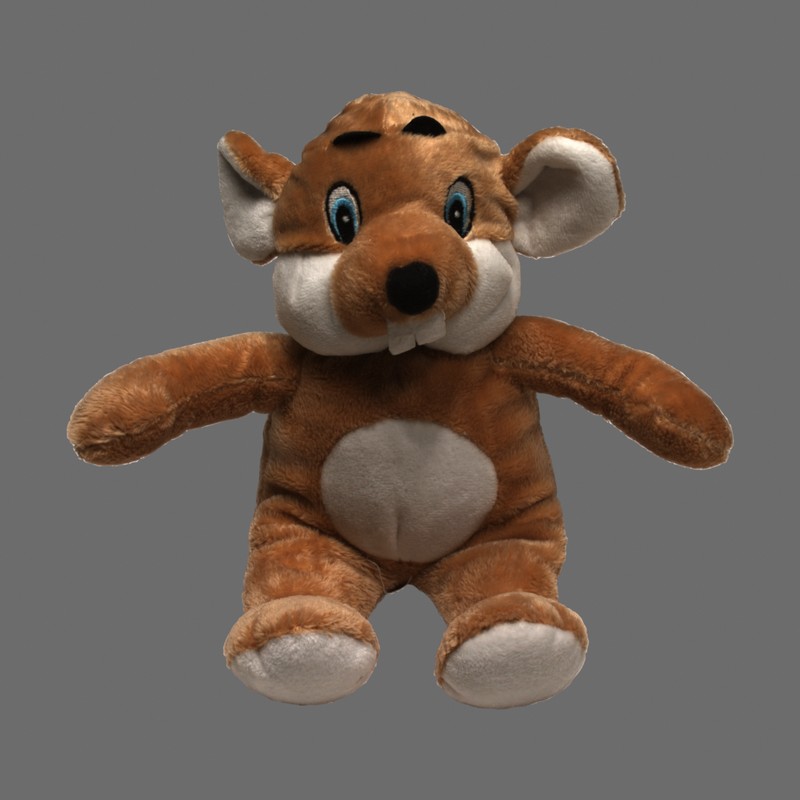} \\
        \includegraphics[width=0.19\textwidth]{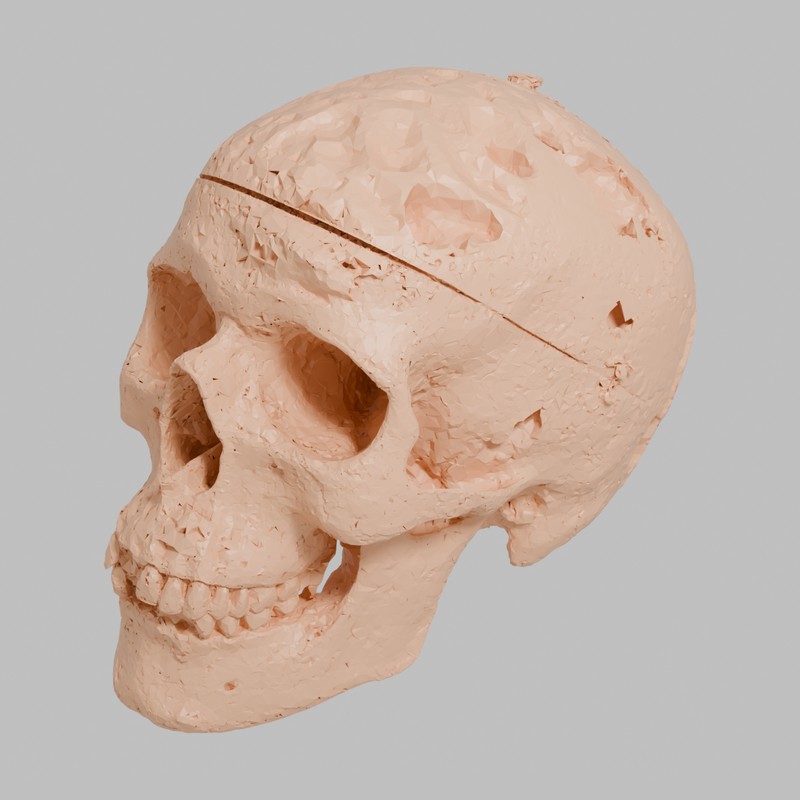} & \includegraphics[width=0.19\textwidth]{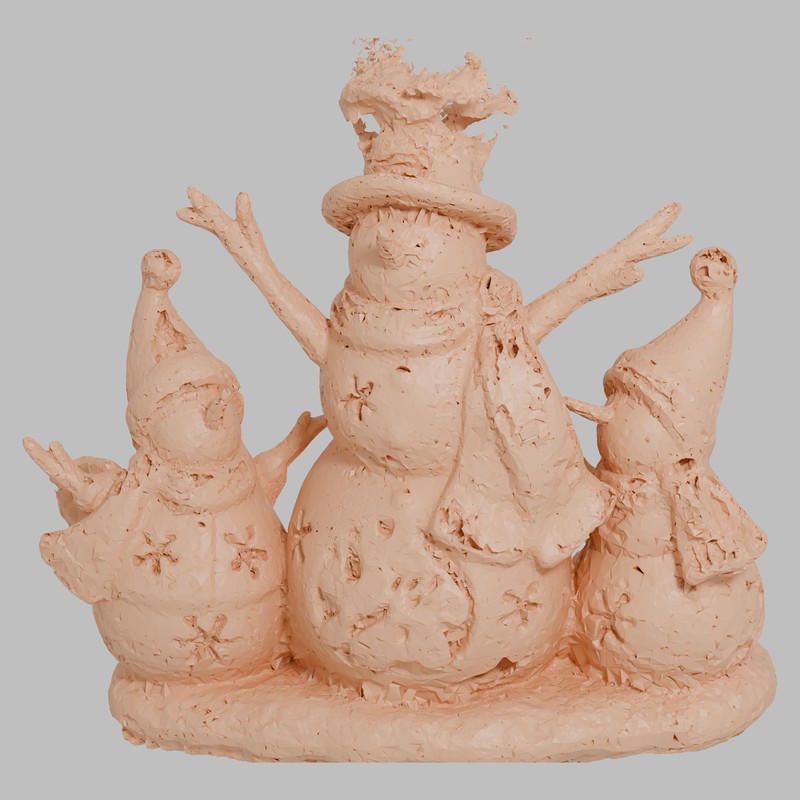} & \includegraphics[width=0.19\textwidth]{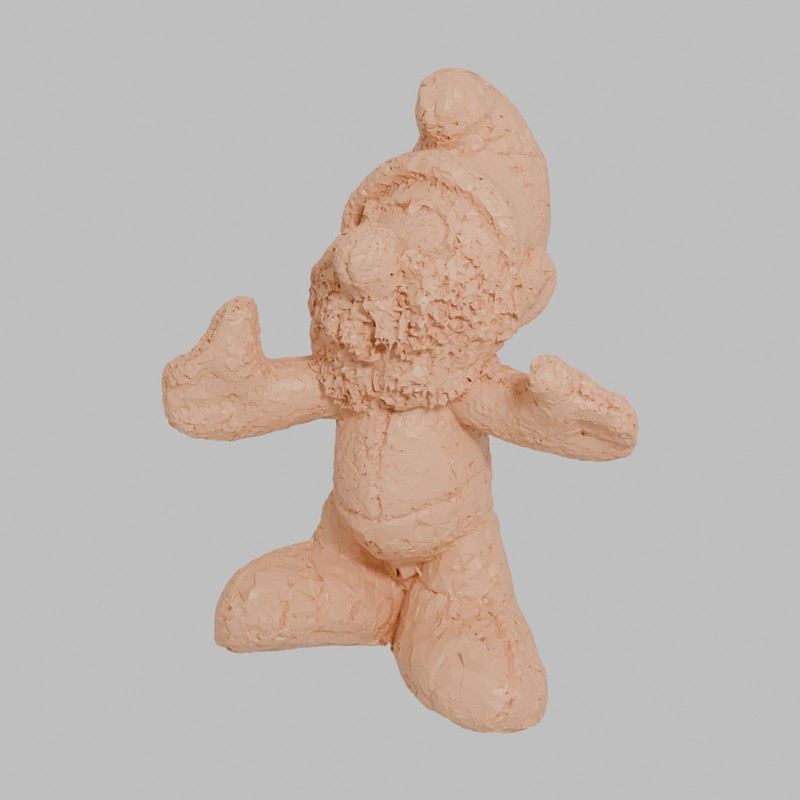} & \includegraphics[width=0.19\textwidth]{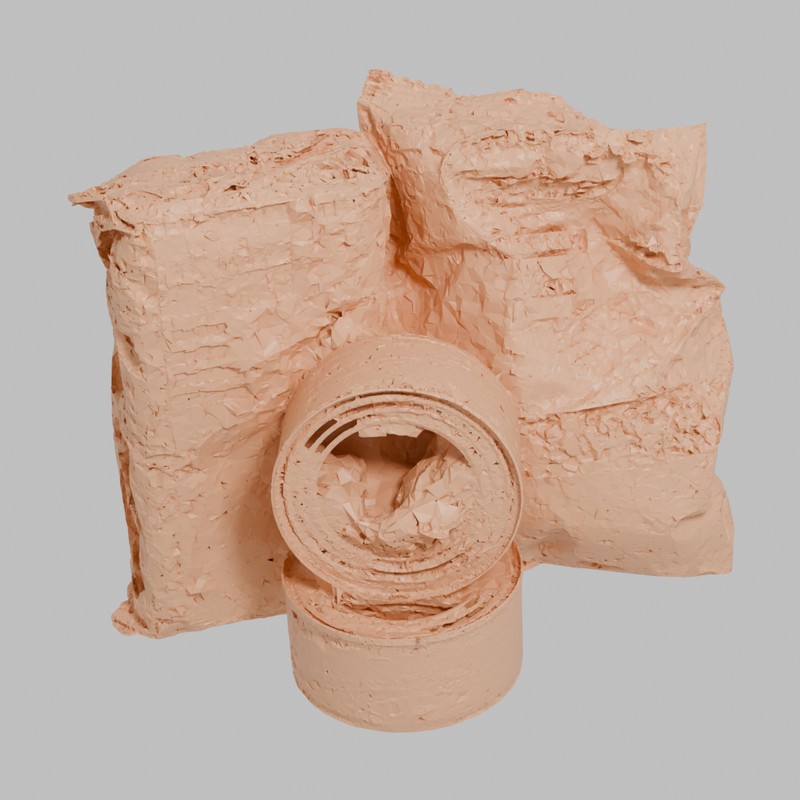} & \includegraphics[width=0.19\textwidth]{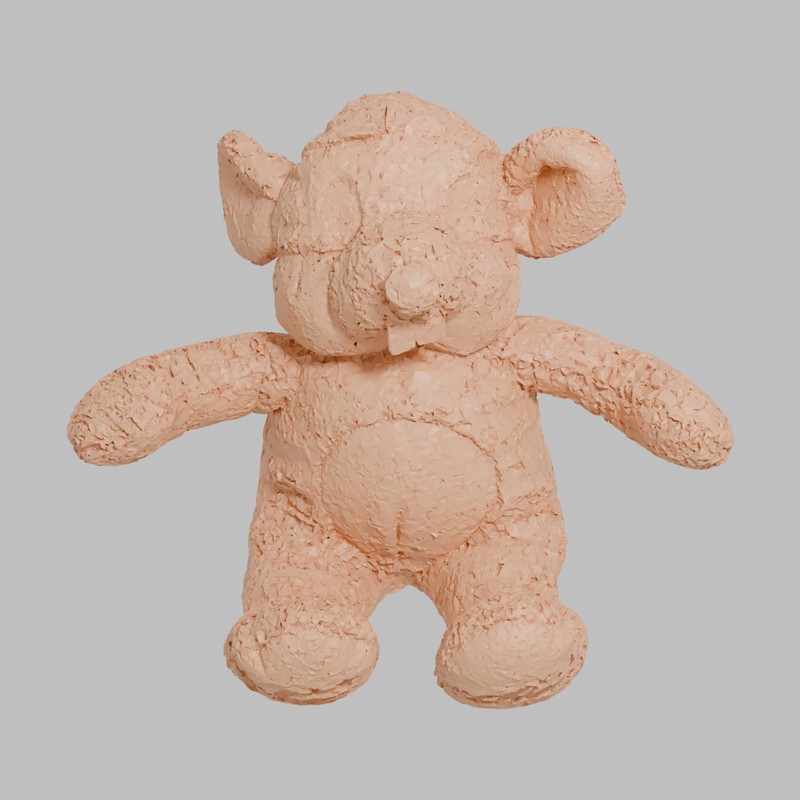} \\[4pt]
        \small 106 & \small 110 & \small 114 & \small 118 & \small 122 \\
        \includegraphics[width=0.19\textwidth]{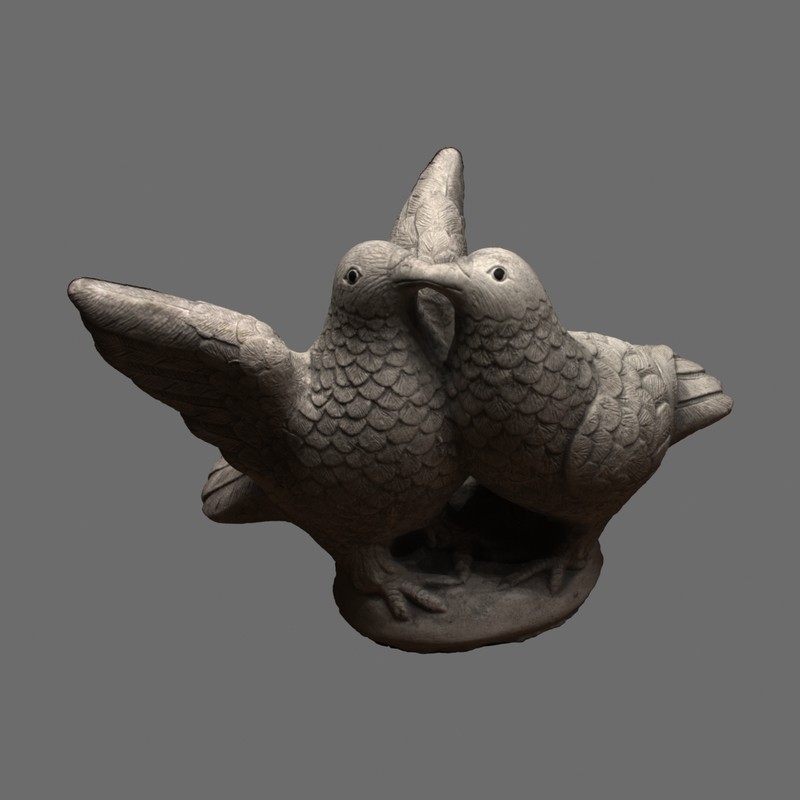} & \includegraphics[width=0.19\textwidth]{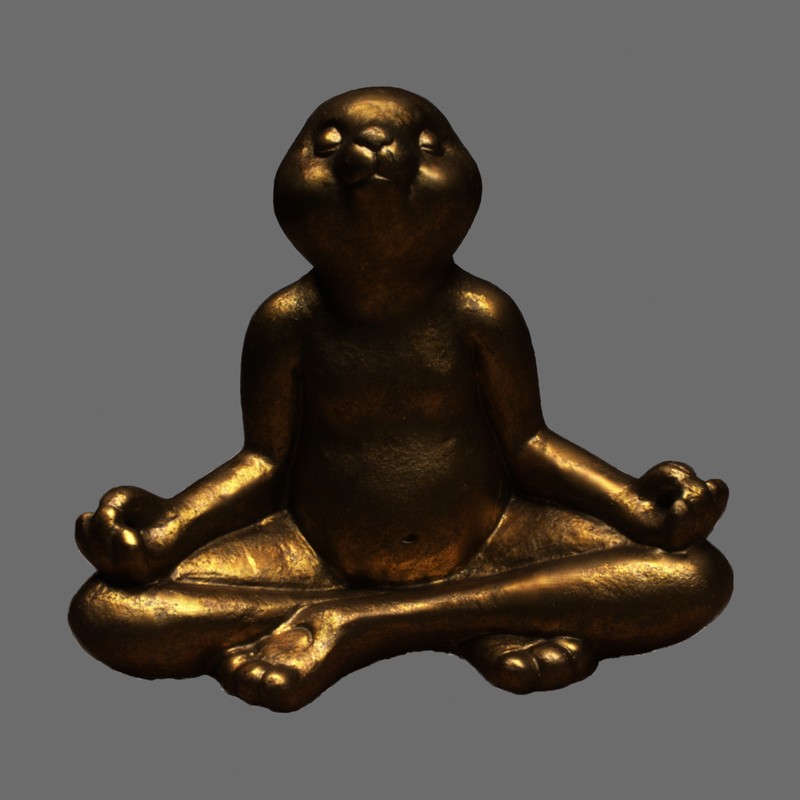} & \includegraphics[width=0.19\textwidth]{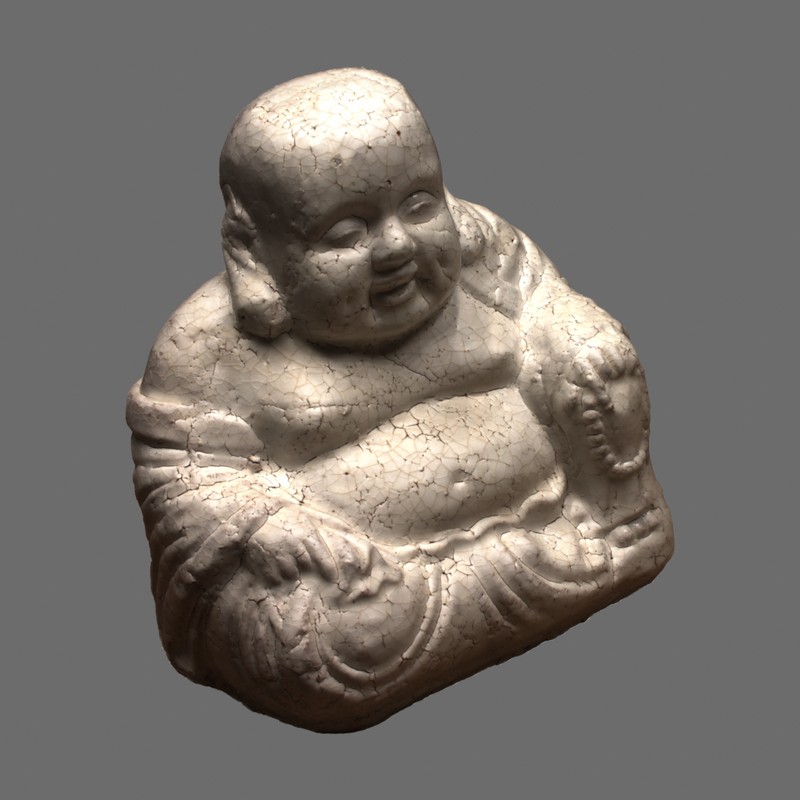} & \includegraphics[width=0.19\textwidth]{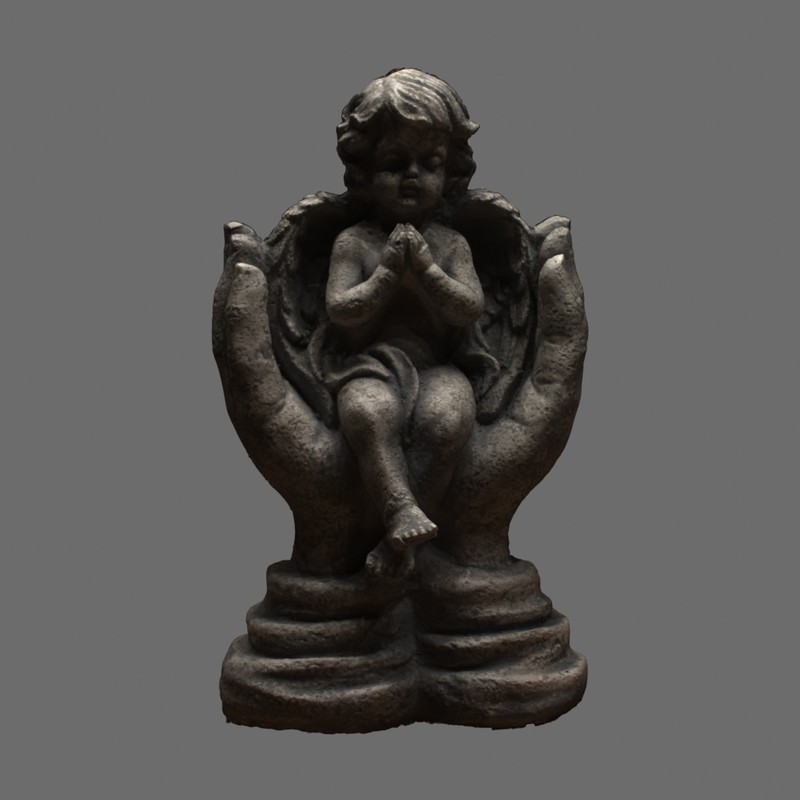} & \includegraphics[width=0.19\textwidth]{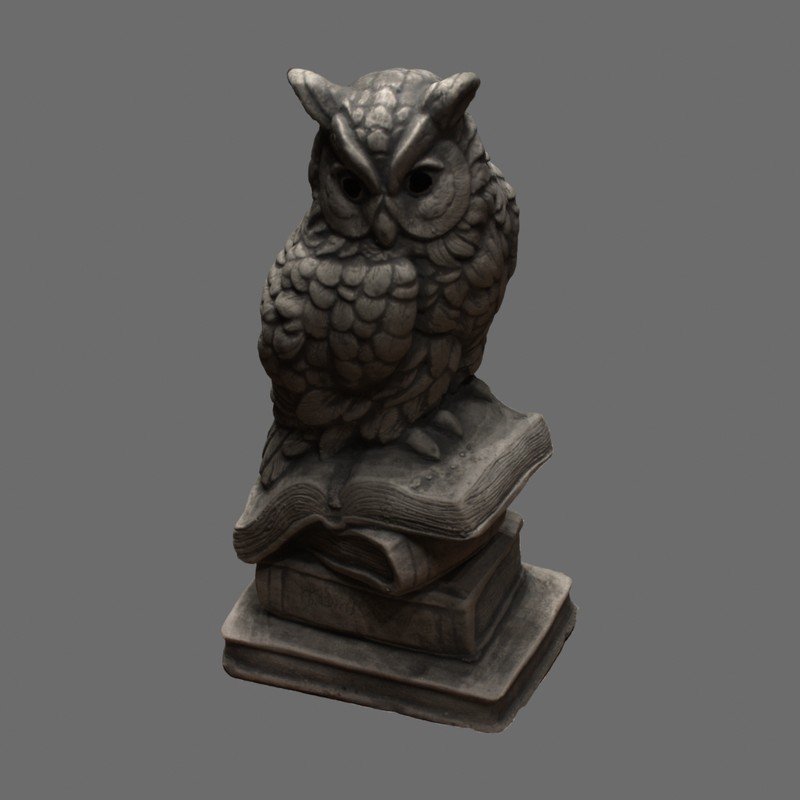} \\
        \includegraphics[width=0.19\textwidth]{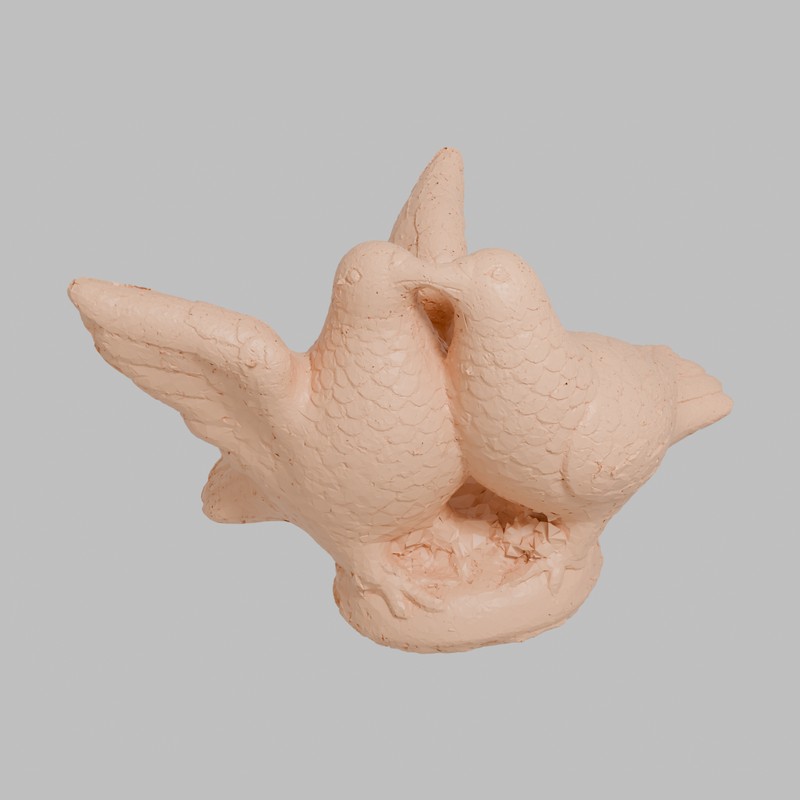} & \includegraphics[width=0.19\textwidth]{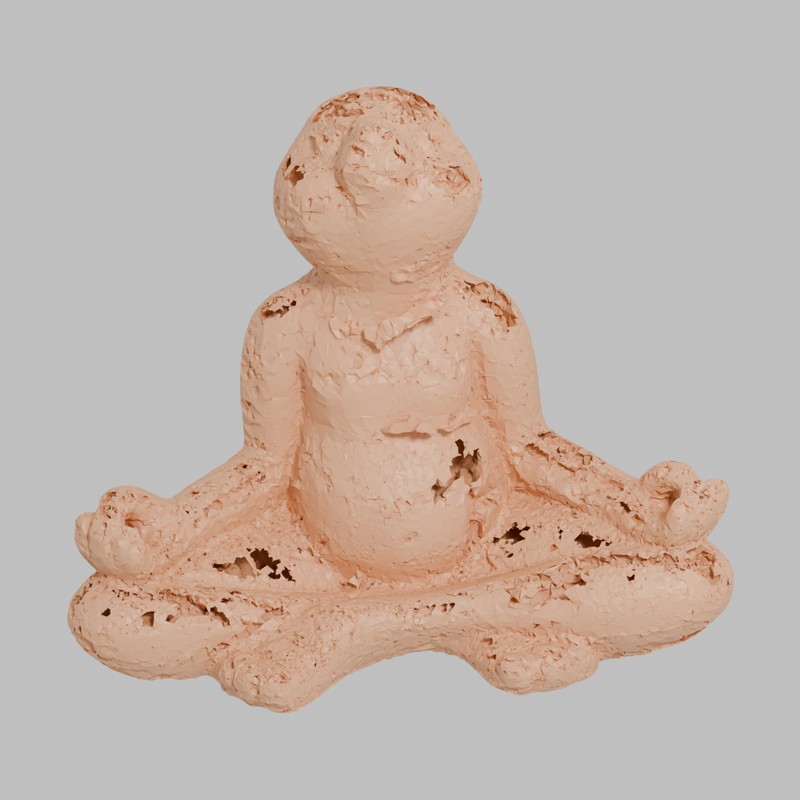} & \includegraphics[width=0.19\textwidth]{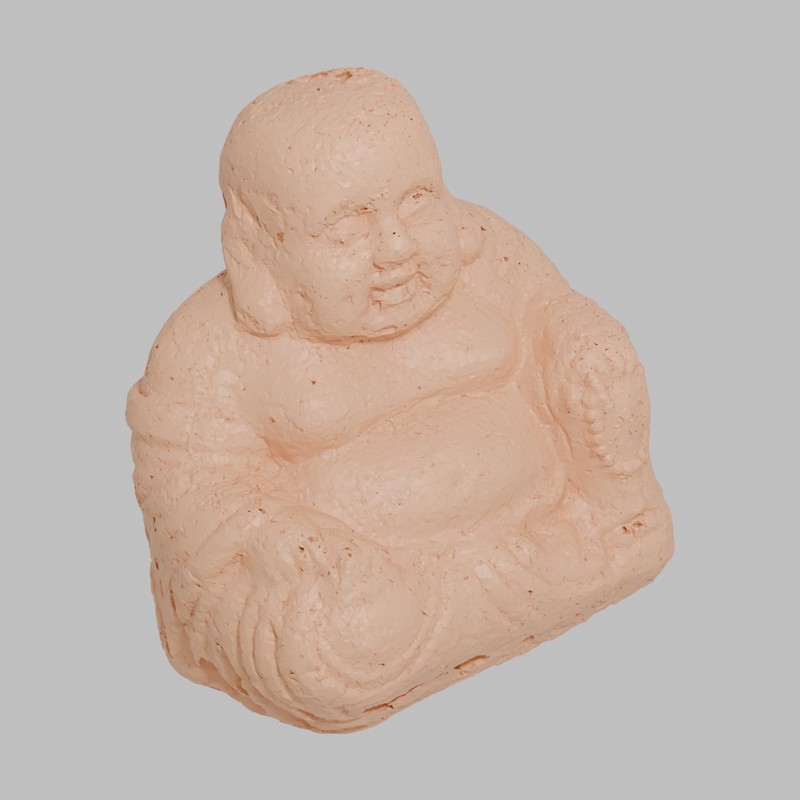} & \includegraphics[width=0.19\textwidth]{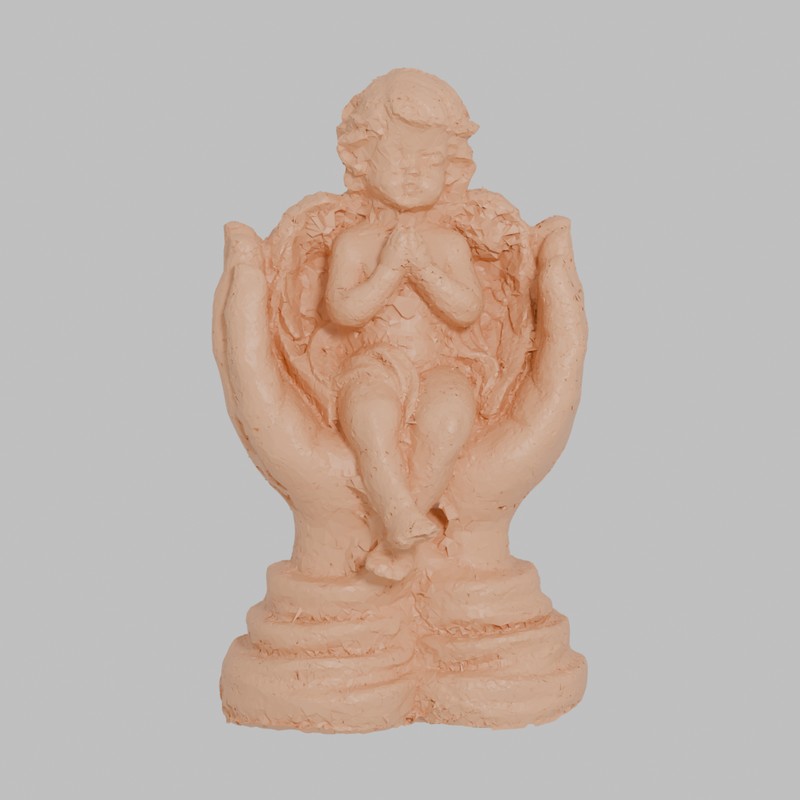} & \includegraphics[width=0.19\textwidth]{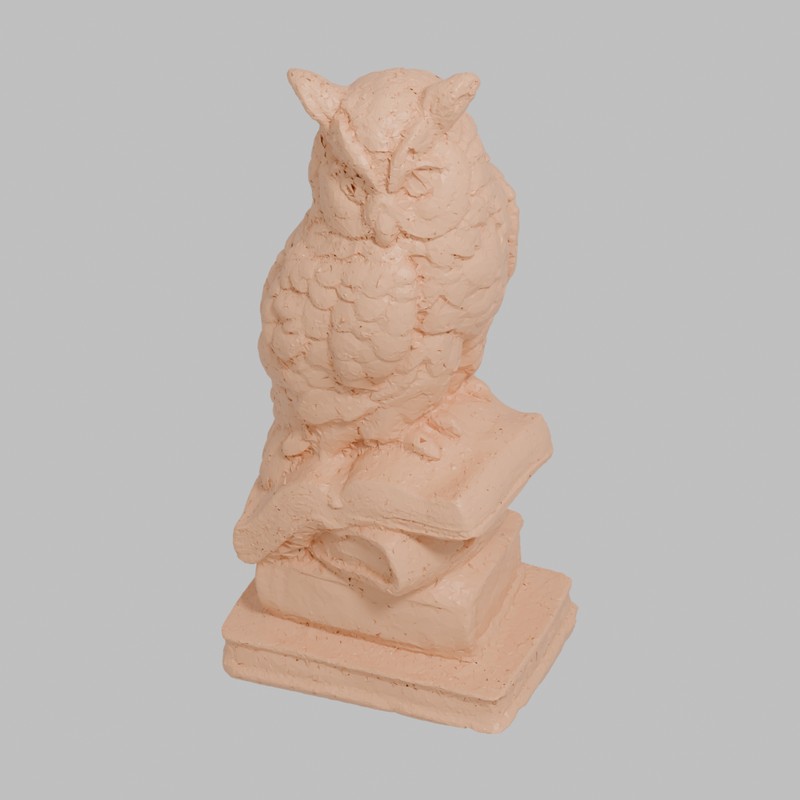} \\[4pt]
    \end{tabular}
    \caption{Textured mesh and mesh geometry for all 15 DTU scans.}
    \label{fig:suppl_dtu_all}
\end{figure*}

\section{Downstream Applications}
\label{sec:suppl_applications}
The interpretable texture atlas allows easy editing of the 2D texture, as seen in Fig.~\ref{fig:suppl_painting}. Our color model encourages the separation of view-dependent color from the underlying texture, which is why the edited texture also displays the learned view-dependent effects.
\begin{figure*}[tbp]
    \centering
    \includegraphics[width=0.49\textwidth]{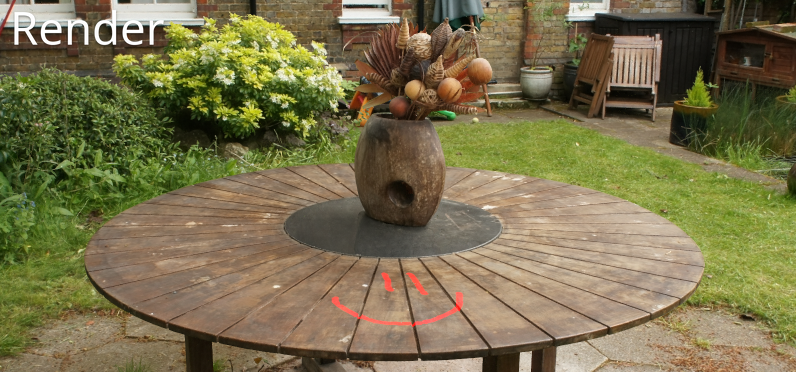}\hfill
    \includegraphics[width=0.49\textwidth]{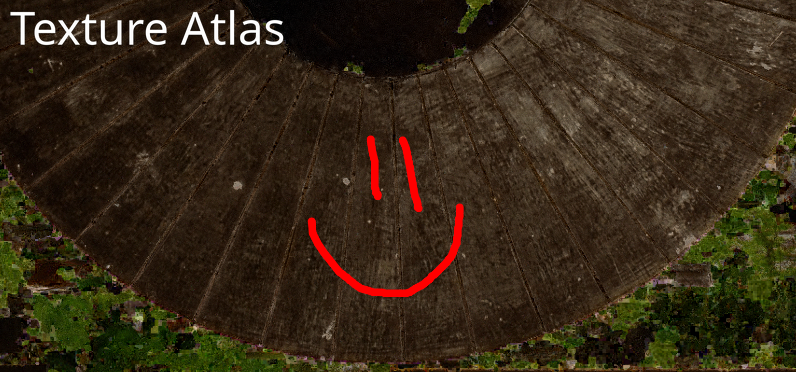}
    \caption{Texture editing. A smiley painted directly into the baked texture atlas (right) appears on the table in the rendered mesh (left) with the view dependence applied as expected.}
    \label{fig:suppl_painting}
\end{figure*}

The exported meshes in the GLB format can be imported directly into Blender. With a shader node for the directional color, the Blender editor view and rendering displays the same quality in novel view synthesis as reported in the paper, enabling editing in existing tooling. Fig.~\ref{fig:suppl_blender} shows the reconstructed \emph{stump} scene from Mip-NeRF 360 directly rendered in Blender.
\begin{figure*}[tbp]
    \centering
    \adjincludegraphics[width=\linewidth, trim={0 0 0 {0.029\height}}, clip]{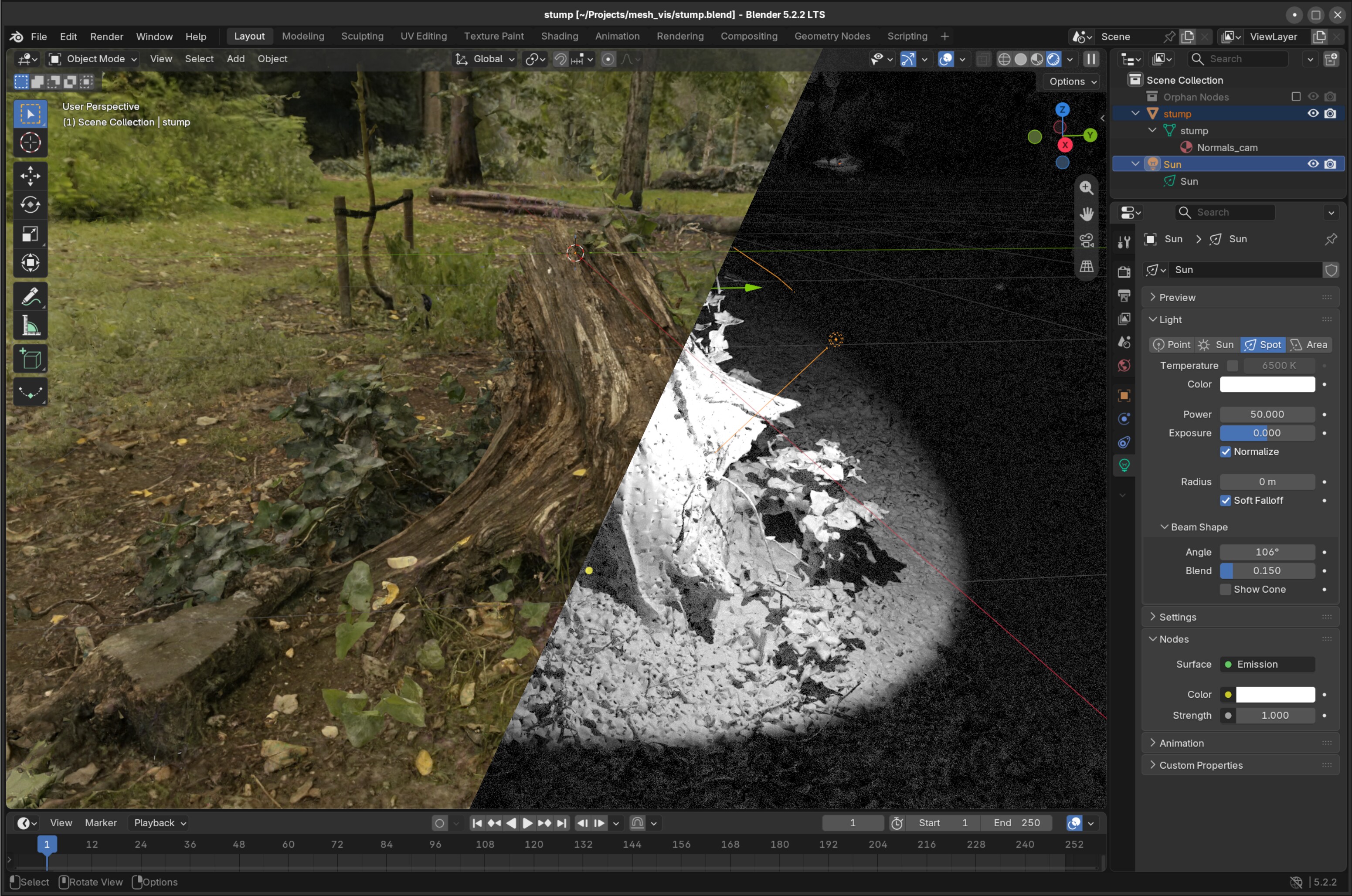}
    \caption{Mesh rendering in Blender. Our meshes can be loaded and rendered in Blender with directional color.}
    \label{fig:suppl_blender}
\end{figure*}

\clearpage
{
    \small
    \bibliographystyle{ieeenat_fullname}
    \bibliography{main}
}

\endgroup

\end{document}